\documentclass[10pt,letterpaper]{article}
\ifdefined\pdfminorversion\pdfminorversion=7\fi
\usepackage[T1]{fontenc}
\usepackage{mathptmx}
\usepackage{graphicx}
\usepackage{amsmath}
\usepackage{booktabs}
\usepackage{capt-of}
\usepackage{placeins}
\usepackage{needspace}
\usepackage{longtable}
\makeatletter
\let\paper@originalmakecaption\@makecaption
\def\paper@figuretype{figure}
\long\def\@makecaption#1#2{%
  \ifx\@captype\paper@figuretype
    \vskip6pt
    \begingroup
      \normalfont\fontsize{9}{11}\selectfont
      \raggedright
      \leftskip=8pt\relax
      \rightskip=8pt plus 1fil\relax
      \parskip=0pt\relax
      \textbf{#1.}\space#2\par
    \endgroup
    \vskip8pt
  \else
    \paper@originalmakecaption{#1}{#2}%
  \fi
}
\makeatother
\ifdefined\pdfimageresolution\pdfimageresolution=72\fi
\usepackage[left=1.5in,right=1.5in,top=1in,bottom=1in,headheight=12pt,headsep=16pt,footskip=28pt]{geometry}
\usepackage{microtype}
\usepackage{xcolor}
\usepackage{fancyhdr}
\usepackage{titlesec}
\usepackage[bookmarksnumbered,hidelinks,pdfversion=1.7]{hyperref}

\newcommand{\modelname}{Faynt}
\newcommand{\paperauthors}{Ali Janati, Nikita Kuzmin\\[3pt] Rohit Swamy\textsuperscript{$\dagger$}, Charles Niu\textsuperscript{$\dagger$}}
\newcommand{\papertitlelineone}{\modelname: Scaling and Optimizing}
\newcommand{\papertitlelinetwo}{Policies for Competitive Melee}

\newcommand{\modelfamilyurl}{https://huggingface.co/collections/frisson-labs/faynt}
\newcommand{\benchmarkurl}{https://github.com/Na00s/Faynt-Benchmarks}
\newcommand{\platformurl}{https://github.com/Na00s/Faynt-Tournament}
\hypersetup{
  pdftitle={\papertitlelineone\space\papertitlelinetwo},
  pdfsubject={Model training, evaluation, and behavior analysis},
  pdfauthor={Ali Janati, Nikita Kuzmin, Rohit Swamy, Charles Niu}
}

\definecolor{muted}{gray}{0.48}
\definecolor{resourcelink}{RGB}{37,65,93}
\titleformat{\section}{\large\bfseries}{\thesection}{1em}{}
\titlespacing*{\section}{0pt}{12pt}{7pt}
\titleformat{\subsection}{\normalsize\bfseries}{\thesubsection}{0.8em}{}
\titlespacing*{\subsection}{0pt}{8pt}{4pt}

\newcommand{\titlemark}{%
  \raisebox{-4pt}{\includegraphics[
    height=20pt,trim=138bp 149bp 115bp 173bp,clip
  ]{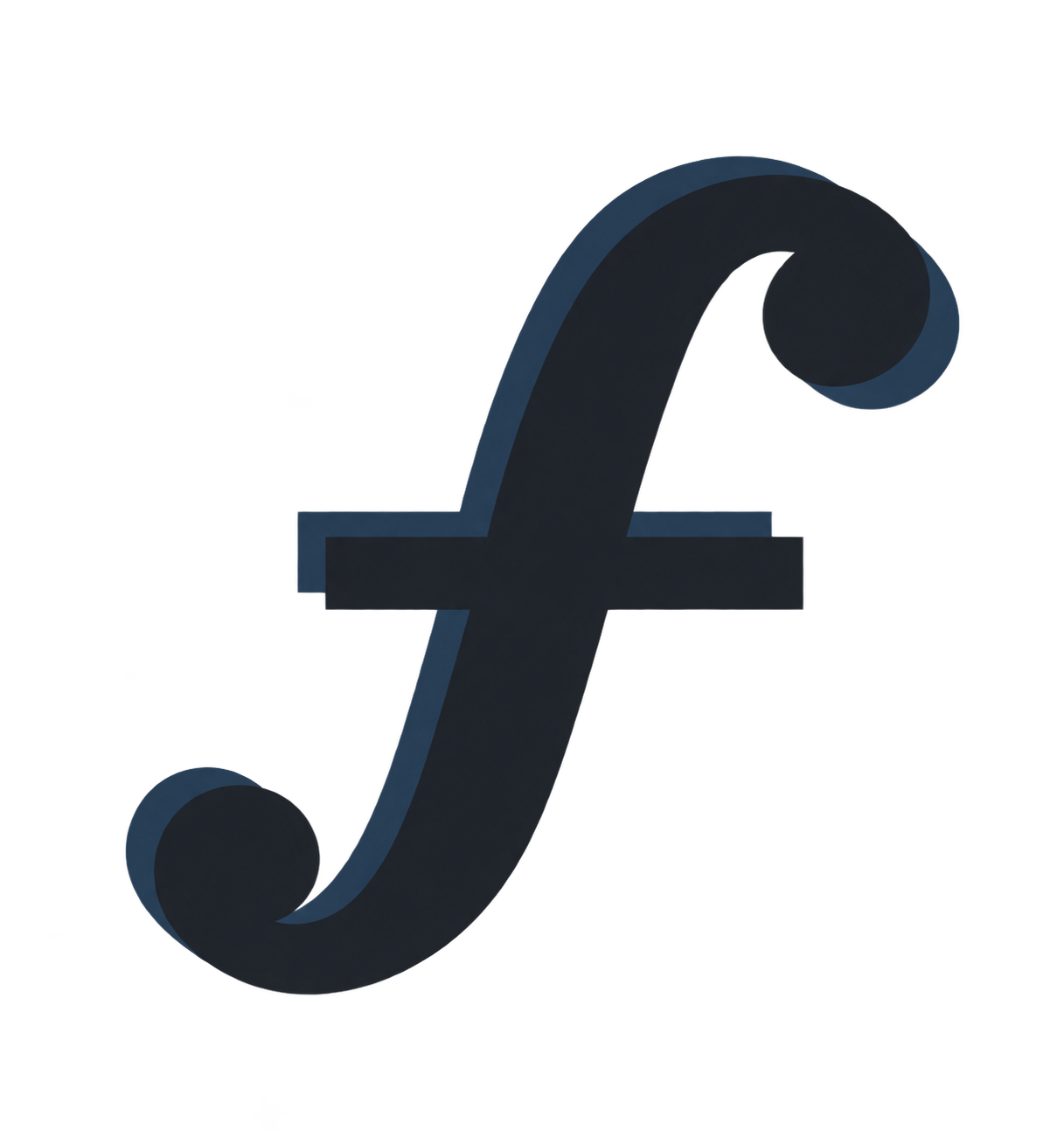}}%
}
\newlength{\titletextwidth}
\newlength{\titlesecondwidth}
\newcommand{\resourceicon}[1]{%
  \raisebox{-2pt}{\includegraphics[height=10pt]{assets/#1.pdf}}\hspace{5pt}%
}
\newcommand{\resourcelinks}{%
  {\centering\fontsize{9}{12}\selectfont
    \begin{tabular}{@{}l@{}}
      \href{\modelfamilyurl}{\resourceicon{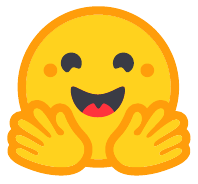}\textcolor{resourcelink}{Model checkpoints}}
    \end{tabular}\par
  }%
}
\newcommand{\titleblock}{%
  \vspace*{6pt}%
  \hrule height 4pt
  \vspace{12pt}%
  {\centering
    \fontsize{17.28}{22}\selectfont\bfseries
    \settowidth{\titletextwidth}{\papertitlelineone}%
    \settowidth{\titlesecondwidth}{\papertitlelinetwo}%
    \ifdim\titlesecondwidth>\titletextwidth
      \setlength{\titletextwidth}{\titlesecondwidth}%
    \fi
    \makebox[\textwidth][c]{%
      \titlemark\hspace{6pt}%
      \parbox[t]{\titletextwidth}{%
        \raggedright\microtypesetup{protrusion=false}%
        \papertitlelineone\\[1pt]%
        \papertitlelinetwo
      }%
    }\par
  }%
  \vspace{13pt}%
  \hrule height 0.6pt
  \vspace{20pt}%
  {\centering\normalsize\bfseries\paperauthors\par}%
  \vspace{4pt}%
  {\centering\normalsize Frisson Labs\par}%
  \vspace{17pt}%
  \resourcelinks
  \vspace{17.37pt}%
  {\centering\large\bfseries Abstract\par}%
  \vspace{9pt}%
  {\normalsize
    \leftskip=20pt\relax
    \rightskip=20pt\relax
    We introduce Faynt, a family of 10M- and 75M-parameter Transformer policies
    for Super Smash Bros. Melee, each controlling all 26 characters with a
    single checkpoint. After reinforcement learning (RL), the 10M wins 240 of 244
    same-character games (98.4\%) against fourteen specialist and multi-character
    releases on their supported rosters, with a winning record against every
    release. These opponents retain 21- or 24-frame action delays; Faynt
    uses no added delay, and we have not isolated the effect of this difference.
    In a separate evaluation against a
    privately supplied zero-delay Slippi-AI model, the 10M wins all 68 games
    across two conditioning settings. We study architecture, optimization,
    scaling, and hyperparameter transfer to guide pretraining on approximately
    840,000 human replays. Post-training combines rank- and outcome-based
    curricula, 75M-to-10M distillation, and RL restricted to Fox mirror matches.
    On the initial 152-game benchmark, the supervised 10M wins 69.7\%
    of games, compared with 45.4\% for the pretrained 75M, despite higher overall
    held-out controller-prediction loss. The weighted validation loss used for
    supervised checkpoint selection agrees with the win-rate ordering of all
    four pretrained and supervised policies. After supervised post-training,
    both models take less damage per minute, build
    larger early leads, and win more often after losing the first life.
    Optimized inference on recorded game states averages 5.2\,ms per decision
    for the 10M and 8.7\,ms for the 75M on an NVIDIA T4, excluding emulator
    execution and communication. We open-source the weights,
    \href{\benchmarkurl}{\textcolor{resourcelink}{both benchmark suites}}, and
    \href{\platformurl}{\textcolor{resourcelink}{a platform for automated model tournaments}}.
    \par
  }%
  \vspace{20pt}%
}

\newcommand{\introductiontext}{%
Super Smash Bros. Melee requires precise control over movement and attacks while
responding to an opponent's decisions. Earlier work demonstrated that
reinforcement-learning agents could compete with professional
players~\cite{firoiu2017}. Slippi-AI subsequently combined imitation learning
from human replays with self-play reinforcement learning. We build on its
structured state representation and autoregressive controller
interface~\cite{slippiai}.
\par
We introduce Faynt, a family of 10M- and 75M-parameter Transformer policies
pretrained on approximately 840,000 ranked human replays. Each RL policy
controls all 26 characters using a single checkpoint. On opponents' supported
characters, the 10M wins 240/244 mirror games against fourteen releases,
with winning records against all fourteen, including all five
specialists. These opponents retain 21- or 24-frame action delays, while
Faynt uses zero added delay. The 10M also wins all 68
games against a privately supplied zero-delay Slippi-AI checkpoint evaluated
under two conditioning settings (Section~\ref{sec:benchmark-zero-delay}). Both models also have
winning full-roster mirror records against all fourteen releases, including
matches on characters outside the opponents' deployed rosters. Some specialists
retain an advantage against the 75M on their native fighters
(Section~\ref{sec:benchmark}). We study how model size, demonstration selection,
distillation, and RL affect controller prediction and playing strength.
\par
Our pretraining study examines optimization and the allocation of
compute between model capacity and training duration. We compare model sizes
and optimizers, conduct hyperparameter sweeps, and measure how selected settings
transfer across model sizes. Using fixed-compute comparisons, as in studies of
language-model scaling, we fit empirical laws relating controller-prediction loss
to model size and processed training frames~\cite{hoffmann2022}. These fits,
together with measured hardware throughput, guide the choice of model sizes and
training budgets. We then compare their predictions with the outcomes of the
final pretraining runs.
\par
Post-training introduces changes to the demonstration distribution and the
learning objective. We first continue supervised training at a lower learning
rate, using a curriculum that increasingly emphasizes wins by higher-ranked
players. We then distill the 75M policy into the 10M, combining recorded
controller targets with the teacher's action distributions~\cite{rusu2016}.
In our experiments, distillation provides further
reductions in held-out prediction loss after gains from supervised continuation
have narrowed. Reinforcement learning follows these stages and optimizes the
policies using rewards from gameplay. RL uses only Fox mirror matches,
with self-play and, in the first runs, the built-in CPU opponent. The final
checkpoints are evaluated across all 26 characters. The two sizes follow
different RL training histories, which we describe in
Section~\ref{sec:reinforcement-learning}.
\par
Our evaluation first compares pretrained and supervised post-trained Faynt
policies with Slippi-AI and MIMIC in a defined 152-game
suite~\cite{slippiai,mimic}. On this suite, the supervised 10M
scores 69.7\%, compared with 45.4\% for the pretrained 75M, despite slightly
higher overall held-out imitation loss. After RL, we expand the opponent set to specialists and
multi-character policies at different skill levels, testing both full-roster
mirrors and additional-character play against opponents on supported fighters.
The final expanded suite contains 1,312 games per RL model, totaling 2,624 games
(Section~\ref{sec:benchmark}).
\par
We also study whether the policies can make a decision within Melee's
16.67\,ms frame budget. Removing host synchronization and replaying computation
through CUDA graphs reduces inference overhead while preserving the policy's
action distributions to within floating-point rounding. On recorded game
states, the optimized T4 decision loop averages 5.2\,ms at 10M and 8.7\,ms
at 75M, with 99th-percentile latencies of 6.8 and 12.1\,ms. These timings
exclude the emulator and communication with it
(Section~\ref{sec:benchmark-realtime}).
\par
We also examine how play changes across training. After supervised
post-training, both policies take less damage per minute, build larger leads
in the opening minute, and win a larger share of games after losing the first
life, or \emph{stock} (Section~\ref{sec:behavior}).
\par
We release the model weights, both benchmark suites, and software for
policy-versus-policy matches and mirrored baseline tournaments. The released
adapters support Faynt comparisons with MIMIC, Slippi-AI, and CPU9;
Appendix~\ref{app:platform} describes their interfaces, scheduling, and
required inputs.
\par
}

\newcommand{\taskandarchitecturetext}{%
\section{Task, Dataset, and Policy Architecture}
\label{sec:task-data-architecture}

\subsection{Task Formulation and Policy Interface}
A policy controls one player in a two-player Melee match. We pretrain it on
sequences of parsed Slippi (\texttt{.slp}) frames. Each frame contains
structured game-state features $s_t$ for both players, companion characters
where applicable, the stage and platforms, and items, together with the
controlled player's current controller input $a_t$. From their history
$H_t=(s_{\leq t},a_{\leq t})$, the policy predicts a distribution over the
next controller command $a_{t+1}$. Imitation training uses 256-frame windows of these
state and controller observations. During imitation training, both come from
human replays; during gameplay, they come from the live game state and the
policy's executed commands.
\par
We use Slippi-AI's \texttt{custom\_v1} controller
discretization~\cite{slippiai}, writing $a_t=(b_t,m_t)$, where
$b_t\in\{0,\ldots,727\}$ encodes a combination of buttons, shoulder pressure,
and C-stick position, and $m_t\in\{0,\ldots,84\}$ encodes the main-stick
position. The predicted pair is decoded into a GameCube controller command.
\par
Our objective is to develop the strongest possible learned Melee policy.
Adding reaction delay would limit its responsiveness and therefore work
against this objective, so we choose zero added policy delay. Our evaluation
harness makes this possible by buffering whichever agent's controller command
arrives first and advancing the Dolphin emulator only once both commands are
available. This synchronization preserves the policy's intended timing within
the game, while inference and communication time affect only the wall-clock
duration of evaluation (Appendix~\ref{app:benchmark-runtime}).
Section~\ref{sec:benchmark-realtime} evaluates whether the decision loop
also fits within a 60\,Hz frame budget without this synchronization.
\par

\subsection{Human Replay Dataset}
We construct the corpus from a fixed snapshot of
\emph{Melee Ranked Replays}, a public collection of anonymized Slippi ranked
matches at Platinum rank and above~\cite{meleereplays}. We select archive
groups by character and rank pair and retain complete games. Deduplication
removes byte-identical files before parsing and replays with identical parsed
training content afterwards. We extract states and controller inputs from
both players' perspectives using a fixed version of the Slippi-AI
parser~\cite{slippiai}. Rank and outcome metadata supply the labels used in
the curricula of Section~\ref{sec:posttraining}.
\par
The final corpus contains 839,942 replays and 17.84 billion valid targets,
including 17.48 billion in the training split. A target is one
player-perspective transition from frame $t$ to $t+1$. For the scaling and
hardware studies, we use a fixed calibration build containing 67,129 replays
and 1.50 billion targets. Training samples can revisit the same frames, so
we distinguish corpus size from the number of targets processed during a run.
\par
Within each build, a deterministic hash assigns entire games to training,
validation, and test splits in approximate proportions of 98\%, 1\%, and 1\%.
Keeping both perspectives in the same split prevents a match from appearing
in training and evaluation through different players. Training windows stay
within a single game. On the final corpus, replay evaluation uses one fixed
256-frame window per held-out player perspective, giving 16,966 validation
windows and 16,752 test windows. The validation windows are held constant
across pretraining and post-training. Test windows are evaluated only after
checkpoint selection.
\par

\subsection{Policy Architecture}
The final 10M and 75M policies follow the same architecture
(Figure~\ref{fig:architecture}), with the dimensions listed in
Table~\ref{tab:model-config}. Section~\ref{sec:pretraining} presents the
model-size and optimization studies that guide these configurations.
\par
\textbf{Frame encoder.}
Following Slippi-AI's structured state representation~\cite{slippiai}, player
and companion features include character identity, in-game action state,
position, facing direction, damage percentage, shield strength, and remaining
jumps. We use learned character and action-state embeddings, shared across
these records. Each action-state embedding receives an additional embedding
indexed by the character--state pair. World features encode the stage,
platform coordinates, and items. Item features pass through a shared
multilayer perceptron and are summed after masking absent items. Concatenating
player, world, and one-hot controller features gives a 2,091-dimensional
vector. A bias-free projection maps this vector to the model width, producing
one temporal token per game frame.
\par
\textbf{Temporal backbone.}
The causal Transformer~\cite{vaswani2017} uses grouped-query attention~\cite{ainslie2023},
RMSNorm~\cite{zhang2019}, rotary position embeddings~\cite{su2021}, and
SwiGLU feed-forward layers~\cite{shazeer2020}. Queries and keys are
RMS-normalized before rotary embeddings, a variant of query--key
normalization~\cite{henry2020}. An elementwise sigmoid
gate scales each attention output before its output projection~\cite{qiu2025}.
Before each attention and feed-forward sublayer, Full Attention
Residuals~\cite{kimi2026} compute a weighted sum of the projected frame input
and all preceding sublayer outputs at the same frame. Learned queries attend
to RMS-normalized versions of these representations to determine the mixing
weights. A final weighted sum followed by RMSNorm produces the backbone
output $h_t$. Imitation training uses a 256-frame context.
\par
\vspace{4pt}
\begin{table}[!htbp]
  \centering
  \small
  \begin{tabular}{@{}lrr@{}}
    \toprule
    Configuration & Faynt-10M & Faynt-75M \\
    \midrule
    Trainable parameters & 10,163,629 & 75,305,709 \\
    Model width & 384 & 768 \\
    Transformer blocks & 5 & 11 \\
    Query / key-value heads & 6 / 2 & 12 / 3 \\
    Head dimension & 64 & 64 \\
    Feed-forward width & 768 & 1,920 \\
    Imitation-training context (frames) & 256 & 256 \\
    \bottomrule
  \end{tabular}
  \vspace{4pt}
  \caption{Model configurations. Parameter counts include the frame
  encoder, Transformer, and controller heads.}
  \label{tab:model-config}
\end{table}
\par\vspace{5pt}
\textbf{Controller heads and training objective.}
We follow Slippi-AI's autoregressive decoding design~\cite{slippiai}: first
predict the joint-controller category, then condition the main-stick
prediction on it.
The backbone output is projected to a 128-dimensional vector $r_t$. The
joint-controller head receives $[r_t,\operatorname{onehot}(b_t)]$. The
main-stick head receives
$[r_t+E(b_{t+1}),\operatorname{onehot}(m_t)]$, where $E$ is a learned
embedding of the next joint-controller category. Each head has two
128-dimensional hidden layers with ReLU activations. The resulting
distribution factors as
\begin{equation}
\pi_\theta(a_{t+1}\mid H_t)
=p_\theta(b_{t+1}\mid h_t,b_t)\,
 p_\theta(m_{t+1}\mid h_t,m_t,b_{t+1}).
\label{eq:controller-factorization}
\end{equation}
The stick head conditions on the recorded next joint-controller target during
imitation training and the sampled category during play.
\par
\noindent\begin{minipage}{\textwidth}
Pretraining uses behavioral cloning, minimizing the mean negative
log-likelihood (NLL) over valid targets
$\mathcal{T}$:
\begin{equation}
\mathcal{L}_{\mathrm{BC}}(\theta)
=-\frac{1}{|\mathcal{T}|}\sum_{t\in\mathcal{T}}
\log\pi_\theta(a_{t+1}\mid H_t).
\label{eq:bc-objective}
\end{equation}
\end{minipage}
\par
The two head cross-entropies enter this loss with equal weight; we report
their sum in nats per target. Appendices~\ref{app:data} and~\ref{app:architecture}
specify the state features, controller codec, parser version, dataset manifests,
and sampling procedure.
\par
\vspace{5pt}
\begin{figure}[!htbp]
  \centering
  \includegraphics[width=\textwidth]{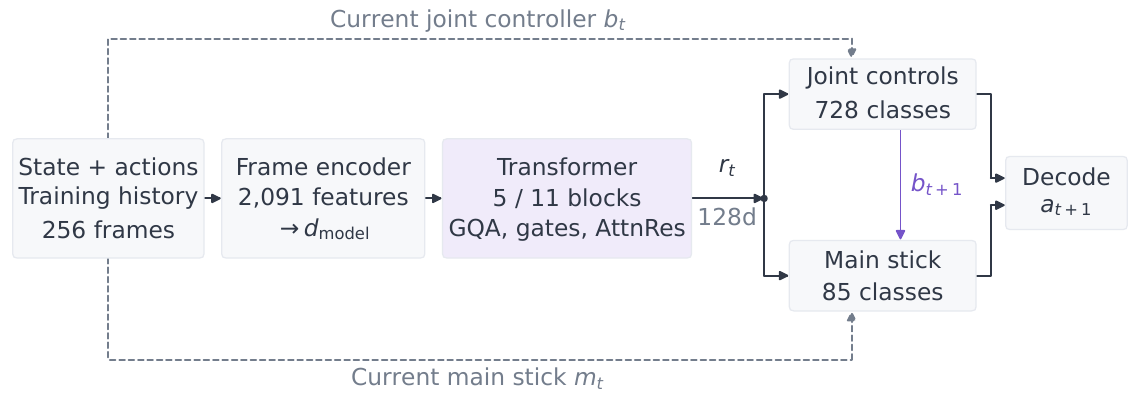}\par
  \caption{\textbf{Faynt policy architecture.} Dashed connections
  supply current controller categories directly to the output heads. The
  main-stick head also conditions on the next joint-controller category,
  using the recorded target during imitation training and a sample during play.
  AttnRes denotes Full Attention Residuals.}\label{fig:architecture}
\end{figure}
\par
}

\newcommand{\scalingfigure}{%
  \par\penalty0\vskip5pt
  \begin{figure}[!htbp]
    \centering
    \includegraphics[width=\textwidth]{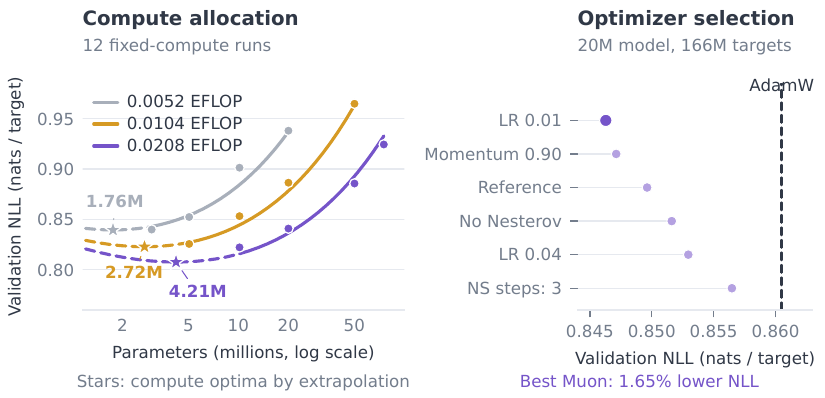}\par
    \caption{\textbf{Compute allocation and optimizer selection.}
    Left: observed endpoints and fitted fixed-compute curves; dashed segments
    extend below the tested size ranges, and stars mark optima found by
    extrapolation. Right: six Muon configurations at 20M parameters and
    166M targets, with the best AdamW run as reference. Muon's reference uses
    learning rate 0.02, momentum 0.95, Nesterov, and five Newton--Schulz iterations;
    row labels identify individual changes.}\label{fig:compute-scaling}
  \end{figure}\par
}
\newcommand{\halvingfigure}{%
  \par\penalty0\vskip5pt
  \begin{figure}[!htbp]
    \centering
    \includegraphics[width=\textwidth]{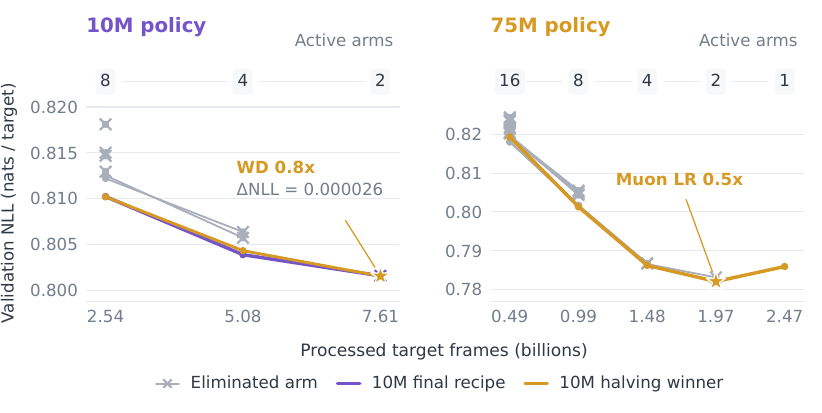}\par
    \caption{\textbf{Successive halving on the full corpus.}
    Curves join scheduled validation evaluations and end when a recipe is
    eliminated (crosses); stars mark the halving selections. Counts give active
    trials. The 10M panel shows selection through rung 3: lower weight decay
    wins narrowly, while lower Muon learning rate is ultimately selected after
    recovery. The 75M winner continues alone in rung 5.}\label{fig:halving}
  \end{figure}\par
}
\newcommand{\pretrainingfigure}{%
  \par\penalty0\vskip5pt
  \begin{figure}[!htbp]
    \centering
    \includegraphics[width=\textwidth]{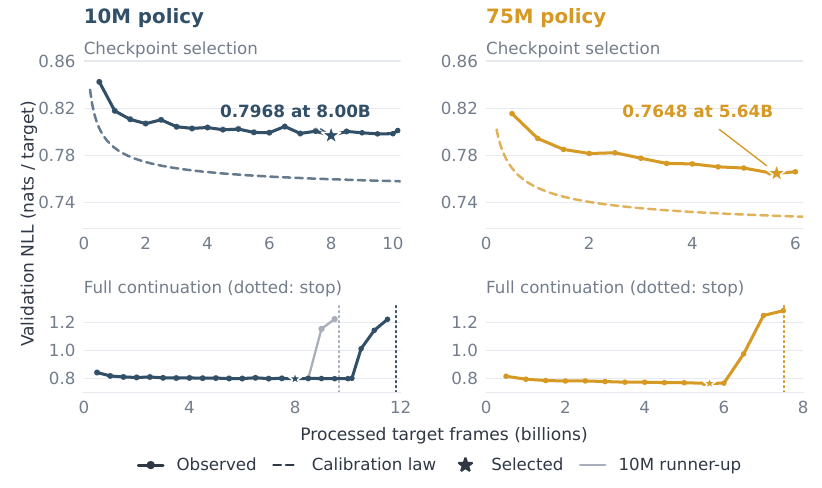}\par
    \caption{\textbf{Final pretraining runs.}
    Top: validation trajectories before divergence and calibration-law
    predictions; stars mark the selected checkpoints. Bottom: full
    continuations, with manual stops dotted. The 10M finalists (navy: lower
    Muon learning rate; gray: lower weight decay) remain within 0.0005 NLL
    through 8B targets, then differ in stability. Points are recorded evaluations;
    lines join successive observations.}\label{fig:pretraining-trajectories}
  \end{figure}\par
}
\newcommand{\pretrainingtext}{%
\section{Pretraining: Optimization and Scaling}
\label{sec:pretraining}
We use the calibration corpus to determine how to allocate pretraining compute
between model size and training exposure, and how to transfer optimizer settings
across sizes. The resulting loss and throughput estimates guide the choice of
two policies, whose recipes we then tune on the full corpus.

\subsection{Optimizer Selection and Hyperparameter Transfer}
We compare AdamW~\cite{adamw2019} and Muon~\cite{muon2024} using a 20M policy
trained for 166M targets. An initial AdamW grid crosses three learning rates with
batches of 65,536 and 262,144 targets, followed by one higher-rate trial. Model
initialization, data, validation windows, and frame budget are shared. The smaller
batch gives lower validation NLL at every matched learning rate; the best AdamW
run reaches 0.8605 at a learning rate of 0.001.
\par
With the batch fixed at 65,536 targets, we test six Muon configurations. Muon
updates the backbone's hidden weight matrices, while auxiliary AdamW updates the
encoder, decoder, embeddings, and remaining parameters. The selected configuration
uses a Muon learning rate of 0.01, auxiliary rate of 0.001, weight decay of 0.01,
momentum of 0.95, Nesterov momentum, and five Newton--Schulz iterations. It reaches
0.8463 validation NLL, a 1.65\% reduction relative to the selected AdamW run in
this single-seed screen (Figure~\ref{fig:compute-scaling}, right).
\par
To transfer this recipe, we run seven trials each at 5M and 50M for 41.94M
targets, varying each learning rate and weight decay separately by
$\{0.5,1,2\}\times$. Each axis favors $2\times$ at 5M and $1\times$ at 50M.
We summarize this pattern with the shared multiplier
$s(N)=\max\{1,(N/20{,}040{,}877)^{-0.5031}\}$ for exact parameter count $N$.
We multiply the three 20M settings by $s(N)$ to initialize subsequent searches
(Appendix~\ref{app:optimizer-screens}).

\subsection{Scaling Model Size and Training Exposure}
We train four model sizes at each of three compute budgets, 0.0052, 0.0104, and
0.0208 EFLOP, for twelve runs spanning 3M to 75M parameters. At each budget,
the smallest tested model achieves the lowest validation loss. Their schedules use
1\% linear warmup and cosine decay to 10\% of the peak, normalized to each run's
frame horizon. Following parametric scaling studies~\cite{hoffmann2022}, we fit
validation loss jointly across all endpoints:
\begin{equation}
\begin{aligned}
\widehat L(N,D)&=L_\infty+A_N(N/N_0)^{-\alpha}+A_D(D/D_0)^{-\beta},\\
C&=D\left(F_{\mathrm{model}}+F_{\mathrm{NS}}/B\right),
\end{aligned}
\label{eq:pretraining-scaling}
\end{equation}
where $D$ counts processed targets, including repeated exposure,
$N_0=10^7$, $D_0=10^8$, and $B=65{,}536$. Compute includes forward and backward
passes per target, $F_{\mathrm{model}}$, and Muon's Newton--Schulz work per update,
$F_{\mathrm{NS}}$. Each run uses its architecture's discrete FLOP count.
\par
The fit gives $L_\infty=0.6433$, $A_N=0.1017$, $A_D=0.1275$,
$\alpha=0.2024$, and $\beta=0.4859$, with an in-sample RMSE of 0.0041 NLL.
For continuous optimization, we fit compute per target as
$\widehat G(N)=6.1654\times10^7(N/10^7)^{1.168}$ FLOPs. We find the compute-optimal model sizes and training budgets by
extrapolating the fitted loss surface. The three optima are 1.76M, 2.72M, and
4.21M parameters, paired with 643M, 772M, and 926M targets
(Figure~\ref{fig:compute-scaling}). The resulting allocation scales as
$N_{\mathrm{opt}}\propto C^{0.631}$ and $D_{\mathrm{opt}}\propto C^{0.263}$
(Appendix~\ref{app:scaling-fit}).

\scalingfigure
\vspace{5pt}

\subsection{Model and Training Budget Selection}
For final training, we consider larger budgets and two model-size ranges:
below 12M parameters for a compact policy smaller than the Slippi-AI reference,
and 30M--75M for the larger policy. End-to-end throughput measurements across
A100, H100, and RTX PRO 6000 GPUs, with varying CPU allocations, provide the
hardware anchors for converting dollar budgets into processed targets. Among the
evaluated candidates, the fitted loss and throughput models select 10M on A100 and 75M on RTX PRO
6000 for planned budgets of \$100 and \$200. The projected training horizons are
74.88B and 24.98B targets. These budgets specify the intended continuations;
sweep and recovery runs incur additional cost. Appendix~\ref{app:throughput}
details the throughput sweeps, recipes, fits, and hardware allocation.

\subsection{Full-Corpus Pretraining}
We tune eight recipes at 10M and sixteen at 75M around the transferred settings,
using successive halving~\cite{jamieson2016}. The 10M search probes
$0.8\times$ and $1.25\times$ along individual axes and one joint increase; the
75M search uses $0.5\times$ and $2\times$ axial probes plus interaction settings.
All trials use BF16 compute, gradient clipping at 1.0, and auxiliary AdamW
betas of $(0.9,0.95)$. At each rung, trials are compared
after equal processed targets on the fixed 4.34M-target validation manifest,
and the lowest-loss half continue with their optimizer and scheduler states.
The five rungs use approximately 2.54B additional targets each at 10M and 0.493B
at 75M, with respective trial counts of $8,4,2,1,1$ and $16,8,4,2,1$
(Figure~\ref{fig:halving}; full results in Appendix~\ref{app:halving-ledger}).
\par
\halvingfigure
\vspace{5pt}

At 75M, halving selects a Muon rate of 0.005, half the transferred center.
At 10M, reducing either weight decay or Muon learning rate to $0.8\times$
gives the two finalists. The lower-decay recipe wins rung 3 by just
0.000026 NLL, comparable to the variation observed when reevaluating identical
weights, then regresses during continuation. This near tie motivates retaining
both 10M finalists for recovery. After the initial continuations diverge, we
restart both finalists and the selected 75M recipe from initialization, saving
immutable checkpoints and validating every 0.5B targets.
\par
The choice between the 10M recipes remains close late into recovery: their
ranking alternates, with losses within 0.0005 NLL through 8B targets. The
selection rule uses minimum validation loss, with stability breaking near ties.
Both recipes reach their minimum at 8.00B targets, where the lower-Muon-rate
run achieves 0.796817 NLL versus 0.797050 for lower weight decay. It also
remains near 0.80 through 10.15B targets, whereas the lower-decay run diverges
between 8.5B and 9B. We therefore select the lower Muon learning rate on both
loss and stability.
\par
The selected 10M checkpoint reaches 0.7968 validation NLL and 0.7977 test NLL
at 8.00B targets; the 75M reaches 0.7648 and 0.7654 at 5.64B targets
(Figure~\ref{fig:pretraining-trajectories}; Appendix~\ref{app:recovery}). At these exposures, calibration-law
predictions are lower by 0.0370 and 0.0360 NLL. Both the data distributions and
progress through learning-rate decay differ from calibration. Longer
continuations diverge while learning rates remain near their peaks. The selected
checkpoints initialize a lower-rate supervised curriculum, followed by knowledge
distillation and reinforcement learning.
\par
\pretrainingfigure
}

\begin{document}
\fontsize{10}{11.5}\selectfont
\setlength{\parskip}{3pt}
\setlength{\abovedisplayskip}{5pt}
\setlength{\belowdisplayskip}{5pt}
\clubpenalty=10000
\widowpenalty=10000
\raggedbottom
\titleblock
\begingroup
\renewcommand{\thefootnote}{}
\footnotetext{\textsuperscript{$\dagger$}Advisory contributions.\quad Contact: Ali Janati, \href{mailto:aj3238@columbia.edu}{\nolinkurl{aj3238@columbia.edu}}; \href{mailto:rohit@frisson-labs.com}{\nolinkurl{rohit@frisson-labs.com}}.}
\endgroup
\section{Introduction and Related Work}
\introductiontext
\taskandarchitecturetext
\pretrainingtext
\section{Post-Training Methods}
\label{sec:posttraining}
Supervised post-training concentrates imitation on high-ranked winning play
and transfers supervision from the 75M policy to the 10M. Both models complete
a curriculum and a second round on a fixed winner mixture; the 10M then
completes two distillation rounds before both models proceed to RL. Validation
trajectories and gradient norms guide the choice of data mixture and learning-rate
schedule, and whether to continue training (Figure~\ref{fig:posttraining-trajectories};
Appendix~\ref{app:posttraining-diagnostics}).

\subsection{Curriculum and Checkpoint Selection}
Player rank identifies the demonstrator's skill bracket, while match outcome
selects the winning perspective. We combine these labels to select
demonstrations. Fine-tuning on higher-rated winning replays was also used in
AlphaStar~\cite{vinyals2019}. The labels come from the raw replays, and all
training uses the existing split (Appendix~\ref{app:curriculum}).
\par
Overall NLL averages over all held-out demonstrations. As training emphasizes
higher-ranked winners, improvements on their actions can be obscured by
regressions elsewhere in the validation set. We therefore rank checkpoints using
\begin{equation}
W=0.9\,L_{\mathrm{Master,won}}+0.1\,L_{\mathrm{Diamond,won}},
\label{eq:posttraining-selection}
\end{equation}
where each $L$ is mean controller NLL in nats per target on a fixed validation
slice. The 90\% weight prioritizes Master winners; the 10\% component keeps
Diamond-winning demonstrations in the selection criterion. Overall NLL tracks
regressions on the broader replay distribution, while per-head entropy monitors
the concentration of predicted actions. Gameplay evaluation follows in
Section~\ref{sec:benchmark}.
\par
The instabilities during pretraining motivate resuming at lower learning
rates. We retain cosine decay and begin without warmup, informed by work on
continued pretraining~\cite{parmar2024}. Sampling proceeds from the natural
distribution to winners of any rank, then Master and Diamond winners, and
finally Master winners. About 90\% of each model's overall NLL reduction in
this round occurs during the initial natural-data stage. Later filtering
increasingly favors Master-winner prediction. During the final Master-only
stage, $W$ continues falling while
overall and Diamond-winner NLL rise. All six rank/outcome slices still finish
below their pretrained losses. The sampling procedure, stage budgets, and
evaluations are given in Appendices~\ref{app:curriculum}
and~\ref{app:posttraining-schedules}.

\subsection{Training on the Target Mixture}
The regression on Diamond-winning demonstrations motivates training on
approximately 90\% Master-winner and 10\% Diamond-winner replay visits,
matching the emphasis of $W$. Gains on the target slices also slow as the
cosine rate falls, so we adopt a warmup-stable-decay (WSD)
schedule~\cite{hu2024}. A 2\% warmup restores the
initial post-training rate, which is held until 80\% of the budget; cosine
decay occupies the final 20\%. This gives the target mixture more exposure
before annealing. Shrinking backbone weight norms also motivate scaling
weight decay with the learning-rate multiplier
(Appendix~\ref{app:posttraining-schedules}).
\par
After 1.2B additional targets at 10M and 1.5B at 75M, $W$ falls from 0.7708
to 0.7644 and from 0.7359 to 0.7249, respectively. Both models initially
regress after warmup and improve during the final decay. Over the final 150M targets of the constant-rate phase, the 10M
levels off, while the 75M still improves by roughly
0.0003 per 50M-target interval (Appendix~\ref{app:posttraining-ledgers}). We try another WSD cycle for the 75M, but its
loss stays above the source checkpoint and then rises alongside the gradient
norms. We stop at 449M targets, with $W=0.7443$, and retain the preceding
75M checkpoint (Appendix~\ref{app:posttraining-stopping}).

\par\penalty0\vskip5pt
\begin{figure}[!htbp]
\centering
\includegraphics[width=\textwidth]{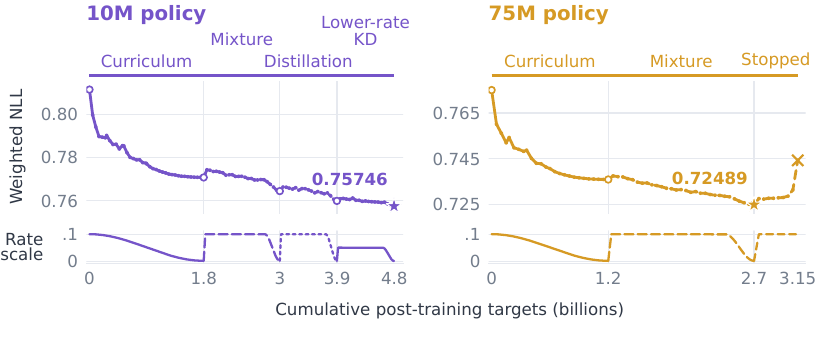}\par
\caption{\textbf{Post-training validation loss and learning-rate
schedules.} Open circles mark each round's source evaluation; stars mark retained
checkpoints. The 10M continues through two distillation rounds. The dashed 75M
continuation is stopped at the crossed endpoint. Rates are
normalized to the reference learning rates in Appendix~\ref{app:posttraining-schedules}.}
\label{fig:posttraining-trajectories}
\end{figure}
\par

\subsection{Distillation and the Final 10M Checkpoint}
We respond to the 10M plateau by adding supervision from the 75M,
which reaches substantially lower loss on the same target slices. We freeze it and
combine recorded-action NLL with its action distributions, following knowledge
and policy distillation~\cite{hinton2015,rusu2016}:
\begin{equation}
\mathcal L_{\mathrm{KD}}=(1-\alpha)\mathcal L_{\mathrm{BC}}
+\alpha T^2\sum_{c\in\{b,m\}}
\operatorname{KL}\!\left(p^{(T)}_{\mathrm{teacher},c}
\,\Vert\,p^{(T)}_{\mathrm{student},c}\right).
\label{eq:posttraining-kd}
\end{equation}
Here $p^{(T)}=\operatorname{softmax}(z/T)$, and each head's KL is averaged
over valid targets. We use teacher weight $\alpha=0.5$ and temperature $T=1$,
giving equal objective weights to recorded actions and the teacher. The teacher
supplies a distribution over commands at every frame, including alternatives
to the recorded action. On the same
winner mixture, 900M targets reduce $W$ from 0.7644 to 0.7598. This improvement
after the preceding supervised plateau motivates a second distillation round.
\par
In the first distillation round, $W$ is above its starting value at the first
evaluation after warmup and first falls below it at 549M targets. We halve
the next held rate and train for another 900M targets,
reaching $W=0.75746$. The gain per distillation round falls from 0.0046 to
0.0024, motivating the end of this supervised sequence. Shorter controls
slightly favor $\alpha=0.7$ at evaluations with the same constant learning
rate, while $T=2$ worsens validation loss. The second 900M distillation round,
with $\alpha=0.5$ and $T=1$, has the lowest $W$ among the 10M trials and
supplies the final checkpoint. Appendix~\ref{app:distillation}
gives the control trajectories and their schedules.
\par
Every selected round also improves weighted and overall test NLL. The final
overall test losses are 0.76974 for 10M and 0.74206 for 75M
(Appendix~\ref{app:posttraining-checkpoints}). These checkpoints are the
supervised endpoints used in the subsequent gameplay and training-stage
comparisons.

\subsection{Reinforcement Learning}
\label{sec:reinforcement-learning}
\label{sec:posttraining-rl}
We continue both supervised policies with proximal policy optimization
(PPO)~\cite{schulman2017}, learning from the consequences of their actions in the game.
Following Slippi-AI~\cite{slippiai}, we combine self-play with a penalty toward the
supervised starting policy, $\pi_{\mathrm{ref}}$. AlphaStar likewise used a KL loss
toward its supervised policy during RL~\cite{vinyals2019}. All training games are
Fox mirrors on the six stages of the expanded benchmark, with zero added policy delay.
They have no stock limit or timer, so play continues after each lost stock. The other
25 characters share the updated weights but receive no RL experience.
\par
\textbf{Reward and objective.} We use Slippi-AI's zero-sum reward. For player $p$
and opponent $q$, the transition from frame $t$ to $t+1$ gives
\begin{equation}
u^p_t=-k^p_t-0.01\delta^p_t-0.02g^p_t-\tfrac{0.1}{60}o^p_t+0.001c^p_t,
\qquad R^p_t=u^p_t-u^q_t.
\label{eq:rl-reward}
\end{equation}
Here $k^p_t$ indicates a lost stock, $\delta^p_t$ is damage received in percentage
points, $g^p_t$ penalizes grabbing the ledge when a vulnerable opponent is on its
stage side, $o^p_t$ penalizes spending time far offstage, and $c^p_t$ rewards
movement toward the opponent. Appendix~\ref{app:rl} gives the exact shaping
conditions. Taking a stock earns $+1$ and dealing one percentage point of damage
$+0.01$; the opponent's shaping terms enter with the opposite sign.
\par
For a command $a_{t+1}$ sampled after history $H_t$, the policy loss is
\begin{equation}
\begin{split}
\ell_t(\theta)={}&-w_{\mathrm{PG}}\min\bigl(\rho_t\hat A_t,\bar\rho_t\hat A_t\bigr)
+\beta\,\mathrm{KL}\bigl(\pi_{\mathrm{old}}\,\|\,\pi_\theta\bigr)\\
&+\lambda_{\to}\,\mathrm{KL}\bigl(\pi_\theta\,\|\,\pi_{\mathrm{ref}}\bigr)
+\lambda_{\leftarrow}\,\mathrm{KL}\bigl(\pi_{\mathrm{ref}}\,\|\,\pi_\theta\bigr),
\end{split}
\label{eq:rl-objective}
\end{equation}
where $\pi_{\mathrm{old}}$ collected the rollout and
$\rho_t=\pi_\theta(a_{t+1}\mid H_t)/\pi_{\mathrm{old}}(a_{t+1}\mid H_t)$.
Following Slippi-AI, clipping acts in log space:
$\bar\rho_t=\exp[\mathrm{clip}(\log\rho_t,-\epsilon,\epsilon)]$, with
$\epsilon=0.01$. We use unnormalized advantages from bootstrapped returns,
\begin{equation}
\hat A_t=G_t-V_\phi(H_t),\qquad G_t=R_t+\gamma G_{t+1},\qquad
\gamma=2^{-1/(60t_{1/2})},
\label{eq:rl-returns}
\end{equation}
where $t_{1/2}$ is the reward half-life in seconds and a separate value network
$V_\phi$ supplies the bootstrap at each rollout's end.
\par
The KL terms sum the two controller heads of Equation~\ref{eq:controller-factorization};
the main-stick term uses the joint-controller category sampled by $\pi_{\mathrm{old}}$.
The reference also reuses the acting policy's stored attention cache. Its weights are
frozen, but this cache reuse approximates the reference distribution, so the penalty
and reported reference KL values carry this approximation
(Appendix~\ref{app:rl}).
\par
\begin{figure}[!htbp]
\centering
\includegraphics[width=\textwidth]{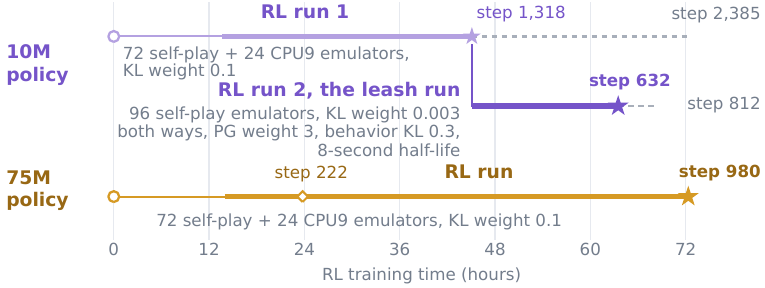}
\caption{\textbf{RL training and evaluated checkpoints.} Stars mark the evaluated
10M and 75M policies; the smaller star marks the 10M checkpoint from which the
leash run continues. The diamond is the earlier 75M checkpoint used in
Section~\ref{sec:behavior}. Dashed segments continue past selected checkpoints;
thin segments precede the implementation speedup described in Appendix~\ref{app:rl}.}
\label{fig:rl-lineage}
\end{figure}

\textbf{Training runs.} Each run uses one NVIDIA L4 GPU and 96 Dolphin
emulators~\cite{dolphin} on 32 CPU cores, driven through libmelee~\cite{libmelee}.
An RL step collects 2,048 frames per emulator and applies two PPO updates.
In the first run for each size, 72 emulators run self-play with both fighters
controlled by the policy, and 24 use the built-in level-9 opponent (CPU9).
Self-play supplies 86\% of decisions. The implementation averages the two groups'
trajectory losses equally, however, giving each group half the policy loss;
a CPU9 decision therefore receives six times the weight of a self-play decision.
The 10M continues in a second, self-play-only run, called the leash run in the
benchmark records. Table~\ref{tab:rl-settings} gives the settings and
Figure~\ref{fig:rl-lineage} the checkpoint history.
\begin{table}[!htbp]
\centering
\small
\begin{tabular}{@{}lccc@{}}
\toprule
Setting & 75M & 10M, run 1 & 10M, leash run \\
\midrule
Starting policy & supervised 75M & supervised 10M & run 1, step 1,318 \\
Self-play share of policy loss & 50\% & 50\% & 100\% \\
Learning rate, policy and value & $4\times10^{-5}$ & $10^{-4}$ & $10^{-4}$ \\
Policy-gradient weight $w_{\mathrm{PG}}$ & 1 & 1 & 3 \\
Reference KL weight $\lambda_{\to}$ & 0.1 & 0.1 & 0.003 \\
Reference KL weight $\lambda_{\leftarrow}$ & 0 & 0 & 0.003 \\
Behavior KL weight $\beta$ & 0 & 0 & 0.3 \\
Reward half-life (s) & 4 & 4 & 8 \\
\midrule
RL steps used / trained & 980 / 980 & 1,318 / 2,385 & 632 / 812 \\
Training time to step used (h) & 72.3 & 45.1 & 18.4 \\
\bottomrule
\end{tabular}
\caption{\textbf{RL settings and checkpoint budgets.} Each run retains its model's
supervised endpoint as the reference policy. The leash run inherits the value
network from step 1,318 and resets both optimizers. Training time is summed over
launches and includes in-loop evaluation; the selected 10M lineage totals 63.5 hours.
Common settings and implementation details are in Appendix~\ref{app:rl}.}
\label{tab:rl-settings}
\end{table}
\par
\textbf{Continuation and checkpoint selection.} About every three hours, an in-loop
evaluation runs 25.6 seconds of play on each of 30 emulators against CPU9,
MIMIC's Master Fox~\cite{mimic}, and a fixed earlier checkpoint. The 75M uses its
final checkpoint, step 980. For the 10M, step 1,318 maximizes in-loop reward per
frame against MIMIC in the first run. Its stock margin against MIMIC does not exceed
1.1 per minute, and separate Fox-mirror matches yield 6/34 wins against Slippi-AI's
Master release, 0/34 against Grandmaster, and 4/34 against the 21-frame
Cody-conditioned Fox specialist. Continued training to step 2,385 gives similar
results within sampling uncertainty (Appendix~\ref{app:rl}).
\par
\begin{figure}[!htbp]
\centering
\includegraphics[width=\textwidth]{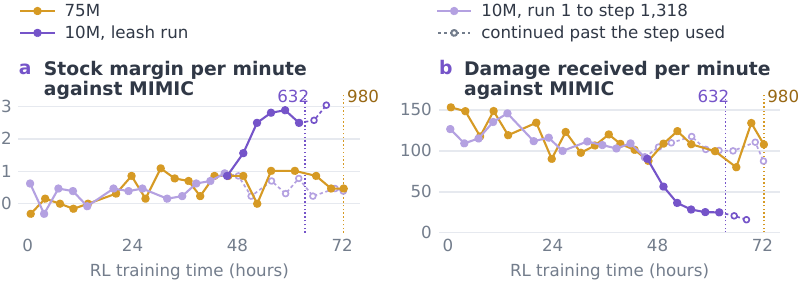}
\caption{\textbf{In-loop evaluation against MIMIC during RL.} Each round contains
25.6 seconds on each of 30 emulators. Stock margin is stocks taken minus stocks
lost per minute. Open markers follow the checkpoints used; dotted lines mark
75M step 980 and 10M step 632, which falls between evaluation rounds at steps
583 and 706. The 10M continuation starts 45 hours into its first run.}
\label{fig:rl-inloop}
\end{figure}

These results motivate the second 10M run. It lowers the reference KL weight,
adds the reverse-direction and behavior KL terms, increases the policy-gradient
weight and reward horizon, removes CPU9, and resets the optimizers. Its settings
follow Slippi-AI's released agents~\cite{slippiai}. Within about seven hours,
stock margin against MIMIC rises from 0.9 to 2.5 per minute; within about eleven
hours, damage received falls by more than two thirds (Figure~\ref{fig:rl-inloop}).
Two sibling runs use reference KL weights of 0.01 and 0.03; all three more than
halve damage received over this period. Several settings changed together, so
these comparisons do not isolate the cause of the improvement.
\par
We choose the 0.003 run after comparing the three runs against Slippi-AI's Master,
Diamond, and Grandmaster releases and an 18-frame Fox specialist; no stock-margin
difference between the 0.003 and 0.01 runs exceeds its sampling uncertainty.
Step 632 is the newest checkpoint at the time of selection. Several of these
opponents also appear in Section~\ref{sec:benchmark}, so results against them
are not independent of selection. The two model sizes also have different RL
histories: the selected 10M follows 1,950 steps across two runs, versus 980
at 75M, with different learning rates and objectives. Each configuration was
run once. Appendix~\ref{app:rl} gives the selection results, their uncertainty,
and the full training budgets.

\section{Benchmark Definition and Main Results}
\label{sec:benchmark}
We first compare pretrained and supervised post-trained Faynt policies with
existing agents in a 152-game suite. After reinforcement learning, we expand
the opponent set to fourteen releases, covering character specialists and
multi-character policies at different skill levels. For each Faynt checkpoint,
the same weights control all 26 characters. We examine both full-roster
performance and matchups against specialists playing their native fighters.

\subsection{Initial Benchmark Suite}
The Faynt schedule contains 10 Fox mirrors against MIMIC, 10 against the
built-in level-9 controller (CPU9), 12 MIMIC core mirrors, 24 Slippi-AI core
mirrors, and 96 roster-extension games. A mirror match uses the same character
on both sides. MIMIC supplies character-specific
policies~\cite{mimic}; the Slippi-AI reference uses \texttt{medium-v2} on its
twelve-character core, with Donkey Kong and Dr. Mario specialists in the
extension~\cite{slippiai}. Outside an opponent's deployed roster, the schedule
uses a fixed fallback checkpoint. Appendix~\ref{app:benchmark-original}
defines the character sets and every block allocation.
\par
We evaluate both Faynt sizes after pretraining and at the supervised
endpoints described in Section~\ref{sec:posttraining}, before RL. The four
checkpoints (Table~\ref{tab:benchmark-checkpoints}) share the same opponents,
character assignments, policy seeds, and ports. We also apply the suite design
to MIMIC and Slippi-AI, substituting the supervised 75M Faynt policy for each
baseline's own-family opponent blocks. Shared games appear in both policies'
views. Games use Final Destination, four stocks, and an eight-minute timer.
Each agent retains its observation interface, action decoder, and configured
delay; emulation waits for inference before advancing. Full execution and
reuse rules are given in Appendix~\ref{app:benchmark-runtime}.

\par\penalty0\vskip5pt
\begin{figure}[!htbp]
\centering
\includegraphics[width=\textwidth]{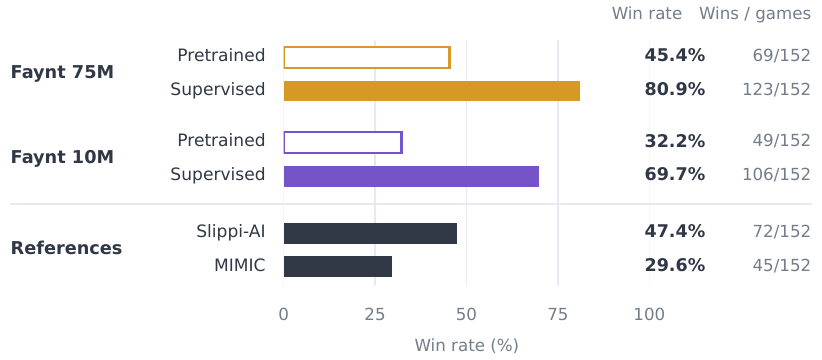}\par
\caption{\textbf{Initial benchmark results before reinforcement learning.}
Bars show win rates and win counts in each policy's 152-game evaluation.
Baseline schedules substitute
the supervised 75M Faynt comparator, so opponent composition differs between
Faynt and the reference views. Slippi-AI includes \texttt{medium-v2} and the
scheduled specialists. All reported outcomes have completed the validation
checks in Appendix~\ref{app:benchmark-runtime}.}
\label{fig:benchmark-initial}
\end{figure}
\par
Supervised post-training raises the 75M policy's win rate from 45.4\% to
80.9\%, and the 10M's from 32.2\% to 69.7\%
(Figure~\ref{fig:benchmark-initial}). Both improve in every suite block. The
reference views score 47.4\% for Slippi-AI and 29.6\% for MIMIC. The
post-trained 10M also exceeds the pretrained 75M's win rate despite
higher overall test NLL, 0.76974 versus 0.76543
(Appendix~\ref{app:posttraining-checkpoints}). This comparison motivates evaluating playing strength
alongside controller-prediction loss.
Section~\ref{sec:behavior} examines training-stage changes;
Appendix~\ref{app:benchmark-ledgers} gives the complete win/loss and stock
records for the pretrained and supervised checkpoints.
Direct pairwise comparisons on the original common native roster are given
in Appendix~\ref{app:benchmark-historical}.

\subsection{Final Benchmark After Reinforcement Learning}
We then evaluate the 75M policy at RL step 980 and the 10M policy at step
632 of its second RL run against fourteen frozen Slippi-AI releases~\cite{slippiai}.
Some releases also informed RL run and checkpoint selection
(Section~\ref{sec:reinforcement-learning}). The opponent set comprises five Fox or Falco specialists,
seven twelve-character policies spanning Silver through Grandmaster releases,
and two earlier four-character policies. Every Faynt checkpoint controls all
26 characters with the same weights. Opponent identities, conditioning, and
deployed rosters are fixed in Appendix~\ref{app:benchmark-expanded}.
\par
For each opponent, we define three conditions. In \emph{supported mirrors},
both agents play the same character from the opponent's deployed roster.
In \emph{extended-roster} games, Faynt uses a character outside that roster
while the opponent keeps a supported fighter. In \emph{forced mirrors}, both
play the same character outside the opponent's deployed roster. Supported
and forced mirrors together cover all 26 characters. Extended-roster games
test Faynt's additional fighters against opponents on supported characters.
\par
Specialists receive twelve supported mirrors across six stages and both ports;
multi-character policies receive two per supported fighter. Each additional
fighter receives two forced mirrors and two extended-roster games, reversing
ports on the assigned stage. The final suite contains \textbf{1,312 games per
RL checkpoint}, giving 2,624 distinct games overall
(Appendix~\ref{app:benchmark-series}).

\par\penalty0\vskip5pt
\begin{figure}[!htbp]
\centering
\includegraphics[width=\textwidth]{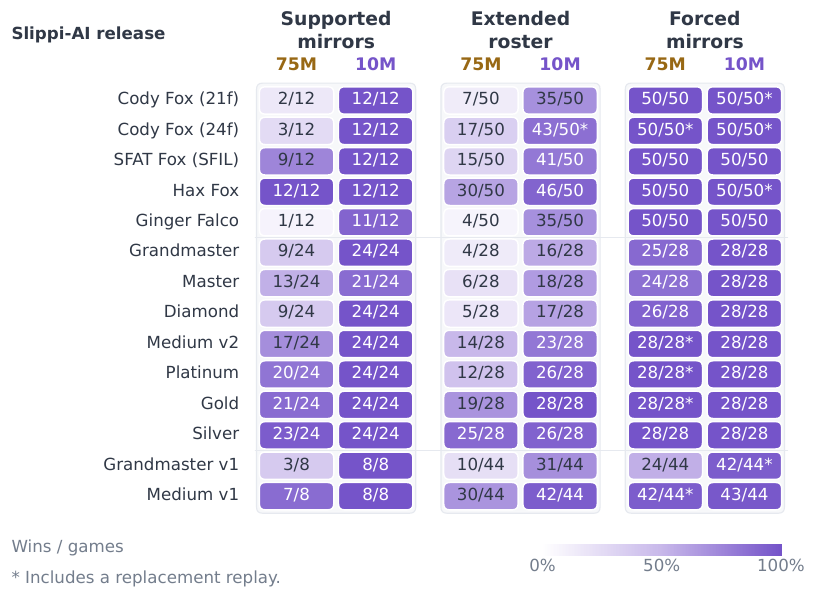}\par
\caption{\textbf{Expanded evaluation after reinforcement learning.}
Cells show Faynt wins/games; color encodes win rate. Row groups contain five
single-character specialists, seven twelve-character releases, and two
four-character releases. Supported and forced mirrors jointly cover all 26
characters; extended-roster games keep the opponent on a supported fighter.
Column pairs compare 75M RL step 980 and 10M RL step 632. Asterisks
mark cells containing separately rerun games
(Appendix~\ref{app:benchmark-audit}).}
\label{fig:benchmark-expanded}
\end{figure}
\par

\subsection{Supported-Character and Full-Roster Performance}
On opponents' supported characters, the 10M wins 240/244 mirrors (98.4\%)
and has a winning record against each of the fourteen releases; the 75M
wins 149/244. Each policy uses one checkpoint across all 26 characters.
In forced mirrors, the 75M and 10M win 503/534 and 531/534 games,
respectively. Combining supported and forced mirrors gives both models
winning full-roster records against all fourteen releases. These totals
include characters outside the opponents' deployed rosters;
Appendix~\ref{app:benchmark-full-roster} gives the per-release results and
their character weighting.
\par
Results on supported characters vary with specialization and
skill level (Figure~\ref{fig:benchmark-expanded}). Against the Hax-conditioned
Fox specialist, the 75M and 10M each win 12/12 mirrors; against SFAT Fox
(SFIL), they win 9/12 and 12/12. Against the 21-frame Cody-conditioned Fox
specialist, the 75M wins 2/12 and the 10M wins 12/12; against the
Ginger-conditioned Falco specialist, they win 1/12 and 11/12. The
twelve-character Grandmaster release retains an advantage over the 75M on
its supported roster, with Faynt winning 9/24 games at 75M and 24/24 at 10M.
\par
Both policies win games against these specialists using other characters.
Against Ginger Falco, the 75M and 10M win 4/50 and 35/50 games with
non-Falco fighters. The 10M's wins include both games as Ice Climbers and
both as Fox. Against the 21-frame Cody-conditioned Fox specialist, the 75M
wins 7/50 games using other fighters: both games as Roy and one of two with
each of Ganondorf, Ice Climbers, Luigi, Peach, and Yoshi. In all these games,
the specialist retains its native fighter. The per-character samples contain
two games; complete condition-level results are in
Appendices~\ref{app:benchmark-expanded-results}--\ref{app:benchmark-forced-results}.
We open-source both the initial
and final benchmark suites, together with the model weights and tournament platform.
\par
\Needspace{8\baselineskip}
\textbf{Timing differences.}
The Slippi-AI checkpoints in the preceding suites queue predicted actions for
18, 21, or 24 game frames~\cite{slippiai}. The supervised HAL comparison uses zero added execution delay and
executes two planned actions before replanning every two frames~\cite{hal}. MIMIC~\cite{mimic}
and Faynt predict a new command each frame without added policy delay.
These timing differences persist when emulation waits for inference
(Appendix~\ref{app:benchmark-runtime}). We have not isolated how much reaction
delay and decision frequency contribute to the reported performance differences.

\subsection{Zero-Delay Comparisons}
\label{sec:benchmark-zero-delay}
We also evaluate both RL policies against seven zero-delay Phillip
specialists~\cite{firoiu2017}, using each opponent's specialist character
on both sides. For each opponent, we play 16 games on Final Destination
and a separate 18-game evaluation across six stages, with three games per
stage. At RL step 980, the 75M wins \textbf{108/112 games (96.4\%)} on
Final Destination and \textbf{126/126} across six stages; the 10M at
RL step 632 wins \textbf{112/112} and \textbf{126/126}, respectively.
Both sides use zero added policy delay in these evaluations.
Appendix~\ref{app:benchmark-zero-delay} gives the opponent-level results.
\par
The Slippi-AI developers~\cite{slippiai} also privately supplied a checkpoint
with zero added action delay, which we evaluate under its ``Master Player''
and ``Cody'' conditioning settings. For each setting, we repeat the 16-game
Final Destination and 18-game six-stage evaluations, giving 68 games per
Faynt checkpoint. The 10M at RL step 632 wins all 68 games, including 61
four-stock victories, in which it loses no stocks. The 75M at RL step 980
wins 58/68 games (85.3\%), comprising 33/34 against Master Player and 25/34
against Cody. Section~\ref{sec:behavior-private-slippi} compares these results
with the supervised checkpoints; Appendix~\ref{app:benchmark-private-slippi}
gives the complete results.

\subsection{Real-Time Inference}
\label{sec:benchmark-realtime}
Our benchmark games advance the emulator only after both agents have responded, so their
results do not depend on inference speed (Appendix~\ref{app:benchmark-runtime}). Live play runs at
60 frames per second. For a policy with no added delay, each decision must therefore fit within
16.67~ms of the arrival of its game state.
\par
On recorded-game inputs, our optimized decision loop takes an average of 5.2~ms for the 10M and
8.7~ms for the 75M on an NVIDIA T4 GPU. Their 99th-percentile latencies are 6.8 and 12.1~ms,
respectively. Both fit within the frame budget at this percentile, although 0.6\% of the 75M's
decisions exceed it. The loop builds the input features, runs one cached policy step, samples and
decodes the controller heads, and issues the command to a recording stub. These timings exclude
the emulator and communication with it. Timing runs use random weights; output comparisons below
use the evaluated RL checkpoints (Appendix~\ref{app:inference-setup}).
\par
We feed frames on a fixed 60~Hz schedule, whether or not the loop is ready. The \emph{real-time
factor} (RTF) is the mean processing time, from picking up a state to issuing its command, divided
by 16.67~ms. Latency is counted from the frame's scheduled arrival and includes any time spent
waiting for the loop. We say a run keeps pace when its RTF is below 1 and its 99th-percentile
latency is within one frame. Both optimized models meet these criteria on recorded-game inputs
on the T4, L4, and A10G GPUs (Appendix~\ref{app:inference-gpu}).
\par
\begin{figure}[!htbp]
  \centering
  \includegraphics[width=\textwidth]{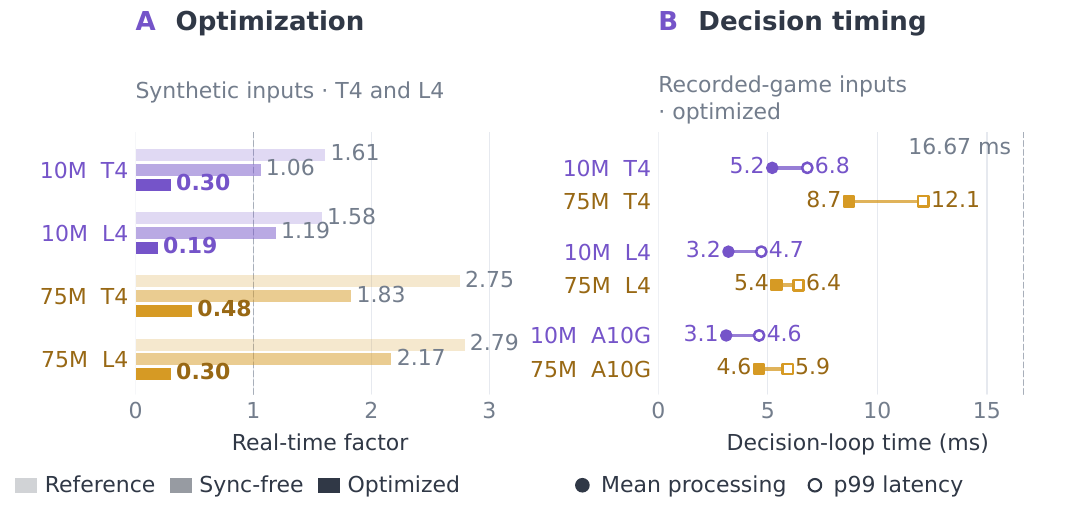}\par
  \caption{\textbf{Inference optimization and the 60~Hz decision budget.}
  Left: real-time factor on synthetic inputs, with darker bars showing successive optimizations.
  Right: mean processing time and 99th-percentile latency on recorded-game inputs, including
  feature conversion. Dashed lines mark one frame. All timings use random weights and exclude the
  emulator and communication with it; output parity is tested separately with trained checkpoints
  (Appendix~\ref{app:inference}).}
  \label{fig:realtime}
\end{figure}

The standard PyTorch~\cite{paszke2019} inference path, our reference implementation, misses this
target on every GPU tested (Figure~\ref{fig:realtime}). A single-frame step launches many small
operations from Python and repeatedly reads cache positions back from the GPU. We remove these
host synchronizations, keep cache bookkeeping on the device, and attend directly over the cache
in storage order. This \emph{sync-free step} computes the same formulas but still misses the
latency target on its own. Recording most of the step as a CUDA graph~\cite{cudagraphs} and
replaying it with one launch brings both models within the target, making the loop 5.4--9.2 times
faster on the T4 and L4 (Appendix~\ref{app:inference-optimized}).
With the evaluated RL checkpoints on a T4 and its host CPU, the reference and optimized
implementations produce action distributions that agree to within floating-point rounding.
Starting from the same random seed, they select the same actions throughout a 1,200-frame
recorded-game test (Appendix~\ref{app:inference-parity}).
\par
Among the cloud CPUs tested, only the optimized 10M on the AMD EPYC hosts keeps pace in the
median run, with an RTF of 0.68 on one thread. Its 99th-percentile latency varies from 14.0 to
18.6~ms across three shared hosts, leaving limited margin. The 10M also keeps pace on the laptop
CPU tested, while the 75M falls behind on every CPU in the study
(Appendix~\ref{app:inference-cpu}). Batched GPU runs serve up to 64 games with 99th-percentile
latency within the frame budget on synthetic inputs; deployment would also need to account for
each game's feature conversion and emulator costs (Appendix~\ref{app:inference-gpu}).
\par

\section{Behavior Across Training Stages}
\label{sec:behavior}
We first compare how the pretrained and supervised post-trained policies play:
how much damage they take and whether they turn early leads into wins.
Replay-based comparisons across these two stages use the same 145 match configurations at 75M
and 139 at 10M, selected from the initial 152-game suite after replay checks.
Appendix~\ref{app:behavior-comparability} describes replay selection and
evaluation conditions.

\subsection{Prediction Loss and Playing Strength}
The supervised 10M wins 69.7\% of the original 152-game suite (106/152),
compared with 45.4\% (69/152) for the pretrained 75M. Yet its overall
validation negative log-likelihood (NLL) is slightly higher: 0.76896 versus
0.76481. Test loss gives the same ordering
(Appendix~\ref{app:posttraining-checkpoints}).
\par
\begin{figure}[!htbp]
\centering
\includegraphics[width=\textwidth]{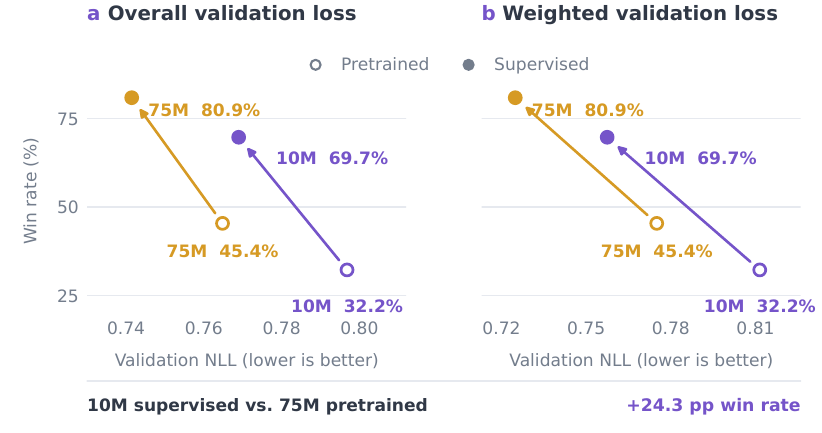}\par
\caption{\textbf{Prediction loss and win rate can favor different models.}
The four pretrained and supervised checkpoints are evaluated on fixed
validation data and the original 152-game suite. Open markers denote
pretrained models; filled markers denote supervised models. Arrows connect
checkpoints of the same size. The supervised 10M wins 37 more games than the
pretrained 75M despite higher overall loss. Their weighted losses follow
their ranking by win rate.}
\label{fig:loss-gameplay}
\end{figure}

Weighting validation loss toward Master- and Diamond-winning demonstrations
changes this ordering (Equation~\ref{eq:posttraining-selection}).
The supervised 10M has lower weighted loss than the pretrained
75M: 0.75746 versus 0.77502 (Figure~\ref{fig:loss-gameplay}). The ordering
by weighted loss agrees with the ordering by win rate for all four pretrained
and supervised checkpoints. The supervised checkpoints were selected using
$W$; Appendix~\ref{app:behavior-checkpoints} discusses the scope of this comparison.

\par\penalty0\vskip5pt

\subsection{Damage Exchange and Early Leads}
Melee displays damage as a percentage, which resets when a fighter loses a
life, called a \emph{stock}. We compute \emph{damage received per minute} by
summing positive changes in this counter between consecutive frames with
the same stock count and dividing by match duration in minutes. For example, a rise from 20\% to 35\% contributes 15 damage points.
The rate includes all damage sources, and each game has equal weight in the
reported average (Appendix~\ref{app:behavior-metrics}).
\par
\begin{figure}[!htbp]
\centering
\includegraphics[width=\textwidth]{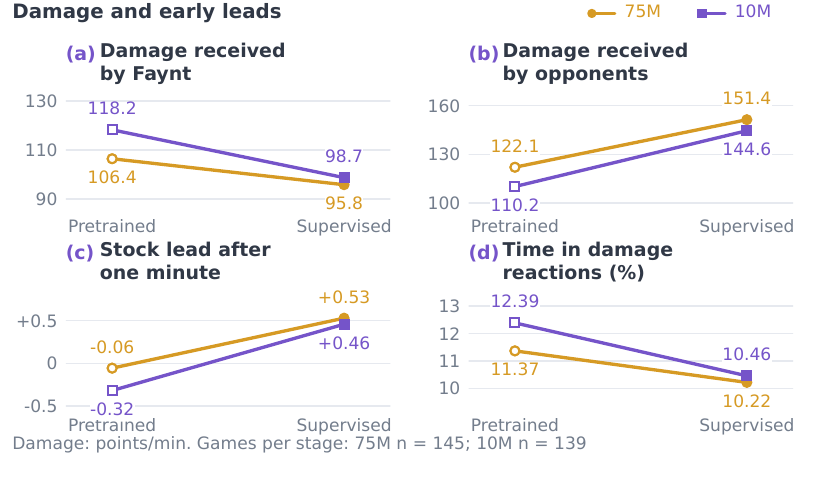}\par
\caption{\textbf{Faynt takes less damage and builds larger early leads.}
Each size uses the same match configurations across the two stages:
145 at 75M and 139 at 10M. Damage rates are percentage points received per
minute, averaged equally across games. Stock lead is the difference in
remaining lives after one minute; terminal scores are retained for games
that end earlier. The last panel shows the fraction of time Faynt spends
in damage-reaction states.}
\label{fig:behavior-trajectories}
\end{figure}

Both models take less damage per minute after supervised post-training,
while their opponents take more (Figure~\ref{fig:behavior-trajectories}).
For 75M, damage received falls from 106.4 after pretraining to 95.8 after
supervised training. Opponent damage rises from 122.1 to 151.4 over the
same stages.
\par
These changes are accompanied by stronger early leads. We define the
\emph{stock lead} as Faynt's remaining lives minus its opponent's. After
one minute, both pretrained policies trail slightly on average. Both supervised
policies lead on average. They also spend less time in damage-reaction states,
the game's action states following a hit. Figure~\ref{fig:behavior-trajectories}
shows the changes at each stage; Appendix~\ref{app:behavior-metrics} gives
the state definitions and treatment of games ending before one minute.

\par\penalty0\vskip5pt
\subsection{Winning from Early Leads and Deficits}
\begin{figure}[!htbp]
\centering
\includegraphics[width=\textwidth]{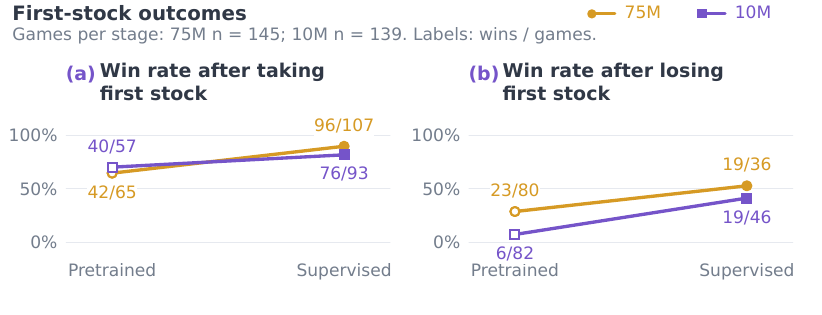}\par
\caption{\textbf{More wins from early leads and deficits.}
Win rates after taking or losing the first stock on the two-stage comparison
set. Labels give wins/games; simultaneous first-stock losses are excluded.}
\label{fig:behavior-outcomes}
\end{figure}

After taking the first stock, both pretrained policies still lose roughly
one game in three. After supervised post-training, the corresponding win
rates are 89.7\% for 75M and 81.7\% for 10M
(Figure~\ref{fig:behavior-outcomes}).
\par
Both policies also win a larger share of games after losing the first stock.
The change is especially large for 10M, whose win rate from this position
rises from 7.3\% after pretraining to 41.3\% after supervised training.
For 75M, the corresponding rates are 28.8\% and 52.8\%.
\par
Which games enter each group changes with the policy. The figure reports
the counts behind each rate. Appendix~\ref{app:behavior-trajectories}
also reports the count of games in which both players lose their first stock simultaneously.

\par\penalty0\vskip5pt

\subsection{Stock Preservation}
\begin{figure}[!htbp]
\centering
\includegraphics[width=0.96\textwidth]{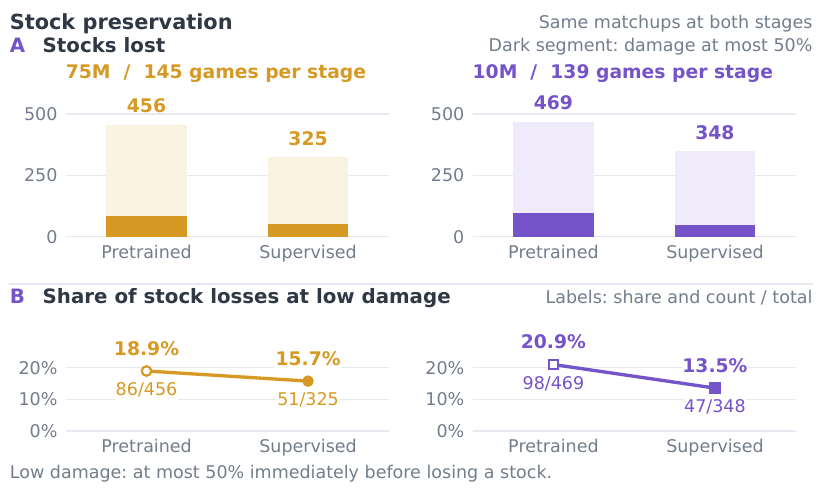}
\caption{\textbf{Stock preservation after supervised training.}
Top: total stock losses, highlighting those at 50\% damage or less.
Bottom: their share of all stocks lost, with counts. Both stages use the
same 145 configurations at 75M and 139 at 10M.}
\label{fig:behavior-preservation}
\end{figure}

After supervised post-training, the share of stock losses at 50\% damage
or less falls from 18.9\% to 15.7\% at 75M and from 20.9\% to 13.5\% at
10M (Figure~\ref{fig:behavior-preservation}). Appendix~\ref{app:behavior-metrics}
defines the metric; Appendix~\ref{app:behavior-stock-ledger} gives stock
margins for the full initial suite.

\subsection{From Supervised Post-Training to Reinforcement Learning}
\label{sec:behavior-private-slippi}
\begin{figure}[!htbp]
\centering
\includegraphics[width=\textwidth]{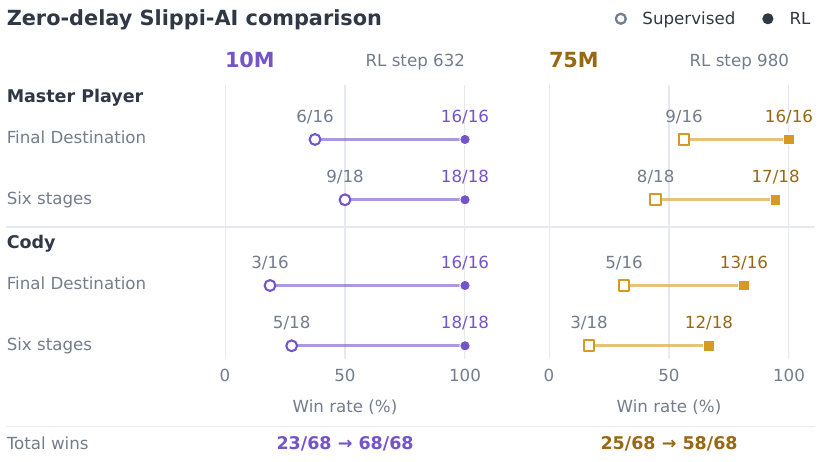}
\caption{\textbf{Playing strength improves after reinforcement learning.}
Open and filled markers show the supervised and RL checkpoints against one
privately supplied zero-delay Slippi-AI model under two conditioning settings.
Labels give wins/games: 16 games on Final Destination (FD) and 18 across six
stages per setting. The RL checkpoints are 10M step 632 and 75M step 980.
Appendix~\ref{app:benchmark-private-slippi} gives the full win/loss records.}
\label{fig:private-slippi-progression}
\end{figure}

Against the privately supplied zero-delay Slippi-AI checkpoint in
Section~\ref{sec:benchmark-zero-delay}, the 10M improves from 23/68 wins
(33.8\%) after supervised post-training to 68/68 (100\%) after RL, and the
75M from 25/68 (36.8\%) to 58/68 (85.3\%)
(Figure~\ref{fig:private-slippi-progression}).
Both models improve under each conditioning setting, on Final Destination
and across six stages. The 10M wins 61 of its 68 games without losing a stock.
For the 75M, the Cody setting remains more difficult, particularly across
six stages, where wins increase from 3/18 to 12/18.

\subsection{Performance Later in Reinforcement Learning}
\label{sec:behavior-rl-progression}
\begin{figure}[!htbp]
\centering
\includegraphics[width=\textwidth]{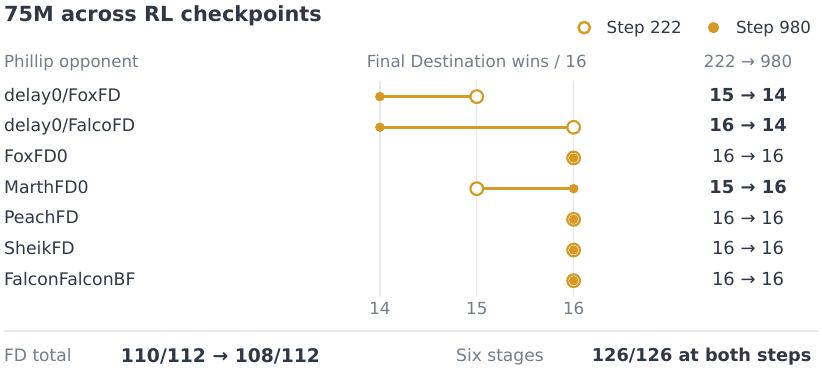}
\caption{\textbf{75M performance at two RL checkpoints.}
Markers show wins in 16 Final Destination games per opponent; labels give
step 222 and step 980 counts. Both checkpoints win 18/18 against every
opponent in the separate six-stage evaluation
(Appendix~\ref{app:benchmark-zero-delay}).}
\label{fig:rl-checkpoint-progression}
\end{figure}

We compare an earlier 75M checkpoint at RL step 222 with step 980 on the
zero-delay specialist evaluations of Section~\ref{sec:benchmark-zero-delay}
(Figure~\ref{fig:rl-checkpoint-progression}).
Final Destination wins change from 110/112 to 108/112, with one fewer win
against \texttt{delay0/FoxFD}, two fewer against \texttt{delay0/FalcoFD},
and one additional win against \texttt{MarthFD0}, while both checkpoints
win all 126 games across six stages.
Both checkpoints perform near ceiling on these evaluations, with small
differences in individual matchups on Final Destination.

\section{Conclusion}
\label{sec:conclusion}
We presented Faynt, a family of policies that controls all 26 Melee characters
using a single checkpoint at each model size. Our study follows their
development through model-size and optimization experiments, scaling
analysis, supervised post-training, distillation, and reinforcement learning
restricted to Fox mirror matches.
The supervised 10M wins 69.7\% of the initial benchmark, compared with
45.4\% for the pretrained 75M, despite slightly higher overall
controller-prediction loss. This result motivates evaluating playing strength
throughout training, alongside prediction loss.
\par
After supervised post-training, both models take less damage
per minute, build larger early leads, and win a larger share of games after
losing the first stock. After RL, the 10M wins 240/244 mirrors against
fourteen releases on their supported characters, with a winning record against
every release, including all five specialists; the 75M wins 149/244. In the additional
zero-delay evaluation against a privately supplied Slippi-AI checkpoint,
the 10M and 75M win 68/68 and 58/68 games across two conditioning settings.
Both have winning
full-roster mirror records against all fourteen releases when games outside
the opponents' deployed rosters are included. The expanded suite retains the
opponents' action delays; we have not isolated their contribution to these
results. The two model sizes also have different RL training histories.
\par
An optimized inference implementation preserves the policy's outputs to
within floating-point rounding. On recorded game states, the T4 decision
loop averages 5.2\,ms for the 10M and 8.7\,ms for the 75M, with
99th-percentile latencies below the 16.67\,ms frame budget. These timings
exclude emulator execution and communication
(Section~\ref{sec:benchmark-realtime}).
\par
We release the weights, both benchmark suites, and the tournament platform.
Several directions follow: varying observation and action delays,
language-based steering, transfer to other video games, and group relative
policy optimization (GRPO)~\cite{shao2024} over long-horizon trajectories.
Training with a selectable delay, as explored in speech processing and
simultaneous translation~\cite{kuzmin2026,elbayad2020}, could support
controlled comparisons across reaction delays within a single policy.
Performance drops
on some game stages also motivate stage-specific training and analysis,
which received limited attention in this study. Another direction is to test
whether resuming from the last stable checkpoint with the parameters trained
by auxiliary AdamW frozen stabilizes divergent training~\cite{janati2026}.

\FloatBarrier

\section*{Author contributions}
Ali Janati and Nikita Kuzmin were the primary contributors. Rohit Swamy
and Charles Niu contributed in advisory roles.

\par\medskip
\textbf{Acknowledgments.} We thank the Slippi-AI developers for the tools and
learning methods that informed this work and for privately supplying the
zero-delay evaluation checkpoint. We thank Project Slippi and the ranked
community for the replay infrastructure and original anonymized collections,
and Erick Martinez for preparing and hosting the \emph{Melee Ranked Replays}
redistribution.

\par\medskip
\textbf{Intended use.} Faynt is intended for offline research and controlled local evaluation.
Do not use or adapt its policies, checkpoints, or released code for Slippi
Online. The \href{https://github.com/project-slippi/slippi-launcher/blob/main/src/renderer/components/slippi_online_rules/slippi_online_rules.messages.ts}{\textcolor{resourcelink}{Slippi Online rules}}
prohibit macros and bots.

  \phantomsection
    \par\bigskip
  \phantomsection
  \addcontentsline{toc}{section}{Appendix}
  \section*{Appendix}

\FloatBarrier
\appendix
\counterwithin{table}{section}
\counterwithin{equation}{section}
\counterwithin{figure}{section}
\fontsize{10}{11.5}\selectfont
\setlength{\parskip}{4pt}
\setlength{\abovedisplayskip}{6pt}
\setlength{\belowdisplayskip}{6pt}
\titlespacing*{\section}{0pt}{14pt}{7pt}
\raggedbottom
\section{Dataset Construction and Sampling}
\label{app:data}

\subsection{Source and Deduplication}
The source snapshot contains 934 compressed archives (1,429.8\,GB) from
\emph{Melee Ranked Replays}~\cite{meleereplays}, selected across 25 character
buckets, six rank-pair buckets, and six archive groups. A mixed-character game
is assigned to its metadata owner's bucket and retained in full. Player ranks
are stored in anonymized netplay names, and final placements are read from
replay end blocks.
\par
SHA-256 deduplication first removes byte-identical files, then removes replays
with identical parsed training content. Complete games are validated before
acceptance. Parsing uses Slippi-AI~\cite{slippiai} commit
\texttt{577965a7731dc53e3472ea63d9e9853a4e9d65fa}.
One zstd-compressed MDS record contains a whole replay with both player
perspectives; shards contain approximately 15 replays. Table~\ref{tab:dataset-builds}
gives the final counts for each build.
\par
\begin{table}[htbp]
\centering\small
\begin{tabular}{@{}lrr@{}}
\toprule
 & Calibration & Final corpus \\
\midrule
Accepted replays & 67,129 & 839,942 \\
Training replays & 65,824 & 823,083 \\
Validation replays & 659 & 8,483 \\
Test replays & 646 & 8,376 \\
All valid targets & 1,500,000,244 & 17,839,509,998 \\
Training targets & 1,470,952,166 & 17,481,727,198 \\
Validation targets & 14,467,358 & 180,230,928 \\
Test targets & 14,580,720 & 177,551,872 \\
Dataset manifest & \texttt{30981903d542} & \texttt{e018460bb8b2} \\
\bottomrule
\end{tabular}
\caption{Dataset builds. A target is one valid transition from frame $t$ to
$t+1$ for one player perspective. Manifest identifiers are the abbreviated
values recorded in the study. The final split identifier is
\texttt{9c4ae826ae97}.}
\label{tab:dataset-builds}
\end{table}

\subsection{Game-Level Splits and Fixed Evaluation Windows}
The split seed is 15031. For replay identity $x$, we hash
\texttt{melee-policy.split.v1}, the seed, and $x$ with SHA-256, then interpret the first
eight digest bytes as an unsigned 64-bit value divided by $2^{64}$ to obtain
$u\in[0,1)$. Training receives $u<0.98$, validation receives
$0.98\leq u<0.99$, and test receives $u\geq0.99$. The calibration build uses
the raw file hash as replay identity; the final build uses the canonical
game-start identity. Assignment is deterministic and has no character
stratification. Both perspectives remain in the same split.
\par
Checkpoint evaluation on the final corpus uses one fixed 256-target window per
held-out perspective: 16,966 validation windows (4,343,296 targets) and
16,752 test windows (4,288,512 targets). These sampled evaluation manifests
are smaller than the complete held-out target populations in
Table~\ref{tab:dataset-builds}. Validation traverses its entire manifest in
stable order with seed 15031. The manifests are shared across pretraining
and post-training, and test evaluation follows checkpoint selection.

\subsection{Training Windows and Resumable Sampling}
The final runs use the \texttt{replay-local-v1} sampler: deterministic rotated
replay traversal, 32 hash-pseudorandom windows sampled with replacement per
replay visit, and a replay cache of size two. Each optimizer update uses
256 windows from eight replay identities, giving $256\times256=65{,}536$
targets. Windows contain consecutive state and current-controller inputs
with next-frame controller targets and remain within one game. Repeated
exposure is included in the processed-target count used by the scaling laws.
\par
Checkpoints store the sampler cursor and random-number-generator state
alongside the weights and optimizer state, allowing sampling to resume from
the same position.

\par
\section{Policy Implementation and Controller Codec}
\label{app:architecture}

\subsection{State Features and Frame Encoding}
Each player record contains character identity, action-state identity, damage
percentage, facing direction, $x$ and $y$ position, invulnerability, remaining
jumps, shield strength, and an on-ground flag. The controlled player and
opponent each have an optional Nana companion record with the same fields
and an existence flag. Character and action spaces contain 33 and 399
categories. Their learned embeddings are 128-dimensional and shared across
all four records. An additional 128-dimensional embedding with index
$399\,c+a$ is added to the action embedding for character $c$ and action $a$.
Remaining jumps use seven one-hot categories.
\par
World inputs comprise a 64-category stage encoding, the $x,y$ coordinates of
the moving platform on Yoshi's Story (Randall), the left and right platform
heights on Fountain of Dreams, and 15 item slots.
Each item contains 238 type categories, 13 state categories,
an existence flag, and $x,y$ coordinates. Its 254-dimensional representation
passes through a shared $254\rightarrow128\rightarrow128$ MLP with a ReLU
between layers. Outputs for absent items are masked before summation.
The encoder concatenates 1,082 player and companion features, 196 world
features, and 813 current-controller features into a 2,091-dimensional vector.
\par
Damage percentage and shield strength are scaled by 0.01; coordinates are
scaled by 0.05. Scaled numeric fields are clamped to $[-10,10]$. Facing is
encoded as $-1$ or $+1$, and flags as 0 or 1. Absent companion records have
zero raw fields and a false existence flag, while their categorical
embedding lookups are still performed. These conventions follow the
structured Slippi-AI representation~\cite{slippiai} with the learned
embeddings and item pooling described in Section~\ref{sec:task-data-architecture}.

\subsection{Shared Transformer Configuration}
Table~\ref{tab:model-config} gives the dimensions of the two final policies.
Both use PyTorch causal scaled dot-product attention, RoPE base 10,000,
RMSNorm $\epsilon=10^{-6}$, and attention scale $1/\sqrt{64}$. Backbone
and gate biases and dropout are disabled; the initialization standard
deviation is 0.02. Full Attention Residuals uses zero-initialized learned
queries.
\par
During imitation training, weights are FP32, while matrix operations and the
key-value cache use BF16. Normalization, the gate product, depth routing, and
softmax use FP32. Live CPU execution also uses FP32. The final launch enables prevalidated inputs
and disables activation checkpointing. Companion features, learned and joint
embeddings, pooled items, and platform features are enabled; the controller
RNN, hybrid embedding, and name conditioning are disabled.

\subsection{Exact Controller Categories}
The \texttt{custom\_v1} codec~\cite{slippiai} forms the joint-controller ID
in mixed-radix order from $B$, $X\lor Y$, $L\lor R$, the combined
$Z$/A/shoulder category, and the C-stick category, with the C-stick varying
fastest. Their cardinalities are $2\times2\times2\times7\times13=728$.
When Z is absent, the seven-way category encodes the Cartesian product of
A and three shoulder levels; Z takes the remaining category and has priority.
Shoulder values above 0.3 enter the light bucket and values above 0.9 enter
the full bucket. Decoding uses levels $0$, $0.35$, and $1$, maps $X\lor Y$
to Y and $L\lor R$ to L, and maps Z to Z, A, and shoulder level $0.35$.
\par
The C-stick has a center and two radial rings with 4 and 8 angles, giving
13 positions. The main stick has a center and three rings with 4, 16, and
64 angles, giving 85 positions. Nonzero radii are logarithmically spaced
from 23 to 80 in raw controller coordinates. Each ring's angles start at
$-\pi$ and are equally spaced; decoded raw coordinates $v$ map to controller
coordinates as $(v+80)/160$. Encoding rounds the input to raw coordinates,
assigns radii at most 22 to the center, and rounds nonzero radii and angles
to their respective bins.

\par
\section{Optimization and Pretraining Details}
\label{app:optimization}
\begingroup
\setlength{\parskip}{3pt}
\subsection{Optimizer Screens and Hyperparameter Transfer}
\label{app:optimizer-screens}
The optimizer screen uses the 20,040,877-parameter calibration policy: six
Transformer layers, width 512, eight attention heads, and 256-frame windows.
All trials share initialization, data, validation windows, and a 166M-target
endpoint. Table~\ref{tab:optimizer-screens} gives all thirteen endpoints.
At batch size 65,536, each microbatch contains eight 256-frame windows,
accumulated over 32 steps. AdamW also uses a 262,144-target batch in three
trials. The selected Muon configuration improves validation NLL by
0.014217 over the best AdamW configuration. These are single-seed measurements.

\begin{table}[!htbp]
\caption{Complete optimizer screen. Muon uses auxiliary AdamW at
0.001 and shared weight decay 0.01. N denotes Nesterov momentum; NS is the
number of Newton--Schulz steps. Bold entries are the selected endpoints.}
\label{tab:optimizer-screens}
\vspace{3pt}
\begin{minipage}[t]{.45\textwidth}\centering
\fontsize{9}{11}\selectfont\setlength{\tabcolsep}{4pt}
\begin{tabular}[t]{@{}rrr@{}}\toprule
\multicolumn{3}{c}{AdamW}\\
Learning rate & Batch & NLL\\\midrule
0.0010 & 65,536 & \textbf{0.860504}\\
0.0003 & 65,536 & 0.870815\\
0.0030 & 65,536 & 0.875388\\
0.0001 & 65,536 & 0.914326\\
0.0010 & 262,144 & 0.914411\\
0.0003 & 262,144 & 0.939331\\
0.0001 & 262,144 & 1.140582\\\bottomrule
\end{tabular}
\end{minipage}\hfill
\begin{minipage}[t]{.52\textwidth}\centering
\fontsize{9}{11}\selectfont\setlength{\tabcolsep}{3pt}
\begin{tabular}[t]{@{}rrccr@{}}\toprule
\multicolumn{5}{c}{Muon}\\
Rate & Momentum & N & NS & NLL\\\midrule
0.01 & 0.95 & yes & 5 & \textbf{0.846287}\\
0.02 & 0.90 & yes & 5 & 0.847126\\
0.02 & 0.95 & yes & 5 & 0.849624\\
0.02 & 0.95 & no & 5 & 0.851615\\
0.04 & 0.95 & yes & 5 & 0.852953\\
0.02 & 0.95 & yes & 3 & 0.856479\\\bottomrule
\end{tabular}
\end{minipage}
\end{table}

Muon~\cite{muon2024} updates the calibration backbone's 42 hidden matrices;
AdamW~\cite{adamw2019} updates the encoder, decoder, embeddings, normalization
parameters, biases, and remaining vectors. Each parameter belongs to one
optimizer. For a matrix of shape $r\times c$, the Muon update uses momentum
$\mu=0.95$, Nesterov, and the shape factor $\sqrt{\max(1,r/c)}$.
Its five-step BF16 Newton--Schulz iteration uses coefficients
$(3.4445,-4.7750,2.0315)$ and Frobenius-normalized input.

\begin{table}[!htbp]
\caption{Hyperparameter transfer at 41.94M targets per run. Each
row varies one setting from the selected 20M recipe. The baseline is shared
across the three rows at each size, yielding fourteen distinct runs.}
\label{tab:transfer-screen}
\centering\fontsize{9}{11}\selectfont\setlength{\tabcolsep}{8pt}
\begin{tabular}{@{}llrrr@{}}\toprule
Size & Setting varied & $0.5\times$ & $1\times$ & $2\times$\\\midrule
5M & Muon rate & 0.943590 & 0.941835 & \textbf{0.939627}\\
5M & Auxiliary rate & 0.950677 & 0.941835 & \textbf{0.926854}\\
5M & Weight decay & 0.939114 & 0.941835 & \textbf{0.937286}\\
50M & Muon rate & 0.915042 & \textbf{0.912638} & 0.927680\\
50M & Auxiliary rate & 0.920879 & \textbf{0.912638} & 0.914731\\
50M & Weight decay & 0.924369 & \textbf{0.912638} & 0.917862\\\bottomrule
\end{tabular}
\end{table}

The three 5M selections come from separate axiswise trials. They define the
search center through $q=\log 2/\log(20{,}040{,}877/5{,}053{,}805)
=0.5031437076$ and the shared multiplier in Section~\ref{sec:pretraining}.
The transferred $(\eta_\mu,\eta_{\rm aux},\lambda)$ are
$(0.025986,0.002599,0.025986)$ at 3M,
$(0.02,0.002,0.02)$ at 5M, and
$(0.014072165624,0.0014072165624,0.014072165624)$ at 10M.
For 20M, 50M, and 75M they remain $(0.01,0.001,0.01)$.

\par
\subsection{Compute Accounting and Scaling Fits}
\label{app:scaling-fit}
For profile $p$, compute is $C_p=D G_p$ with
$G_p=F_{\mathrm{model},p}+F_{\mathrm{NS},p}/B$ and $B=65{,}536$.
The count includes forward and backward passes and the dominant five-step
Newton--Schulz matrix multiplications. It excludes elementwise optimizer
operations, validation, checkpointing, and input-pipeline work.
Table~\ref{tab:scaling-ledger} lists the discrete costs and all twelve endpoints.
Integer update counts produce small deviations from the nominal compute budgets.

\begin{table}[!htbp]
\caption{Exact architecture costs and calibration endpoints.
$G_p$ is FLOPs per target. C1, C2, and C3 denote 0.0052, 0.0104, and
0.0208 EFLOP. The smallest measured model has the lowest loss in each group.}
\label{tab:scaling-ledger}
\centering\fontsize{9}{11}\selectfont\setlength{\tabcolsep}{7pt}
\begin{tabular}{@{}lrr@{}}\toprule
Profile & Parameters & $G_p$\\\midrule
3M & 3,003,373 & 12,847,328\\
5M & 5,053,805 & 29,304,944\\
10M & 10,163,629 & 72,262,720\\
20M & 20,040,877 & 151,385,600\\
50M & 50,127,021 & 394,013,760\\
75M & 75,305,709 & 596,823,632\\\bottomrule
\end{tabular}
\par\vspace{5pt}
\begin{tabular}{@{}llrrr@{}}\toprule
Group & Profile & Updates & Processed targets & Validation NLL\\\midrule
C1 & 3M & 6,176 & 404,750,336 & \textbf{0.839922}\\
C1 & 5M & 2,708 & 177,471,488 & 0.852395\\
C1 & 10M & 1,098 & 71,958,528 & 0.901357\\
C1 & 20M & 524 & 34,340,864 & 0.938103\\
\addlinespace[2pt]
C2 & 5M & 5,415 & 354,877,440 & \textbf{0.825642}\\
C2 & 10M & 2,196 & 143,917,056 & 0.853261\\
C2 & 20M & 1,048 & 68,681,728 & 0.886462\\
C2 & 50M & 403 & 26,411,008 & 0.964864\\
\addlinespace[2pt]
C3 & 10M & 4,392 & 287,834,112 & \textbf{0.822259}\\
C3 & 20M & 2,097 & 137,428,992 & 0.840928\\
C3 & 50M & 806 & 52,822,016 & 0.885656\\
C3 & 75M & 532 & 34,865,152 & 0.924414\\
\bottomrule\end{tabular}
\end{table}

We fit the parametric loss surface~\cite{hoffmann2022} by unweighted nonlinear
least squares in raw endpoint NLL. The fitted coefficients are
$L_\infty=0.643289120$, $A_N=0.101704664$, $A_D=0.127464887$,
$\alpha=0.202435790$, and $\beta=0.485913629$, using
$N_0=10^7$ and $D_0=10^8$. The in-sample RMSE is 0.004082 NLL and
$R^2=0.991643$. A log-space fit to the six discrete costs gives
$\widehat G(N)=61{,}654{,}448(N/10^7)^{1.167961}$ FLOPs per target.

Substituting $D=C/\widehat G(N)$ into the loss surface and minimizing over $N$
gives the compute optima by extrapolation. With $\gamma=1.167961$ and
$C_{\rm ref}=1.04\times10^{16}$ FLOP, the allocation exponents are
$\beta/(\alpha+\gamma\beta)=0.631086$ for parameters and
$\alpha/(\alpha+\gamma\beta)=0.262916$ for processed targets:
\begin{align*}
\widehat N_{\rm opt}(C)&=2{,}719{,}574(C/C_{\rm ref})^{0.631086},\\
\widehat D_{\rm opt}(C)&=771{,}880{,}565(C/C_{\rm ref})^{0.262916},\\
\widehat L_{\rm opt}(C)&=0.643289+0.179598(C/C_{\rm ref})^{-0.127754}.
\end{align*}
At C1, C2, and C3, the predicted losses at the optima are 0.839516,
0.822887, and 0.807667. The reported goodness of fit describes the twelve
calibration endpoints; the study contains one seed per scaling coordinate.

\par
\subsection{Full-Corpus Search Recipes and Halving Results}
\label{app:halving-ledger}
The 10M search uses a narrow neighborhood of the transferred center because
its size lies between the 5M and 20M anchors. The 75M search uses a broader
neighborhood because its center extends the plateau measured at 50M.
Table~\ref{tab:halving-recipes} defines every candidate and gives all scheduled
losses. Multipliers act on $(\eta_\mu,\eta_{\rm aux},\lambda)$ in that order.
The centers are
\[
(\eta_\mu,\eta_{\rm aux},\lambda)=
\begin{cases}
(0.014072165624,0.0014072165624,0.014072165624), & 10\mathrm{M},\\
(0.01,0.001,0.01), & 75\mathrm{M}.
\end{cases}
\]

\begin{table}[!htbp]
\caption{All 24 full-corpus recipes and 47 rung evaluations.
Empty cells follow elimination. For interaction recipes,
$\ell=1/\sqrt{2}$ and $h=\sqrt{2}$; their three-letter names give the
multipliers in order. The 75M local-fit recipe uses
$(0.77798125836,1.2289867927,1.1422710484)$.}
\label{tab:halving-recipes}
\centering\fontsize{8.5}{10.4}\selectfont\setlength{\tabcolsep}{3pt}
\begin{tabular}{@{}llrrrrr@{}}\toprule
Recipe & Multipliers & R1 & R2 & R3 & R4 & R5\\\midrule
\multicolumn{7}{@{}l}{\textbf{10M}}\\[2pt]
wd-low & $1,1,0.8$ & 0.81019 & 0.80431 & 0.801534 & 0.99334 & 0.94213\\
muon-low & $0.8,1,1$ & 0.81022 & 0.80387 & 0.801560 &  & \\
aux-low & $1,0.8,1$ & 0.81252 & 0.80572 &  &  & \\
center & $1,1,1$ & 0.81220 & 0.80635 &  &  & \\
aux-high & $1,1.25,1$ & 0.81294 &  &  &  & \\
wd-high & $1,1,1.25$ & 0.81467 &  &  &  & \\
muon-high & $1.25,1,1$ & 0.81501 &  &  &  & \\
joint-high & $1.25,1.25,1.25$ & 0.81811 &  &  &  & \\
\midrule\multicolumn{7}{@{}l}{\textbf{75M}}\\[2pt]
muon-low & $0.5,1,1$ & 0.81929 & 0.80129 & 0.78620 & 0.782058 & 0.785923\\
lll & $\ell,\ell,\ell$ & 0.81942 & 0.80167 & 0.78655 & 0.783126 & \\
lhl & $\ell,h,\ell$ & 0.81806 & 0.80171 & 0.78670 &  & \\
wd-low & $1,1,0.5$ & 0.81802 & 0.80153 & 0.78674 &  & \\
local fit & see caption & 0.81956 & 0.80435 &  &  & \\
lhh & $\ell,h,h$ & 0.81975 & 0.80461 &  &  & \\
aux-high & $1,2,1$ & 0.81907 & 0.80510 &  &  & \\
hhl & $h,h,\ell$ & 0.81950 & 0.80548 &  &  & \\
hll & $h,\ell,\ell$ & 0.82021 &  &  &  & \\
center & $1,1,1$ & 0.82046 &  &  &  & \\
aux-low & $1,0.5,1$ & 0.82136 &  &  &  & \\
llh & $\ell,\ell,h$ & 0.82182 &  &  &  & \\
hhh & $h,h,h$ & 0.82332 &  &  &  & \\
muon-high & $2,1,1$ & 0.82361 &  &  &  & \\
wd-high & $1,1,2$ & 0.82393 &  &  &  & \\
hlh & $h,\ell,h$ & 0.82437 &  &  &  & \\
\bottomrule\end{tabular}
\end{table}

The exploratory 75M local-fit candidate combines separate three-point local
responses from the 50M transfer screen. Each axis fit has zero residual degrees
of freedom, so this candidate serves as an additional search proposal.
Successive halving~\cite{jamieson2016} retains the lowest-loss half at each
boundary. Ties are resolved using earlier-rung losses and then distance from
the transferred center. Table~\ref{tab:rung-schedule} specifies the exact update
boundaries; multiplying by 65,536 recovers cumulative targets.

\begin{table}[!htbp]
\caption{Matched-target halving schedule. Active counts refer to the
trials evaluated at the end of each rung. The 10M winner continues alone in
rungs 4 and 5; the 75M winner continues alone in rung 5.}
\label{tab:rung-schedule}
\centering\fontsize{9}{11}\selectfont\setlength{\tabcolsep}{10pt}
\begin{tabular}{@{}crrrr@{}}\toprule
& \multicolumn{2}{c}{10M} & \multicolumn{2}{c}{75M}\\
Rung & Updates & Active & Updates & Active\\\midrule
1 & 38,730 & 8 & 7,526 & 16\\
2 & 77,461 & 4 & 15,052 & 8\\
3 & 116,192 & 2 & 22,579 & 4\\
4 & 154,923 & 1 & 30,105 & 2\\
5 & 193,654 & 1 & 37,631 & 1\\\bottomrule
\end{tabular}
\end{table}

\par
\subsection{Schedules, Recovery, and Checkpoint Selection}
\label{app:recovery}
All pretraining schedules use 1\% linear warmup followed by cosine decay to
10\% of the peak. For processed targets $f$ and the run's planned horizon $F$,
\[
\eta(f)=\eta_{\rm peak}
\begin{cases}
f/(0.01F), & f<0.01F,\\
0.1+0.45\!\left[1+\cos\!\left(\pi\dfrac{f-0.01F}{0.99F}\right)\right],
& f\geq0.01F.
\end{cases}
\]
The schedule advances with processed targets and keeps its original horizon
through halving and continuation. Full-corpus trials use BF16, gradient clipping
at 1.0, auxiliary AdamW betas $(0.9,0.95)$, Muon momentum 0.95 with Nesterov,
five Newton--Schulz steps, a 256-frame context, and a 65,536-target global batch.
Survivors retain both optimizer states and scheduler state.

Sampling and fixed validation windows follow Appendix~\ref{app:data}.
Checkpoints preserve weights, both optimizers, the scheduler, processed-target
and sampler cursors, random-number states, and dataset identity.

At 10M, the lower-weight-decay and lower-Muon-rate recipes differ by only
0.000026 NLL at rung 3. Re-evaluation of identical weights on another container
showed approximately 0.00003 NLL variation, providing an empirical estimate
of numerical variability. The initially selected
lower-decay recipe then regressed. After the original 10M and 75M continuations diverged, recovery restarted
the two 10M finalists and the selected 75M recipe from initialization with
immutable checkpoints. Validation ran every 0.5B targets and at the original rung boundaries.

The selection rule uses minimum validation loss, with longer stability resolving
comparisons within the observed re-evaluation variation. The 10M recipes remain
within 0.0005 NLL through 8B targets and both reach their minimum at step
122,064. The lower-Muon-rate recipe reaches 0.796817 versus 0.797050 for
lower weight decay, then remains near 0.80 through 10.15B targets. The lower-decay
run diverges between 8.5B and 9B. The extended trajectories resolve the stability
comparison and support the final selection: the selected recipe has both the
lower minimum and the longer stable continuation.

\begin{table}[!htbp]
\caption{Recovery checkpoints and complete continuation lengths.
The test split is evaluated once for each frozen winner. A blank test entry
indicates that the runner-up was excluded from this evaluation.}
\label{tab:recovery-checkpoints}
\centering\fontsize{9}{11}\selectfont\setlength{\tabcolsep}{4pt}
\begin{tabular}{@{}lrrrr@{}}\toprule
Recipe & Best step & Validation & Test & Stop targets\\\midrule
10M Muon-low & 122,064 & 0.796817 & 0.797658 & 11,811,160,064\\
10M WD-low & 122,064 & 0.797050 & & 9,663,676,416\\
75M Muon-low & 86,016 & 0.764821 & 0.765435 & 7,516,192,768\\\bottomrule
\end{tabular}
\end{table}

The frozen 10M and 75M checkpoints contain 7,999,586,304 and 5,637,144,576
processed targets. The component losses are approximately 0.26732 and 0.25287
for the button head and 0.52950 and 0.51194 for the main-stick head.
Evaluating the calibration fit at these exact exposures predicts 0.759819 and
0.728842 NLL, giving observed offsets of 0.036998 and 0.035979.
These evaluations also differ in schedule progress: calibration endpoints
complete their cosine decay, whereas the selected full-corpus checkpoints
precede most of it.

Divergence is accompanied by a rise in gradient norms from a healthy median
near 0.07 to medians of 5 to 6, with spikes above 30. At the affected exposures,
the long planned horizons leave learning-rate multipliers near
0.95--0.98 for 10M and 0.82--0.89 for 75M. Runs are stopped after validation
confirms the degradation, and the best earlier checkpoints are retained.
\endgroup

\par
\section{Throughput Optimization and Hardware Allocation}
\label{app:throughput}

\subsection{Controlled Execution and Input-Delivery Sweeps}
\label{app:throughput-protocol}
We benchmark three exact policy profiles (20,040,877, 50,127,021, and
75,305,709 parameters) on A100, H100 SXM, and RTX PRO 6000 GPUs. Each
GPU/profile pair is evaluated with 8, 16, and 32 physical CPU cores and
64 GiB of host memory, giving 27 configurations. Global batch size remains
65,536 targets, with 256-frame contexts, BF16 compute, and Muon plus auxiliary
AdamW. Model initialization, replay ordinals, collation, and input tensors are
held fixed.

An initial 32-core study selects execution geometry for each GPU/profile pair
with live parsing and collation removed. The subsequent delivery sweeps keep
that geometry fixed across CPU tiers and vary the number of loader workers and
prefetched batches. The worker grids are
\begin{equation}
\begin{aligned}
\mathcal W_8&=\{0,2,4,6,8,10,12\},\\
\mathcal W_{16}&=\{0,4,8,12,16,20,24\},\\
\mathcal W_{32}&=\{0,8,16,24,32,40,48\}.
\end{aligned}
\end{equation}
Each positive worker count is crossed with prefetch factors $\{1,2,4\}$;
the zero-worker setting supplies one baseline. This gives 19 candidates per configuration
and 513 delivery candidates overall. All selected recipes use compiled
execution without activation checkpointing.

For $K$ synchronized updates of duration $t_i$ seconds and batch size
$B=65{,}536$, we measure throughput as
\begin{equation}
T_{\mathrm{agg}}=\frac{KB}{\sum_{i=1}^{K}t_i}
=\left(\frac{1}{K}\sum_{i=1}^{K}\frac{1}{f_i}\right)^{-1},
\qquad f_i=B/t_i.
\label{eq:throughput-aggregate}
\end{equation}
This aggregate includes input stalls, host-to-device transfer, forward and
backward passes, and optimizer updates. Each candidate is timed for 24 updates.
Selection maximizes the moving-block-bootstrap 95\% lower confidence bound;
the selected recipe then receives three non-overlapping 96-update runs.
Final intervals resample whole repetitions and circular moving blocks of
ordered updates.

Every configuration passes a byte-exact comparison against the zero-worker
reference on 4,096 windows, covering replay identity, tensor shapes, dtypes,
target counts, and contents. Data delivery preserves order with
\texttt{shuffle=False}, \texttt{in\_order=True}, and
\texttt{drop\_last=False}. Separate controls cover 4,096 producer windows and
an exact page-locked compute corpus. The compute control retains transfer and
training work while removing live parsing and collation. Producer-only timing
includes iterator and worker startup, so it is interpreted separately from
steady training throughput.

\subsection{Measured Recipes and Model-Size Assignments}
\label{app:throughput-recipes}
Table~\ref{tab:throughput-recipes} gives all 27 selected recipes and their
repeated-run measurements. The fastest measured configuration changes with
model size: A100 with 32 cores at 20M (371.5k targets/s), H100 SXM with
32 cores at 50M (245.3k), and RTX PRO 6000 with 8 cores at 75M (137.0k).
Increasing CPU allocation does not consistently improve delivery. For example,
the 75M RTX PRO 6000 runs reach 137.0k, 123.0k, and 119.4k targets/s with
8, 16, and 32 cores.

The compute-only and end-to-end rankings can also differ. At 75M, H100 reaches
216.2k targets/s in the compute control versus 170.4k on RTX PRO 6000, while
their end-to-end rates are 121.7k and 137.0k. We therefore use end-to-end
measurements for hardware allocation. Speed and cost favor different CPU tiers
at 50M: H100 with 32 cores is fastest, while H100 with 8 cores has the lowest
tested cost at \$6.54 per billion targets.

\par
\begin{table}[!ht]
\centering
\fontsize{8.5}{10.5}\selectfont
\setlength{\tabcolsep}{4pt}
\renewcommand{\arraystretch}{1.14}
\begin{tabular}{@{}rrccrr@{}}
\toprule
\shortstack{Model\\(M)} & \shortstack{CPU\\cores} &
\shortstack{Microbatch\\/ accumulation} & \shortstack{Workers\\/ prefetch} &
\shortstack{Targets/s, thousands\\(95\% interval)} & \shortstack{USD\\/ 1B targets}\\
\midrule
\multicolumn{6}{@{}l}{\textbf{RTX PRO 6000}}\\[2pt]
20.04 & 8  & 64/4 & 6/2  & 165.0 [160.3, 170.8] & 6.60\\
20.04 & 16 & 64/4 & 12/1 & 146.8 [142.9, 150.1] & 8.14\\
20.04 & 32 & 64/4 & 0/2  & \textbf{187.1} [179.1, 195.5] & 7.51\\
50.13 & 8  & 64/4 & 0/2  & 131.3 [125.4, 136.3] & 8.30\\
50.13 & 16 & 64/4 & 8/2  & \textbf{139.2} [136.3, 142.4] & 8.58\\
50.13 & 32 & 64/4 & 40/1 & 115.4 [111.1, 121.5] & 12.17\\
75.31 & 8  & 64/4 & 6/2  & \textbf{137.0} [135.3, 138.6] & 7.95\\
75.31 & 16 & 64/4 & 16/1 & 123.0 [119.7, 126.8] & 9.71\\
75.31 & 32 & 64/4 & 48/1 & 119.4 [112.2, 127.2] & 11.76\\
\midrule
\multicolumn{6}{@{}l}{\textbf{H100 SXM}}\\[2pt]
20.04 & 8  & 128/2 & 12/2 & \textbf{345.9} [334.1, 356.9] & 3.89\\
20.04 & 16 & 128/2 & 24/1 & 197.8 [174.5, 225.5] & 7.33\\
20.04 & 32 & 128/2 & 48/1 & 276.8 [260.4, 295.6] & 6.00\\
50.13 & 8  & 128/2 & 10/4 & 205.6 [201.4, 210.0] & 6.54\\
50.13 & 16 & 128/2 & 24/1 & 180.9 [176.5, 186.2] & 8.01\\
50.13 & 32 & 128/2 & 48/2 & \textbf{245.3} [234.6, 257.8] & 6.77\\
75.31 & 8  & 64/4 & 12/2 & \textbf{121.7} [106.3, 132.5] & 11.04\\
75.31 & 16 & 64/4 & 4/2  & 94.9 [89.9, 99.5] & 15.27\\
75.31 & 32 & 64/4 & 0/2  & 86.5 [83.3, 89.2] & 19.19\\
\midrule
\multicolumn{6}{@{}l}{\textbf{A100}}\\[2pt]
20.04 & 8  & 256/1 & 12/1 & 331.7 [325.2, 338.7] & 2.84\\
20.04 & 16 & 256/1 & 24/4 & 260.4 [245.8, 276.2] & 4.02\\
20.04 & 32 & 256/1 & 48/1 & \textbf{371.5} [356.2, 386.9] & 3.38\\
50.13 & 8  & 128/2 & 4/4  & 114.1 [111.7, 117.0] & 8.25\\
50.13 & 16 & 128/2 & 8/2  & 118.4 [110.1, 124.9] & 8.84\\
50.13 & 32 & 128/2 & 48/4 & \textbf{150.5} [145.3, 156.3] & 8.35\\
75.31 & 8  & 64/4 & 4/2  & 75.8 [73.0, 77.9] & 12.42\\
75.31 & 16 & 64/4 & 0/2  & 76.3 [74.8, 77.9] & 13.71\\
75.31 & 32 & 64/4 & 40/2 & \textbf{105.5} [104.5, 106.6] & 11.91\\
\bottomrule
\end{tabular}
\caption{Complete selected execution and delivery recipes. Microbatch size
counts 256-frame windows; accumulation preserves the 65,536-target global
batch. Bold rates identify the fastest CPU tier for each GPU/model pair.
Prefetch is inactive when there are zero workers; its stored setting is
retained. All hosts have 64 GiB RAM. Costs use the GPU, CPU, and memory rates
recorded on August 27, 2026 and exclude startup.}
\label{tab:throughput-recipes}
\end{table}
\par
\subsection{Throughput Fits}
\label{app:throughput-fits}
For GPU family $g$ and CPU allocation $c$, the three measured model sizes
define a power-law fit
\begin{equation}
\widehat T_{g,c}(N)=A_{g,c}
\left(\frac{N}{50{,}127{,}021}\right)^{-\beta_{g,c}},
\qquad
\widehat T_{g,\mathrm{env}}(N)
=\max_{c\in\{8,16,32\}}\widehat T_{g,c}(N).
\label{eq:throughput-law}
\end{equation}
Here $T$ and $A$ are measured in targets/s. The maximum across fitted CPU
branches describes how the preferred host allocation varies with model size.
We also fit a separate compact law of the same form to each GPU's three
measured maximum-throughput anchors. Table~\ref{tab:throughput-fits} gives
both sets of coefficients, rounded to the displayed precision. At measured
sizes, allocation uses the measured rate. The compact fits supply interpolated
30M and 40M values for maximum-throughput planning. The fixed-core fits support
CPU-tier comparisons and projections under a dollar budget.

\begin{table}[htbp]
\centering\small
\setlength{\tabcolsep}{5pt}
\renewcommand{\arraystretch}{1.12}
\begin{tabular}{@{}lrrrrr@{}}
\toprule
GPU & CPU cores & $A$ (k/s) & $\beta$ & $R^2_{\log}$ & Max. residual\\
\midrule
\multicolumn{6}{@{}l}{\textbf{Fixed-core fits}}\\[2pt]
RTX PRO 6000 & 8  & 139.9 & 0.159 & 0.785 & 6.2\%\\
RTX PRO 6000 & 16 & 133.2 & 0.121 & 0.815 & 4.5\%\\
RTX PRO 6000 & 32 & 128.7 & 0.371 & 0.870 & 10.4\%\\
H100 SXM & 8  & 180.7 & 0.751 & 0.952 & 13.8\%\\
H100 SXM & 16 & 138.6 & 0.477 & 0.652 & 30.5\%\\
H100 SXM & 32 & 158.8 & 0.752 & 0.636 & 54.5\%\\
A100 & 8  & 117.4 & 1.123 & 0.999 & 2.8\%\\
A100 & 16 & 113.8 & 0.916 & 0.997 & 4.0\%\\
A100 & 32 & 153.6 & 0.957 & 0.999 & 2.0\%\\
\midrule
\multicolumn{6}{@{}l}{\textbf{Compact fits to measured maxima}}\\[2pt]
RTX PRO 6000 & & 146.5 & 0.250 & 0.934 & \\
H100 SXM & & 192.7 & 0.719 & 0.838 & \\
A100 & & 153.6 & 0.957 & 0.999 & \\
\bottomrule
\end{tabular}
\caption{Throughput coefficients and fit diagnostics. The normalization size
is 50,127,021 parameters. Each fixed-core fit uses three measurements at one
CPU allocation; each compact fit uses the best measured CPU tier at each size.
Maximum residual is $\max_i |T_i-\widehat T_i|/\widehat T_i$ over the three anchors.}
\label{tab:throughput-fits}
\end{table}

Each law has three anchors and one residual degree of freedom. The H100
16-core and 32-core curves have log-space $R^2$ of 0.652 and 0.636; the latter
misses an anchor by 54.5\%. We retain the measured winner at each anchor when
the fitted ordering differs. For example, compact-curve intersections place
the H100/RTX PRO 6000 crossover at 90.05M, while the measured speed leader
already changes between 50.13M and 75.31M. Bootstrap intervals characterize
timing variation under the fitted form; they do not quantify its structural
error at unmeasured sizes.

\par
\subsection{Fixed-Dollar Model and Hardware Selection}
\label{app:hardware-budget}
We combine the loss surface of Equation~\ref{eq:pretraining-scaling} with the
targets delivered by each GPU/CPU tier. For dollar budget $S$, hourly price
$r_{g,c}$, model-dependent startup allowance $\tau_{N,g,c}$ seconds, and batch size $B$, the
batch-aligned exposure is
\begin{equation}
D(N,g,c;S)=B\left\lfloor\frac{T_{g,c}(N)}{B}
\max\left\{\frac{3{,}600S}{r_{g,c}}-\tau_{N,g,c},0\right\}\right\rfloor.
\label{eq:hardware-budget}
\end{equation}
We select $(N,g,c)$ to minimize $\widehat L(N,D(N,g,c;S))$ over the candidate
set. Prices include the GPU, physical CPU cores, and 64 GiB of host memory.
The steady-state cost per billion targets is
\[
\operatorname{cost}_{1\mathrm B}=\frac{r_{g,c}10^9}{3{,}600T_{g,c}}.
\]

The compact-policy candidate set remains below the approximately 12M
parameters of the Slippi-AI reference policy. The larger-policy search covers
30M to 75M. Table~\ref{tab:hardware-budget} records the evaluated choices.
The compact-policy comparison uses a 29.5-hour planning horizon, within the
\$100 budget. The selected 10M policy uses A100 with 8 cores and the fixed-core fit
\begin{equation}
\widehat T_{\mathrm{A100},8}(N)=117{,}405.4065
\left(\frac{N}{50{,}127{,}021}\right)^{-1.12340643},
\label{eq:a100-8core}
\end{equation}
whose log-space $R^2$ is 0.998907. At the exact 10,163,629-parameter count,
it predicts 705,074 targets/s by extrapolation. The same CPU branch gives the
30M and 40M A100 projections in Table~\ref{tab:hardware-budget}.
The execution recipe transfers
from the 20M A100/8-core measurement: 256 windows per microbatch, one
accumulation step, 12 workers, and prefetch factor 1. The selected 75M policy
uses the directly measured RTX PRO 6000/8-core recipe: 64 windows per
microbatch, four accumulation steps, six workers, and prefetch factor 2,
delivering 137,014 targets/s.

\begin{table}[htbp]
\centering\small
\setlength{\tabcolsep}{4pt}
\renewcommand{\arraystretch}{1.15}
\begin{tabular}{@{}rlrrrr@{}}
\toprule
Model (M) & GPU (8 CPU cores) & Hours & Targets/s & Targets (B) & Pred. NLL\\
\midrule
\multicolumn{6}{@{}l}{\textbf{Compact-policy candidates, \$100}}\\[2pt]
5.054 & A100 & 29.50 & 1,545,647 E & 164.148 & 0.763553\\
7 & A100 & 29.50 & 1,071,942 E & 113.840 & 0.756780\\
8 & A100 & 29.50 & 922,620 E & 97.982 & 0.754180\\
\textbf{10.164} & \textbf{A100} & 29.50 & 705,074 E & 74.879 & \textbf{0.749773}\\
\midrule
\multicolumn{6}{@{}l}{\textbf{Larger-policy candidates, \$200}}\\[2pt]
30 & A100 & 59.02 & 209,003 F & 44.406 & 0.731305\\
40 & A100 & 59.02 & 151,285 F & 32.143 & 0.727819\\
50.127 & H100 SXM & 41.33 & 205,625 M & 30.591 & 0.724576\\
\textbf{75.306} & \textbf{RTX PRO 6000} & 51.00 & 137,014 M & 24.982 & \textbf{0.719589}\\
\bottomrule
\end{tabular}
\caption{Fixed-dollar planning comparisons. M denotes measured throughput,
F fitted interpolation, and E extrapolation. Bold entries are selected within
each candidate set. Exposures count processed targets, including repeated
frames. The 75M budget includes the measured cold-start allowance.}
\label{tab:hardware-budget}
\end{table}

The hourly rates used for the final selections are \$3.3888 for A100/8-core
and \$3.9216 for RTX PRO 6000/8-core. The \$100 and \$200 budgets apply to
the intended winning continuations; the winner's first five sweep hours count
toward its allocation. Other sweep arms add 11 and 26 GPU-hours, respectively.
The three recovery runs subsequently use 17.7 hours for the lower-Muon-rate
10M recipe, 23.6 hours for the lower-decay 10M recipe, and 37.9 hours for
75M, approximately \$289 in total. These additional runs are separate from
the original planning budgets.

\par
\subsection{Hardware Provenance and Measurement Scope}
\label{app:hardware-provenance}
The hardware study uses source snapshot \texttt{229b26f2a955}, software-image
commit \texttt{268031e7bdde}, dataset manifest \texttt{30981903d542}, and
split manifest \texttt{526369d572ad}. RTX PRO 6000 allocations report NVIDIA
RTX PRO 6000 Blackwell Server Edition at 600 W; H100 allocations report
NVIDIA H100 80GB HBM3 at 700 W. The A100 family includes SXM4-80GB and
80GB PCIe allocations. All three A100 maximum-throughput anchors use SXM4
at 500 W. Its 8-core and 16-core fits each include one PCIe allocation
(50M/8-core and 75M/16-core, respectively), so those curves also reflect
the subtype mix delivered by the provider.

Repeated-run throughput is below the selected 24-update tuning estimate in
25 of 27 configurations, with a median ratio of 0.893. Seven finalist
intervals have width exceeding 10\% of their point estimate. Eleven selected
recipes use the largest successful worker count in their sweep. These results
describe the tested configurations; longer comparisons of every candidate or
additional CPU tiers could change the winners. At 20M, the winning A100 and
H100 timing intervals overlap slightly. The reported cross-GPU rankings use
point estimates from the repeated runs.

\par
\section{Post-Training Details}
\label{app:posttraining}
\begingroup
\setlength{\parskip}{3pt}
\subsection{Labels, Curriculum Pools, and Evaluation}
\label{app:curriculum}
Rank labels are read per port from the anonymized netplay-name field of each
raw replay. End-of-game placement identifies the winner, with final stocks
used when placement is unavailable. The label pass covers 849,935 raw replays;
849,895 have both rank labels. Placement supplies 579,281 outcomes and final
stocks supply 270,533; 121 remain undecided. Within the existing training split,
823,083 replays are available and 823,057 have a decided winner.

On the full label pass, master players win 61.0\% of 140,043 games against
diamond players and 66.2\% of 119,909 against platinum players. Diamond players
win 59.1\% of 184,023 games against platinum players. Rank and game outcome
therefore provide complementary filters for selecting demonstrations.

\begin{table}[!htbp]
\caption{Eligible replay pools. Later rounds use every master-winning
replay and a deterministic sample of diamond-winning replays, giving
approximately 90\% master and 10\% diamond visits.}
\label{tab:posttraining-pools}
\centering\fontsize{9}{11}\selectfont\setlength{\tabcolsep}{8pt}
\begin{tabular}{@{}lrl@{}}\toprule
Pool & Replays & Eligible perspective\\\midrule
Natural distribution & 823,083 & Both players\\
All decided games & 823,057 & Winner\\
Master or diamond winners & 471,316 & Winner\\
Master winners & 256,532 & Winner\\
Diamond winners & 214,784 & Winner\\
Diamond subset in later mixture & 28,504 & Winner\\\bottomrule
\end{tabular}
\end{table}

Filtering changes eligible replay lists and the imitated perspective. Each visit
still draws 32 windows, and each optimizer update combines eight replay
identities. Eligible lists are shuffled in contiguous chunks of 128 replays.
This addresses rank-correlated storage order while preserving shard locality.
A sequential traversal would expose the 10M natural stage to 0.7\%
master--diamond games, compared with 16.5\% in its pool, and the 75M master
stage to 59.8\% master--master games, compared with 38.0\% in its pool.

Validation uses the same 8,483 held-out games throughout: one 256-frame window
per player, totaling 16,966 windows and 4,343,296 targets. Windows are collated
once per container and cached in host memory, occupying approximately 2.5\,GB.
Evaluation runs at initialization and approximately every 50M additional targets,
with additional evaluations at stage and run boundaries.

\begin{table}[!htbp]
\caption{Rank--outcome validation slices. An additional undecided
master game contributes two perspectives and 512 targets; it is excluded
from the six decided-outcome slices.}
\label{tab:posttraining-slices}
\centering\fontsize{9}{11}\selectfont\setlength{\tabcolsep}{12pt}
\begin{tabular}{@{}lrr@{}}\toprule
Slice & Perspectives & Targets\\\midrule
Master won & 2,635 & 674,560\\
Master lost & 1,932 & 494,592\\
Diamond won & 2,235 & 572,160\\
Diamond lost & 2,208 & 565,248\\
Platinum won & 3,612 & 924,672\\
Platinum lost & 4,342 & 1,111,552\\\bottomrule
\end{tabular}
\end{table}

For slice $S$, $L_S=|S|^{-1}\sum_{t\in S}\ell_t$, where $\ell_t$ is the
teacher-forced two-head controller NLL. Checkpoint selection minimizes
$W=0.9L_{\rm master\text{-}won}+0.1L_{\rm diamond\text{-}won}$.
As the curriculum narrows toward stronger players' winning actions, overall
NLL can rise even while the target slice improves. The fixed weights prioritize
Master-winning demonstrations while retaining Diamond winners in checkpoint
selection. Later training mixtures follow this weighting. The all-target NLL
continues to track performance on the broader replay distribution.
The fixed test manifest contains 8,376 games and 16,752 windows and is evaluated
once per selected checkpoint. Test results do not enter selection.

\par
\subsection{Training Schedules and the Four-Stage Curriculum}
\label{app:posttraining-schedules}
The post-training schedules use a 10M Muon reference rate of 0.011257732499 and an auxiliary
AdamW reference of 0.00140722. The corresponding 75M references are 0.005
and 0.001.
The multipliers below apply to those references. Optimizer state carries over
between rounds. Context length 256, global batch 65,536, BF16, gradient
clipping at 1.0, and the per-card microbatch and accumulation geometry remain
fixed. The 10M uses an SXM A100-80GB and the 75M uses an RTX PRO 6000.

Round 1 starts from the frozen pretraining checkpoints and applies one cosine
schedule over its full horizon $F$, with no warmup:
\[
m(f)=0.001+\frac{0.099}{2}\left[1+\cos(\pi f/F)\right].
\]
For 10M, this decays the Muon learning rate from 0.001125773250 to
0.0000112577325.
The four data stages consume 15\%, 15\%, 20\%, and 50\% of the budget.
Stage boundaries change the eligible data while the same cosine continues.
Weight decay retains its pretraining value in this round.

\begin{table}[!htbp]
\caption{Round 1 stage endpoints. Frames are targets processed within
the stage, in millions. $L$ is overall validation NLL and $W$ is the weighted
selection loss. The full-corpus stage precedes every filtered stage.}
\label{tab:curriculum-stage-endpoints}
\centering\fontsize{9}{11}\selectfont\setlength{\tabcolsep}{5pt}
\begin{tabular}{@{}llrrrr@{}}\toprule
Model & Stage & Frames & $L$ & Master-won & $W$\\\midrule
10M & Natural & 270 & 0.77510 & 0.79191 & 0.79024\\
10M & All winners & 270 & 0.77233 & 0.78693 & 0.78543\\
10M & Master + diamond winners & 360 & 0.77166 & 0.77798 & 0.77708\\
10M & Master winners & 900 & 0.77282 & 0.77078 & 0.77080\\\addlinespace[2pt]
75M & Natural & 180 & 0.74519 & 0.75550 & 0.75426\\
75M & All winners & 180 & 0.74230 & 0.74977 & 0.74873\\
75M & Master + diamond winners & 240 & 0.74214 & 0.74109 & 0.74069\\
75M & Master winners & 600 & 0.74286 & 0.73566 & 0.73591\\\bottomrule
\end{tabular}
\end{table}

Later rounds hold the rate before a final decay~\cite{hu2024}. For hold $h$,
floor $e=h/100$, warmup end $F_w=0.02F$, and decay start $F_d=0.8F$,
\[
m(f)=\begin{cases}
e+(h-e)f/F_w, & f<F_w,\\
h, & F_w\leq f<F_d,\\
e+\dfrac{h-e}{2}\!\left[1+\cos\!\left(\pi\dfrac{f-F_d}{F-F_d}\right)\right],
& f\geq F_d.
\end{cases}
\]
Weight decay follows $\lambda(f)=\lambda_0m(f)$. This change follows a
12--15\% reduction in backbone weight norms during round 1. Later rounds use
the master/diamond winner mixture throughout.

\begin{table}[!htbp]
\caption{Executed rounds. Horizons are planned processed targets.
A blank horizon for the stopped 75M continuation indicates that only its
449M-target stopping point is used here. KD controls each run for 300M targets.}
\label{tab:posttraining-rounds}
\centering\fontsize{9}{11}\selectfont\setlength{\tabcolsep}{6pt}
\begin{tabular}{@{}llrll@{}}\toprule
Round & Model & Horizon & Schedule & Objective\\\midrule
1 & 10M / 75M & 1.8B / 1.2B & Cosine $0.1\to0.001$ & NLL\\
2 & 10M / 75M & 1.2B / 1.5B & Hold $h=0.1$ & NLL\\
3 & 10M & 900M & Hold $h=0.1$ & KD, $\alpha=0.5$, $T=1$\\
3, stopped & 75M & & Hold $h=0.1$ & NLL\\
4 & 10M & 900M & Hold $h=0.05$ & KD, $\alpha=0.5$, $T=1$\\
4, control & 10M & 300M & Hold $h=0.05$ & KD, $\alpha=0.7$, $T=1$\\
4, control & 10M & 300M & Hold $h=0.05$ & KD, $\alpha=0.5$, $T=2$\\\bottomrule
\end{tabular}
\end{table}

\par
\subsection{Knowledge Distillation and the Stopped 75M Continuation}
\label{app:distillation}
\label{app:posttraining-stopping}
The teacher is the frozen 75M round 2 checkpoint, step 127,214. The student
starts round 3 from the 10M round 2 checkpoint and round 4 from its round 3
winner. Both use the teacher throughout. Distillation~\cite{hinton2015,rusu2016}
combines demonstration NLL with the sum of teacher-to-student KL divergences
over the two controller heads, averaged over valid target positions:
\[
\begin{gathered}
\mathcal L=(1-\alpha)L_{\rm NLL}
+\alpha T^2\sum_c\mathrm{KL}(p^{(c)}_T\Vert q^{(c)}_T),\\
p_T=\operatorname{softmax}(z_{\rm teacher}/T),\quad
q_T=\operatorname{softmax}(z_{\rm student}/T).
\end{gathered}
\]
Validation always evaluates the student's NLL at $T=1$. Round 3 uses an eager
teacher in 64-window chunks, reaching 76k targets/s at 11.7\,GB peak memory.
Round 4 compiles the teacher and increases the host allocation from 8 cores and
12 loader workers to 16 cores and 16 workers, reaching 196k targets/s after a
28-minute compilation. The changes are applied together.
In round 3, training-batch KL to the teacher averages approximately 0.045 nats.
This diagnostic uses the sum over both controller heads, as in the objective.

\begin{table}[!htbp]
\caption{Round 4 control trajectories. All runs start at step 181,516.
The 50--200M evaluations share the held rate; warmup lasts 6M targets for
controls and 18M for the main run. Controls begin decaying at 240M, and the
900M main run at 720M. $L$ is overall NLL.}
\label{tab:kd-controls}
\centering\fontsize{9}{11}\selectfont\setlength{\tabcolsep}{6pt}
\begin{tabular}{@{}rrrrrr@{}}\toprule
& & \multicolumn{2}{c}{$\alpha=0.7$} & \multicolumn{2}{c}{$T=2$}\\
Targets (M) & Main $W$ & $W$ & $L$ & $W$ & $L$\\\midrule
0 & 0.75990 & 0.75984 & 0.77071 & 0.75982 & 0.77070\\
50 & 0.76086 & 0.76075 & 0.77074 & 0.76268 & 0.77257\\
100 & 0.76106 & 0.76085 & 0.77042 & 0.76230 & 0.77175\\
150 & 0.76075 & 0.76041 & 0.77141 & 0.76289 & 0.77362\\
200 & 0.76053 & 0.76025 & 0.76993 & 0.76289 & 0.77246\\
250 & 0.76123 & 0.76073 & 0.76957 & 0.76376 & 0.77249\\
300 & 0.76016 & 0.75896 & 0.76939 & 0.76159 & 0.77178\\
300, final tail & & 0.75898 & 0.76941 & 0.76158 & 0.77178\\\bottomrule
\end{tabular}
\end{table}

At the four evaluations with the same constant learning rate, increasing the teacher weight reduces
$W$ by 0.00011--0.00034, while $T=2$ increases it by 0.00124--0.00236.
The 300M heavier-teacher control selects step 186,088; the temperature control
keeps its source because every trained checkpoint is worse. The main run
continues to 900M and supplies the final 10M checkpoint. Relative to each run's
re-scored source, the weighted gains are 0.00244 in round 4 and 0.00461 in
round 3. These diminishing gains motivated
the end of the supervised 10M sequence.

\begin{table}[!htbp]
\caption{The 75M continuation after round 2. The learning-rate
multiplier is 0.1 at every trained checkpoint. All metrics are validation
measurements except the interval median of the training gradient norm.}
\label{tab:75m-stopped-posttraining}
\centering\fontsize{9}{11}\selectfont\setlength{\tabcolsep}{6pt}
\begin{tabular}{@{}rrrrr@{}}\toprule
Step & Targets (M) & $W$ & Overall NLL & Gradient median\\\midrule
127,214 & 0 & 0.72492 & 0.74149 & \\
127,976 & 50 & 0.72756 & 0.74431 & 0.1205\\
128,738 & 100 & 0.72734 & 0.74400 & 0.1255\\
129,500 & 150 & 0.72771 & 0.74383 & 0.1263\\
130,262 & 200 & 0.72771 & 0.74504 & 0.1304\\
131,024 & 250 & 0.72790 & 0.74235 & 0.1350\\
131,786 & 300 & 0.72804 & 0.74503 & 0.1341\\
132,548 & 350 & 0.72865 & 0.74621 & 0.1405\\
133,310 & 400 & 0.73137 & 0.74950 & 0.1500\\
134,072 & 449 & 0.74430 & 0.76510 & 0.1925\\\bottomrule
\end{tabular}
\end{table}

The additional 75M cycle never recovers its starting loss. It is stopped
manually at 449M targets as losses and gradient norms rise, before the automatic
guards trigger. Step 127,214 remains the 75M checkpoint and the distillation
teacher. RL checkpoint results are reported in Section~\ref{sec:benchmark};
the RL methods are described in Section~\ref{sec:reinforcement-learning}.

\par
\subsection{Selected Checkpoints and Slice Diagnostics}
\label{app:posttraining-checkpoints}
\begin{table}[!htbp]
\caption{Validation metrics at every selected checkpoint. Round names
follow the chronological experiments. Values come from the selected checkpoint
rows; final alignment steps can differ slightly from those rows.}
\label{tab:posttraining-selected-validation}
\centering\fontsize{9}{11}\selectfont\setlength{\tabcolsep}{5pt}
\begin{tabular}{@{}llrrrrr@{}}\toprule
Model & Checkpoint & Step & Overall & Master-won & Diamond-won & $W$\\\midrule
10M & Pretrained & 122,064 & 0.79682 & 0.81315 & 0.79691 & 0.81152\\
10M & Round 1 & 149,512 & 0.77281 & 0.77076 & 0.77094 & 0.77078\\
10M & Round 2 & 167,800 & 0.77220 & 0.76381 & 0.77001 & 0.76443\\
10M & Round 3 & 181,516 & 0.77071 & 0.75889 & 0.76825 & 0.75983\\
10M & Round 4 & 195,248 & 0.76896 & 0.75647 & 0.76641 & 0.75746\\
75M & Pretrained & 86,016 & 0.76481 & 0.77617 & 0.76467 & 0.77502\\
75M & Round 1 & 104,326 & 0.74286 & 0.73566 & 0.73810 & 0.73591\\
75M & Round 2 & 127,214 & 0.74146 & 0.72375 & 0.73516 & 0.72489\\\bottomrule
\end{tabular}
\end{table}

\begin{table}[!htbp]
\caption{Test results at the same checkpoints. Every selected round
improves all four reported metrics relative to the preceding selected round.}
\label{tab:posttraining-selected-test}
\centering\fontsize{9}{11}\selectfont\setlength{\tabcolsep}{8pt}
\begin{tabular}{@{}llrrrr@{}}\toprule
Model & Checkpoint & Overall & Master-won & Diamond-won & $W$\\\midrule
10M & Pretrained & 0.79766 & 0.79694 & 0.79928 & 0.79717\\
10M & Round 1 & 0.77356 & 0.75469 & 0.77223 & 0.75644\\
10M & Round 2 & 0.77291 & 0.74758 & 0.77119 & 0.74994\\
10M & Round 3 & 0.77137 & 0.74280 & 0.76886 & 0.74541\\
10M & Round 4 & 0.76974 & 0.74044 & 0.76722 & 0.74312\\
75M & Pretrained & 0.76543 & 0.75915 & 0.76714 & 0.75995\\
75M & Round 1 & 0.74338 & 0.72097 & 0.73899 & 0.72277\\
75M & Round 2 & 0.74206 & 0.70906 & 0.73570 & 0.71173\\\bottomrule
\end{tabular}
\end{table}

\begin{table}[!htbp]
\caption{Every rank--outcome slice before and at the end of round 1.
Both sizes improve on every slice relative to pretraining, with the largest
reductions on master-winning actions.}
\label{tab:curriculum-all-slices}
\centering\fontsize{9}{11}\selectfont\setlength{\tabcolsep}{9pt}
\begin{tabular}{@{}lrrrr@{}}\toprule
& \multicolumn{2}{c}{10M} & \multicolumn{2}{c}{75M}\\
Slice & Pretrained & Round 1 & Pretrained & Round 1\\\midrule
Master won & 0.81315 & 0.77078 & 0.77617 & 0.73566\\
Master lost & 0.79648 & 0.76365 & 0.76250 & 0.73142\\
Diamond won & 0.79691 & 0.77098 & 0.76467 & 0.73810\\
Diamond lost & 0.79777 & 0.77552 & 0.76672 & 0.74479\\
Platinum won & 0.79945 & 0.78122 & 0.76855 & 0.75292\\
Platinum lost & 0.78444 & 0.77083 & 0.75502 & 0.74552\\\bottomrule
\end{tabular}
\end{table}

\begin{table}[!htbp]
\caption{Controller-head diagnostics on all validation slices.
$H$ denotes the mean predictive entropy of a head, in nats. The final checkpoints
retain button entropy within 10\% of their pretrained values.}
\label{tab:posttraining-head-diagnostics}
\centering\fontsize{9}{11}\selectfont\setlength{\tabcolsep}{7pt}
\begin{tabular}{@{}llrrrr@{}}\toprule
Model & Checkpoint & Button NLL & Stick NLL & Button $H$ & Stick $H$\\\midrule
10M & Pretrained & 0.26732 & 0.52950 & 0.2612 & 0.5300\\
10M & Round 4 & 0.25502 & 0.51394 & 0.2498 & 0.5178\\
75M & Pretrained & 0.25287 & 0.51194 & 0.2521 & 0.5127\\
75M & Round 2 & 0.24311 & 0.49835 & 0.2367 & 0.5001\\\bottomrule
\end{tabular}
\end{table}

The first natural-data stage accounts for approximately 90\% of round 1's
net overall improvement. Learning rate, training exposure, and data composition
change together during the curriculum. Each model follows one trajectory,
without matched-budget continuations on natural or Master-only data. The KD
controls also use one run per setting. These comparisons leave uncertainty
about the contribution of each intervention and variation across training seeds.

\par
\subsection{Schedule and Stopping Diagnostics}
\label{app:posttraining-ledgers}
\label{app:posttraining-diagnostics}
The validation trajectories in Figure~\ref{fig:posttraining-trajectories}
guide the continuation decisions. Table~\ref{tab:posttraining-hold}
reports the final four constant-rate evaluations in round 2. The 10M's
weighted loss changes little over these 150M targets, while the 75M continues
to improve. The complete trajectories, including optimizer steps and
round-specific source evaluations, accompany the figure data.

\begin{table}[!htbp]
\caption{Final constant-rate evaluations in round 2. Targets count
additional training within the round. The learning-rate multiplier is 0.1
throughout. $L$ is overall validation NLL; $W$ is the weighted selection loss.}
\label{tab:posttraining-hold}
\centering\fontsize{9}{11}\selectfont\setlength{\tabcolsep}{10pt}
\begin{tabular}{@{}lrrr@{}}\toprule
Model & Targets (M) & $L$ & $W$\\\midrule
10M & 799 & 0.77472 & 0.76962\\
10M & 849 & 0.77552 & 0.76960\\
10M & 899 & 0.77423 & 0.76979\\
10M & 949 & 0.77441 & 0.76973\\\addlinespace[2pt]
75M & 1049 & 0.74495 & 0.72936\\
75M & 1099 & 0.74367 & 0.72905\\
75M & 1149 & 0.74393 & 0.72895\\
75M & 1199 & 0.74546 & 0.72859\\\bottomrule
\end{tabular}
\end{table}

\textbf{Checkpoint averaging.} Averaging the last 3, 5, or 8 validated
checkpoints after round 1 gives $W=0.77082$, $0.77080$, and $0.77089$ for 10M,
and $0.73595$, $0.73599$, and $0.73608$ for 75M. The best single checkpoints
score 0.77078 and 0.73591. Later rounds retain single checkpoints.

\textbf{Monitoring and stopping.} For each 50M-target interval, the gradient
median is compared with the median of the preceding four intervals in the same
stage, with a stop threshold of $3\times$. A rolling 64-step median is checked
against the same baseline at each step. A rise above 0.10 in interval-mean
training loss or between consecutive overall validation losses also stops the
run. A 10\% drop in per-head entropy triggers a warning. Immutable checkpoints
are written at each validation and every 1,024 optimizer steps.

Figure~\ref{fig:posttraining-controls} shows the additional 75M cycle and the
10M distillation controls. Comparisons at 50--200M targets share the constant
learning rate. The final-decay panel shows each run relative to its own budget;
the controls use 300M targets and the main run uses 900M.
\par
\begin{figure}[!htbp]
\centering
\includegraphics[width=\textwidth]{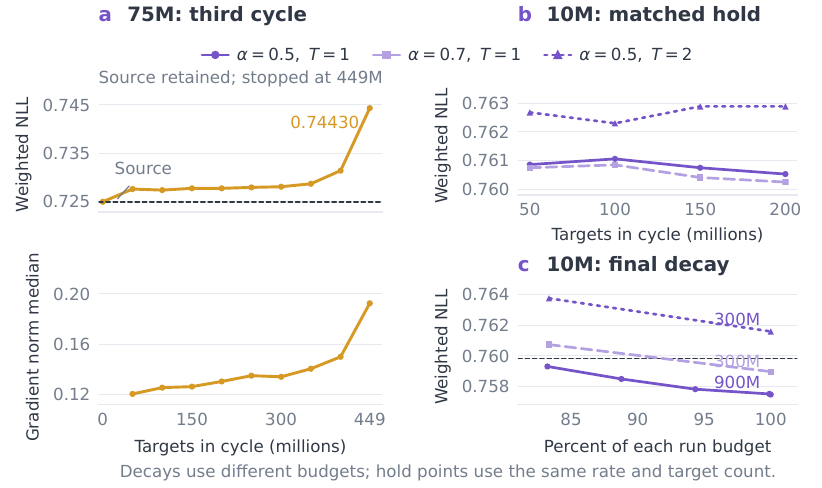}\par
\caption{\textbf{Post-training diagnostics.} (a) Weighted validation NLL and
training gradient-norm medians rise during the additional 75M cycle, which is
stopped at 449M targets. (b) The 10M controls at evaluations with the same held
rate of 0.05: teacher weight 0.7 gives lower weighted NLL at all four points,
while temperature 2 gives higher NLL. (c) Each run's own final decay, shown as
a percentage of its budget. The dashed line is the temperature control's source
loss, 0.75982; its trained endpoint is worse and the source is retained. Lines
connect recorded observations.}
\label{fig:posttraining-controls}
\end{figure}
\endgroup

\subsection{Reinforcement-Learning Details and Additional Results}
\label{app:rl}
This subsection gives the reward implementation, updates, and checkpoint-selection
results for Section~\ref{sec:posttraining-rl}. Training statistics are learner values,
reported as medians over 20 RL steps unless stated otherwise. KL divergences and
entropies are in nats per decision; logged KL values and the update guard average
over decisions. Figure~\ref{fig:rl-lineage} shows the training history.
\par
\textbf{Reward shaping.} In Equation~\ref{eq:rl-reward}, $g^p_t=1$ when player $p$
grabs the ledge while the opponent is not invulnerable and is on the stage side of
that ledge. The offstage indicator $o^p_t=1$ on each frame spent more than 50 units
from the region spanning the stage between its ledges and up to 60 units above it.
The term $c^p_t$ is displacement toward the opponent. The same terms are evaluated
for both players and subtracted to obtain the zero-sum reward. The stock-loss
indicator is one for each lost stock. Games continue without a stock limit or timer;
returns end at an emulator restart, which occurs only at launch or after a fault.
\par
\Needspace{3\baselineskip}
\textbf{Rollouts and updates.} Each step collects 16 consecutive rollouts of 128
frames from each of 96 emulators. These are split by emulator group into 32
trajectories, corresponding to 16 rollouts of two groups. The value network first
takes one Adam update per trajectory. The policy then takes two PPO epochs, with
one Adam update per epoch on the mean of the trajectories' mean losses. With
self-play and CPU9, averaging group losses this way gives a CPU9 decision six times
the weight of a self-play decision, rather than the per-decision weighting intended
in the run configuration.
\par
The update guard restores the policy and its optimizer state to the start of the RL
step if mean $\mathrm{KL}(\pi_{\mathrm{old}}\,\|\,\pi_\theta)$ after both epochs
exceeds 0.005; it never triggered. After warm-up, the median step up to each selected
checkpoint changes the policy by a mean KL of $1.0\times10^{-5}$ at 75M,
$1.3\times10^{-5}$ in the first 10M run, and $2.5\times10^{-5}$ in the leash run.
The guard's threshold is 50 times that of Slippi-AI's released agents~\cite{slippiai}.
\par
All runs use Adam~\cite{kingma2015} with PyTorch defaults~\cite{paszke2019}, no gradient
clipping, sampling at temperature 1, and no entropy bonus. Each starts with six
value-only steps and three zero-learning-rate steps. Learning rates were chosen
in short trial runs; Slippi-AI's releases use $3\times10^{-5}$ and 128 warm-up steps
of each kind. The value network has two Transformer blocks of width 128, its own
frame encoder, and grouped-query attention with two query heads and one key-value
head~\cite{ainslie2023}. It reads the 128 frames preceding each rollout together
with that rollout and minimizes squared error to the bootstrapped return.
\par
\textbf{Policy context and reference approximation.} Each rollout stores the acting
policy's attention cache at its first frame. The learner uses this cache to score
decisions with the same context, up to the 256-frame window, available during play.
At unchanged weights, it reproduces the actor's action distributions to within a
mean KL of $5\times10^{-10}$ per decision. The reference policy also reads this cache;
its own representation of the history would differ once its weights and the acting
policy's weights diverge. The reference penalty and its logged KL therefore use an
approximation. For each KL term, the joint-controller head is evaluated exactly,
while the main-stick term is evaluated at the joint-controller category sampled by
$\pi_{\mathrm{old}}$. For the reference KL terms, this sampling distribution can
also differ from the distribution defining the corresponding joint-policy KL.
Our PyTorch implementation follows Slippi-AI's RL code at revision
\texttt{577965a}~\cite{slippiai}, with the return-boundary, cache, and group-weighting
differences described above.
\par
\textbf{Training budgets and implementation.} The first runs resume across launches
until approximately 72 hours of training, reaching 980 steps at 75M and 2,385 at 10M.
The supervised starting checkpoints are steps 127,214 and 195,248, respectively.
The first 10M run's selected step 1,318 supplies both policy and value weights to
the leash run; its reference remains the supervised 10M. No earlier checkpoint or
other learned agent serves as a training opponent.
\par
After about 14 hours, both first runs resume with an optimized actor, learner, and
emulator exchange that preserve action distributions up to floating-point rounding.
Median time per RL step falls from 256 to 86 seconds at 10M and from 595 to 196
seconds at 75M. The actor uses the sync-free policy step replayed as a CUDA
graph~\cite{cudagraphs}, batched over emulators
(Appendix~\ref{app:inference-optimized}). The learner's forward pass is reorganized
for the stored-cache layout and compiled with \texttt{torch.compile}~\cite{ansel2024}.
Emulator processes exchange frames through shared memory, with the two groups
stepped in alternation. The leash run uses this implementation throughout, at a
median 88 seconds per step up to step 632. Training times include in-loop evaluation.
\par
\textbf{In-loop evaluation.} Evaluation runs about every three hours, at the start
of each run, and at the normal end of each launch. Each opponent is evaluated on
30 emulators, five per stage, for 1,536 frames each: 12.8 minutes of play. The fixed
earlier opponent is the supervised starting checkpoint in the first runs and step
1,318 in the leash run. These rounds are short and noisy. Step 1,318, evaluated four
times across the first run and the three KL-weight runs, gives stock margins of
0.86--1.09 per minute and damage received of 90--93 percentage points per minute
against MIMIC~\cite{mimic}. Each stock-margin estimate has a Poisson standard error
of about 0.4 per minute. Neither first run exceeds a stock margin of 1.1 per minute
or falls below 80 percentage points of damage received per minute against MIMIC.
At step 1,318, $\mathrm{KL}(\pi_\theta\,\|\,\pi_{\mathrm{ref}})=0.0043$; it reaches
0.0047 by step 2,385, another 1,067 steps later.
\par
\textbf{Matches before the leash run.} Each Slippi-AI release~\cite{slippiai} plays
at its own action delay. The comparison uses Fox mirrors, with 16 games on Final
Destination and 18 across six stages. Step 1,318 wins 6/34 against \texttt{master},
0/34 against \texttt{gm}, 19/34 against \texttt{diamond}, and 4/34 against
\texttt{fox\_d21\_ditto\_v4}, the 21-frame Cody-conditioned Fox specialist.
Step 2,385 wins 5/34 against \texttt{gm} and 3/34 against
\texttt{fox\_d21\_ditto\_v4}. The stock-margin differences from step 1,318 are
$+0.44$ $[-0.12,+1.00]$ and $+0.03$ $[-0.62,+0.65]$ stocks per game, respectively
(95\% bootstrap intervals). We design the leash run after these matches; all four
opponents also appear in Section~\ref{sec:benchmark}.
\par
\textbf{Comparison of the three KL-weight runs.} The three runs continue from
step 1,318 with $\lambda_{\to}=\lambda_{\leftarrow}$ set to 0.03, 0.01, or 0.003,
otherwise sharing settings. The selected 0.003 run is called the leash run in
Appendix~\ref{app:benchmark}. Each has a 22-hour cap and reaches step 808, 788, or
812, respectively. The 0.003 run restarts in place once early on, renewing its cap,
and trains for 23 hours. Including all three continuations, the 10M's RL experiments
take 140 hours, compared with 72 hours for 75M.
\par
We compare the newest checkpoints of the three runs: steps 592, 568, and 544 against
\texttt{master} and \texttt{diamond}, and steps 608, 592, and 576 against
\texttt{gm} and \texttt{fox\_d18\_ditto\_v4}, an 18-frame Fox specialist, in the
same order. All three win all 50 games against \texttt{master} and
\texttt{diamond}, totaling 150 wins. Our planned criterion, mean stock margin
against \texttt{master}, favors the 0.01 run over 0.003 by 0.06 stocks per game.
Against \texttt{gm} and \texttt{fox\_d18\_ditto\_v4}, stock margins rank 0.003,
0.01, and 0.03 from best to worst in each of the three sets of 16 or 18 games.
Across all six sets, none of the six stock-margin differences between 0.003 and
0.01 has a 95\% bootstrap interval excluding zero, although damage received per
minute favors 0.003 in three sets. We depart from the planned criterion and choose
0.003 on its results against \texttt{gm} and \texttt{fox\_d18\_ditto\_v4}.
A Falco comparison after this choice, including step 632, favors 0.01 on Final
Destination within sampling uncertainty. These selection opponents partly overlap
with the final benchmark.
\par
In in-loop evaluation, all three runs reduce damage received from MIMIC from
90--93 to 24--41 percentage points per minute within about eleven hours. Over the
leash run's first 632 steps, $\mathrm{KL}(\pi_\theta\,\|\,\pi_{\mathrm{ref}})$ rises
from 0.004 at the fork, the median over its nine warm-up steps with unchanged
policy weights, to 0.14. The reverse KL reaches 0.30, and policy entropy falls from
0.74 to 0.33. These KL values use the approximations described above. Random seeds
are fixed, but emulator game randomness is not; each configuration is run once.
\par
\FloatBarrier

\par
\section{Benchmark Protocol and Additional Results}
\label{app:benchmark}
\begingroup
\setlength{\parskip}{3pt}
\subsection{Checkpoints and Evaluation Cohorts}
The \href{\benchmarkurl}{\textcolor{resourcelink}{original panel}} compares four frozen Faynt checkpoints. The subsequent RL
cohort evaluates the 75M checkpoint at RL step 980 and the 10M checkpoint at
step 632 of its leash run. These step counts belong to their respective
training runs.
The earlier suite and the expanded Slippi-AI panel are retained as separate
cohorts because their character assignments, stages, and port schedules differ.

\begin{table}[!htbp]
\caption{Frozen Faynt checkpoints. The supervised post-training stage
includes curriculum learning for both sizes and knowledge distillation for 10M.}
\label{tab:benchmark-checkpoints}
\centering\fontsize{8.5}{10}\selectfont\setlength{\tabcolsep}{11pt}
\begin{tabular}{@{}llrl@{}}\toprule
Size & Stage & Checkpoint step & Original run designation\\\midrule
75M & Pretrained & 86,016 & Final pretraining\\
75M & Supervised post-training & 127,214 & \texttt{cur-3}\\
10M & Pretrained & 122,064 & Final pretraining v2\\
10M & Supervised post-training & 195,248 & \texttt{cur-6-kd}\\
75M & RL & 980 & RL run\\
10M & RL & 632 & RL leash run\\\bottomrule
\end{tabular}
\end{table}

\subsection{The Original 152-Game Panel}
\label{app:benchmark-original}
The four original Faynt checkpoints face identical opponent checkpoints,
characters, policy seeds, and ports. Every game is a character mirror. The
152-game totals exclude the five shared 10M--75M games and each supervised
post-trained model's ten ancestor games. The HAL sets below are supplementary.

\begin{table}[!htbp]
\caption{Original external-panel allocation for each Faynt checkpoint.}
\label{tab:benchmark-original-allocation}
\centering\fontsize{8.5}{10}\selectfont\setlength{\tabcolsep}{5pt}
\begin{tabular}{@{}p{112pt}rp{238pt}@{}}\toprule
Block & Games & Assignment\\\midrule
MIMIC Master Fox & 10 & Fox mirrors against \texttt{fox-master}\\
CPU9 Fox & 10 & Fox mirrors against the level-9 controller\\
MIMIC core mirrors & 12 & Six native characters, two games each\\
Slippi-AI core mirrors & 24 & Twelve native characters, two games each\\
Roster extension & 96 & Allocation in Table~\ref{tab:benchmark-extension}\\\bottomrule
\end{tabular}
\end{table}

The common six-character core is Fox, Falco, Marth, Sheik, Captain Falcon,
and Luigi. The Slippi-AI core adds Jigglypuff, Peach, Yoshi, Ice Climbers,
Pikachu, and Samus. MIMIC uses the corresponding character-trained release;
Slippi-AI uses one shared \texttt{medium-v2} policy for its twelve core
characters~\cite{mimic,slippiai}. Faynt is trained for all 26 fighters.

Let $E$ contain the fourteen fighters outside the Slippi-AI core: Mario,
Donkey Kong, Kirby, Bowser, Link, Ness, Mewtwo, Zelda, Young Link, Dr. Mario,
Pichu, Mr. Game \& Watch, Ganondorf, and Roy.

\begin{table}[!htbp]
\caption{The 96 original extension games. Each assigned fighter is
played twice. Forced mirrors place the opponent on a fighter outside its deployed roster.}
\label{tab:benchmark-extension}
\centering\fontsize{8.5}{10}\selectfont\setlength{\tabcolsep}{5pt}
\begin{tabular}{@{}p{85pt}rp{265pt}@{}}\toprule
Opponent & Games & Character and checkpoint assignment\\\midrule
CPU9 & 28 & All fourteen fighters in $E$\\
MIMIC native & 34 & Seventeen native fighters outside the common core\\
MIMIC forced & 6 & \texttt{fox-master} on Kirby, Zelda, and Pichu\\
Slippi specialists & 4 & Donkey Kong: \texttt{dk\_d18\_imitation\_v2};
Dr. Mario: \texttt{doc\_d18\_imitation\_v3}\\
Slippi forced & 24 & \texttt{medium-v2} on $E$ except Donkey Kong and Dr. Mario\\\bottomrule
\end{tabular}
\end{table}

MIMIC's seventeen native extension fighters are $E$ excluding Kirby, Zelda,
and Pichu, together with the six additional Slippi-AI core characters.
CPU9 therefore covers fifteen fighters in this panel: Fox and $E$.

\par
\subsection{Execution, Scoring, and Reuse}
\label{app:benchmark-runtime}
\paragraph{Original game conditions.}
The original panel uses Melee NTSC 1.02, Final Destination, four stocks, an
eight-minute timer, items off, UCF enabled, and a 30,000-frame safety cap.
Faynt occupies port 1 and the opponent port 2 throughout. Ten-game anchors
use policy seeds 101, 211, 307, 401, 503, 601, 701, 809, 907, and 1009;
core mirrors use 1201/1301, and extension mirrors use 1401/1501. These
seeds govern policy sampling. Dolphin's game RNG is recorded separately and
is not fully determined by the policy seeds.

\paragraph{Inference and controller timing.}
Checkpoints, preprocessing, and controller decoders remain frozen. The harness
waits for the required inference calls, commits the paired controller packets,
and then advances emulation. Audited steady-state controller/replay alignment
is one frame; startup uses its separately verified transport mapping. Slippi-AI
retains its policy delay of 21 frames for \texttt{medium-v2} and 18 frames for
the two specialists. Its delayed-agent wrapper queues predicted actions before
execution; these delays remain in game frames when emulation waits for
inference~\cite{slippiai}. The expanded panel uses 21-frame delays except for
\texttt{fox\_d24\_ditto\_v4}, which uses 24
(Table~\ref{tab:benchmark-opponent-configurations}).
The mixed launcher uses zero console-delay frames,
whereas upstream \texttt{eval\_two} uses two. Results apply to this integration;
inference speed does not affect these outcome comparisons.
Section~\ref{sec:benchmark-realtime} evaluates the decision loop separately.

MIMIC's released local-play implementation predicts and sends a controller
command at every frame without an added policy-delay queue~\cite{mimic}.
Its documented v2 training setup pairs the state at frame $t$ with the
controller input at $t+1$, with the reaction-delay setting at zero. The local-play launcher uses
zero console-delay frames. The separate Direct Connect launcher configures
two frames of online delay; that setting does not describe the local benchmark.

\paragraph{Outcome and stock accounting.}
For final stocks $s_M$ and $s_O$, the focal policy takes $4-s_O$ stocks and
concedes $4-s_M$. Stock losses include self-destructs. W--L records game
outcomes, win rate is $W/(W+L)$, and mean stock margin is the difference
between total stocks taken and conceded divided by games. All reported
original-panel outcomes are decisive stock-outs. No reported original result
uses timeout or tiebreak scoring. Technical failures and frame-limit exits
receive no automatic score.

\paragraph{Acceptance and audit status.}
Strict acceptance checks checkpoint identity, state and controller continuity,
terminal outcome, and a complete attributed \texttt{.slp} replay. All 304
pretrained external games, all 304 supervised post-training external games,
and all 138 newly played baseline games passed. All nine
HAL games passed synchronization, native-controller, and independent replay
audits.

\paragraph{Baseline views and shared games.}
The original MIMIC and Slippi-AI 152-game views use the supervised
post-trained 75M checkpoint as their Faynt comparator. MIMIC reuses 62 Faynt
games (10 Fox anchors, 12 core mirrors, and 40 extension games); Slippi-AI
reuses 52 (24 core and 28 extension games), reversing outcomes and stock
counts. Their own-family core block becomes a Faynt block. The MIMIC
extension comprises 40 Faynt, 28 CPU9, and 28 Slippi-AI games; the Slippi-AI
extension comprises 28 Faynt, 28 CPU9, and 40 MIMIC games. These views
therefore have different opponent compositions. They are descriptive summaries,
and their totals are not a common-opponent ranking.

The 138 new baseline games comprise 38 CPU9 games per baseline and 62
MIMIC--Slippi-AI games. Shared games are played once. MIMIC always occupies
port 2; Slippi-AI occupies port 2 against Faynt and port 1 otherwise. CPU9
occupies the opposite port from the evaluated baseline.

\paragraph{Frozen records.}
The original schedules are identified by abbreviated SHA-256 values:
pretraining v5, \texttt{6580bcf8...d7c1e7}; supervised post-training,
\texttt{566eb5e8...a2bd}; and baseline completion,
\texttt{56d0d414...8e7ae}. Per-game summaries and scoring replays supply the
outcome records. Fixed ports, one stage, and two games per character limit
generalization. Different checkpoints also have different training exposures,
so their outcomes compare selected models across the recorded training stages.

\par
\subsection{Original Pairwise Comparisons and the HAL Supplement}
\label{app:benchmark-historical}
The full-roster comparisons below use two same-character games per fighter.
The ten-game Fox anchor and CPU9 games are excluded. The common native core
contains the six characters listed in Appendix~\ref{app:benchmark-original};
all three frameworks use native-character policies on that subset.

\begin{table}[!htbp]
\caption{Original pairwise results. Faynt checkpoints are supervised
post-trained. Margins are stocks taken minus stocks conceded per game.
The common-core columns reuse a subset of the full-roster games.}
\label{tab:benchmark-original-pairwise}
\centering\fontsize{8.5}{10}\selectfont\setlength{\tabcolsep}{5pt}
\begin{tabular}{@{}llrrrr@{}}\toprule
& & \multicolumn{2}{c}{All 26 characters} & \multicolumn{2}{c}{Common native core}\\
Policy & Opponent & W--L & Margin & W--L & Margin\\\midrule
Faynt 10M & MIMIC & 35--17 & $+1.02$ & 8--4 & $+0.67$\\
Faynt 10M & Slippi-AI & 34--18 & $+0.79$ & 4--8 & $-0.83$\\
Faynt 75M & MIMIC & 46--6 & $+1.71$ & 10--2 & $+1.08$\\
Faynt 75M & Slippi-AI & 40--12 & $+1.31$ & 7--5 & $+0.17$\\
Slippi-AI & MIMIC & 34--18 & $+0.38$ & 11--1 & $+1.42$\\\bottomrule
\end{tabular}
\end{table}

These comparisons contain 260 distinct accepted games: 208 Faynt--baseline
games and 52 MIMIC--Slippi-AI games. The full roster includes forced mirrors. Faynt 10M's Slippi-AI win rate is 65.4\% across the full roster
and 33.3\% on the common native core. The corresponding 75M rates are
76.9\% and 58.3\%. For Jigglypuff, Peach, Yoshi, Ice Climbers, Pikachu, and
Samus, Faynt--MIMIC uses seeds 1401/1501 while other pairings use 1201/1301.
The small per-character samples and fixed ports limit these comparisons.

\paragraph{HAL checkpoint and execution.}
The supplementary Fox sets use the HAL O50 checkpoint, update
16,384~\cite{hal}, with released normalization statistics and strict native
architecture loading. Its abbreviated checkpoint SHA-256 is
\texttt{bd1301d3...bab10d0}; source revision is
\texttt{cdf84f90...0346ba}. Fox uses the recommended \texttt{IBDW\#0}
player conditioning. Both sets use Final Destination, four stocks, an
eight-minute timer, and first to three wins. Policy seeds start at zero;
HAL occupies port 1 in odd-numbered games and port 2 in even-numbered games.

The HAL checkpoint was trained with a two-frame delay. Our benchmark uses
its native D0/R2 deployment option~\cite{hal}, with zero added execution delay: it observes every frame, plans four
actions, executes the first two, and replans at 30 game-time Hz. The second
action is executed before another inference. Faynt runs at 60 game-time Hz. Both
required current-frame policy steps finish before the paired controller
transaction advances Melee. Each policy retains its native feature encoding,
history, sampling, and controller decoder.

The supervised 75M wins its five-game set 3--2, with 15--16 stocks taken;
the supervised 10M wins its four-game set 3--1, with 14--11 stocks taken.
These nine games are supplementary to the 152-game panels and the pairwise
comparisons above. The per-game records are retained in
\nolinkurl{hal-o50/bo5-20260908/series.json} and its scoring replays.
\endgroup

\par
\subsection{Expanded Slippi-AI Panel}
\label{app:benchmark-expanded}
\label{app:benchmark-series}
The final panel contains 1,312 games per RL checkpoint against fourteen
frozen releases, giving 2,624 distinct games overall. The evaluated checkpoints
are 75M RL step 980 and 10M step 632 of the second RL run.
Each game starts with four stocks. The six stages are
Battlefield, Final Destination, Dream Land, Fountain of Dreams, Yoshi's Story,
and Pok\'emon Stadium. Single-character opponents receive twelve supported
mirrors, covering all six stages and both port assignments. Multi-character
opponents receive two supported mirrors per deployed fighter. Each
outside-roster fighter receives two extended-roster games and two forced
mirrors, with reversed ports on its scheduled stage. The extension samples
assigned matchups; it does not enumerate every character pair.
\par
The four Fox specialists are \texttt{fox\_d21\_ditto\_v4} and
\texttt{fox\_d24\_ditto\_v4}, together with \texttt{SFIL} and
\texttt{fox\_d21\_ditto\_hax\_v3}. The Falco specialist is
\texttt{falco\_d21\_ditto\_v4}. Each has 25 outside-roster characters.
The seven twelve-character releases are \texttt{gm}, \texttt{master},
\texttt{diamond}, \texttt{medium-v2}, \texttt{plat}, \texttt{gold}, and
\texttt{silver}. Their deployed roster is the twelve-character Slippi-AI core
in Appendix~\ref{app:benchmark-original}; its complement is $E$.
The four-character \texttt{gm-v1} and \texttt{medium-v1} releases deploy Fox,
Falco, Marth, and Sheik. Their 22 outside-roster fighters occur in the broader
behavior-cloning declaration, so ``outside roster'' refers specifically to
the deployed configuration.
\par
Table~\ref{tab:benchmark-opponent-configurations} lists the releases,
conditioning, and delays.
\begin{table}[!htbp]
\caption{Expanded opponent configurations. Roster size is deployed character count; delay is in game frames.}
\label{tab:benchmark-opponent-configurations}
\centering\fontsize{8.5}{10}\selectfont\setlength{\tabcolsep}{4pt}
\begin{tabular*}{\textwidth}{@{\extracolsep{\fill}}lrrl@{}}\toprule
Release & Roster & Delay & Conditioning\\\midrule
\texttt{SFIL} & 1 & 21 & SFAT\\
\texttt{diamond} & 12 & 21 & Master Player\\
\texttt{falco\_d21\_ditto\_v4} & 1 & 21 & Ginger\\
\texttt{fox\_d21\_ditto\_hax\_v3} & 1 & 21 & Hax\\
\texttt{fox\_d21\_ditto\_v4} & 1 & 21 & Cody\\
\texttt{fox\_d24\_ditto\_v4} & 1 & 24 & Cody\\
\texttt{gm} & 12 & 21 & Master Player\\
\texttt{gm-v1} & 4 & 21 & Master Player\\
\texttt{gold} & 12 & 21 & Platinum Player\\
\texttt{master} & 12 & 21 & Master Player\\
\texttt{medium-v1} & 4 & 21 & Platinum Player\\
\texttt{medium-v2} & 12 & 21 & Master Player\\
\texttt{plat} & 12 & 21 & Master Player\\
\texttt{silver} & 12 & 21 & Platinum Player\\
\bottomrule\end{tabular*}
\end{table}

\subsection{Acceptance and Reruns}
\label{app:benchmark-audit}
Of the 2,624 scored games, 2,613 passed the frozen queue's strict acceptance
checks. Eleven reached its retry bound without a completed native game.
Each was rerun exactly once as a separate single-game execution using the
unchanged worker and game plan, with only the local output path changed.
All eleven reruns passed the frozen terminal validator and have retained
SLP replays. They are included once in the reported totals and identified
below; asterisks mark the corresponding cells
in the condition tables and Figure~\ref{fig:benchmark-expanded}.
\par
Every scored game has a terminal fourth-stock loss. The replay contains a
formal game-end event for 2,407 games; for the remaining 217, the worker
accepts the recorded loss of the fourth stock as the terminal outcome.
Stock totals agree with the terminal result in every case. No scored game
uses an adjudicated outcome. The full game ledger is provided in
\nolinkurl{data/rl_current_scores.json}, with aggregates in
\nolinkurl{data/rl_current_summary.json}.

\begin{minipage}{\textwidth}
\centering\fontsize{8.5}{10}\selectfont\setlength{\tabcolsep}{4pt}
\begin{tabular*}{\textwidth}{@{\extracolsep{\fill}}lllr@{}}\toprule
Faynt & Opponent release & Faynt character / condition & Stocks\\\midrule
75M & \texttt{fox\_d24\_ditto\_v4} & Ness / forced mirror & 4--0\\
75M & \texttt{gold} & Link / forced mirror & 4--1\\
75M & \texttt{medium-v1} & Mr. Game \& Watch / forced mirror & 4--2\\
75M & \texttt{medium-v2} & Pichu / forced mirror & 4--2\\
75M & \texttt{plat} & Dr. Mario / forced mirror & 4--1\\
\midrule
10M & \texttt{fox\_d21\_ditto\_hax\_v3} & Peach / forced mirror & 4--0\\
10M & \texttt{fox\_d21\_ditto\_v4} & Bowser / forced mirror & 4--0\\
10M & \texttt{fox\_d21\_ditto\_v4} & Samus / forced mirror & 4--0\\
10M & \texttt{fox\_d24\_ditto\_v4} & Kirby / extended roster & 4--2\\
10M & \texttt{fox\_d24\_ditto\_v4} & Jigglypuff / forced mirror & 4--1\\
10M & \texttt{gm-v1} & Yoshi / forced mirror & 4--2\\\bottomrule
\end{tabular*}
\end{minipage}
\par All eleven reruns are Faynt wins. Stocks read taken--conceded.

\par
\subsection{Fixed-Panel Win and Stock Ledgers}
\label{app:benchmark-ledgers}
Each checkpoint contributes 152 verified outcomes. All W--L and stock pairs read Faynt--opponent.
\par\Needspace{8\baselineskip}\vspace{6pt}\textbf{75M checkpoints}\par\vspace{4pt}
{\fontsize{8.5}{10}\selectfont\setlength{\tabcolsep}{2.5pt}
\addtocounter{table}{-1}
\renewcommand{\theHtable}{benchmark.fixed.tables.1}
\begin{longtable}{@{\extracolsep{\fill}}lrrrr@{}}\toprule
 & \multicolumn{2}{c}{Pretrained} & \multicolumn{2}{c}{Supervised}\\
Block & W--L & Stocks & W--L & Stocks\\\midrule\endfirsthead
\toprule
 & \multicolumn{2}{c}{Pretrained} & \multicolumn{2}{c}{Supervised}\\
Block & W--L & Stocks & W--L & Stocks\\\midrule\endhead
\bottomrule\endfoot
\endlastfoot

MIMIC Fox & 1--9 & 20--39 & 3--7 & 32--32\\[2pt]
CPU9 Fox & 8--2 & 38--22 & 10--0 & 40--7\\[2pt]
MIMIC core & 2--10 & 33--46 & 10--2 & 45--32\\[2pt]
Slippi-AI core & 2--22 & 55--93 & 15--9 & 82--75\\[2pt]
Extension & 56--40 & 320--278 & 85--11 & 373--195\\*[2pt]
\midrule Total & 69--83 & 466--478 & 123--29 & 572--341\\\bottomrule
\end{longtable}}
\par\vspace{6pt}\textbf{10M checkpoints}\par\vspace{4pt}
{\fontsize{8.5}{10}\selectfont\setlength{\tabcolsep}{2.5pt}
\addtocounter{table}{-1}
\renewcommand{\theHtable}{benchmark.fixed.tables.2}
\begin{longtable}{@{\extracolsep{\fill}}lrrrr@{}}\toprule
 & \multicolumn{2}{c}{Pretrained} & \multicolumn{2}{c}{Supervised}\\
Block & W--L & Stocks & W--L & Stocks\\\midrule\endfirsthead
\toprule
 & \multicolumn{2}{c}{Pretrained} & \multicolumn{2}{c}{Supervised}\\
Block & W--L & Stocks & W--L & Stocks\\\midrule\endhead
\bottomrule\endfoot
\endlastfoot

MIMIC Fox & 1--9 & 15--38 & 5--5 & 33--32\\[2pt]
CPU9 Fox & 8--2 & 38--23 & 10--0 & 40--7\\[2pt]
MIMIC core & 2--10 & 24--45 & 8--4 & 40--32\\[2pt]
Slippi-AI core & 1--23 & 38--95 & 8--16 & 71--87\\[2pt]
Extension & 37--59 & 274--316 & 75--21 & 355--219\\*[2pt]
\midrule Total & 49--103 & 389--517 & 106--46 & 539--377\\\bottomrule
\end{longtable}}
\par The corresponding total stock margins per game are $-0.08$ and $+1.52$ at
75M, and $-0.84$ and $+1.07$ at 10M, in pretrained and supervised order.
\par\Needspace{6\baselineskip}\vspace{8pt}\textbf{Original baseline views}\par
The Faynt comparator is the supervised 75M checkpoint, step 127,214. These views reuse shared games as specified in Appendix~\ref{app:benchmark-runtime}. Their opponent compositions differ from each other and from the Faynt panel.
\par\vspace{4pt}
{\fontsize{8.5}{10}\selectfont\setlength{\tabcolsep}{4pt}
\addtocounter{table}{-1}
\renewcommand{\theHtable}{benchmark.fixed.tables.3}
\begin{longtable}{@{\extracolsep{\fill}}lrrr@{}}\toprule
\textbf{MIMIC}: block & Games & W--L & Stocks\\\midrule\endfirsthead
\toprule
\textbf{MIMIC}: block & Games & W--L & Stocks\\\midrule\endhead
\bottomrule\endfoot
\endlastfoot

Faynt Fox & 10 & 7--3 & 32--32\\
CPU9 Fox & 10 & 10--0 & 40--15\\
Faynt mirrors & 12 & 2--10 & 32--45\\
Slippi-AI mirrors & 24 & 2--22 & 59--93\\
Roster extension & 96 & 24--72 & 232--337\\*
\midrule Total & 152 & 45--107 & 395--522\\\bottomrule
\end{longtable}}
\par\Needspace{10\baselineskip}\vspace{4pt}
{\fontsize{8.5}{10}\selectfont\setlength{\tabcolsep}{4pt}
\addtocounter{table}{-1}
\renewcommand{\theHtable}{benchmark.fixed.tables.4}
\begin{longtable}{@{\extracolsep{\fill}}lrrr@{}}\toprule
\textbf{Slippi-AI}: block & Games & W--L & Stocks\\\midrule\endfirsthead
\toprule
\textbf{Slippi-AI}: block & Games & W--L & Stocks\\\midrule\endhead
\bottomrule\endfoot
\endlastfoot

MIMIC Master Fox & 10 & 8--2 & 37--27\\
CPU9 Fox & 10 & 10--0 & 40--4\\
MIMIC mirrors & 12 & 11--1 & 46--29\\
Faynt mirrors & 24 & 9--15 & 75--82\\
Roster extension & 96 & 34--62 & 233--322\\*
\midrule Total & 152 & 72--80 & 431--464\\\bottomrule
\end{longtable}}

\par
\subsection{Supported-Character Mirrors}
\label{app:benchmark-expanded-results}
Both agents control the same fighter inside the opponent release's deployed roster. Each entry is from Faynt\textquotesingle s perspective. Stars identify cells containing separately rerun games described in Appendix~\ref{app:benchmark-audit}.
\par\vspace{5pt}
{\fontsize{8.5}{10}\selectfont\setlength{\tabcolsep}{3pt}
\addtocounter{table}{-1}
\renewcommand{\theHtable}{benchmark.expanded.tables.1}
\begin{longtable}{@{\extracolsep{\fill}}lrrrrr@{}}\toprule
 & & \multicolumn{2}{c}{75M RL} & \multicolumn{2}{c}{10M RL}\\
\cmidrule(lr){3-4}\cmidrule(l){5-6}
Opponent release & Games & W--L & Stocks & W--L & Stocks\\\midrule\endfirsthead
\toprule
 & & \multicolumn{2}{c}{75M RL} & \multicolumn{2}{c}{10M RL}\\
\cmidrule(lr){3-4}\cmidrule(l){5-6}
Opponent release & Games & W--L & Stocks & W--L & Stocks\\\midrule\endhead
\bottomrule\endfoot
\endlastfoot

\texttt{fox\_d21\_ditto\_v4} & 12 & 2--10 & 20--46 & 12--0 & 48--4\\[1pt]
\texttt{fox\_d24\_ditto\_v4} & 12 & 3--9 & 33--45 & 12--0 & 48--6\\[1pt]
\texttt{SFIL} & 12 & 9--3 & 44--32 & 12--0 & 48--1\\[1pt]
\texttt{fox\_d21\_ditto\_hax\_v3} & 12 & 12--0 & 48--23 & 12--0 & 48--2\\[1pt]
\texttt{falco\_d21\_ditto\_v4} & 12 & 1--11 & 22--47 & 11--1 & 47--22\\[1pt]
\texttt{gm} & 24 & 9--15 & 73--83 & 24--0 & 96--41\\[1pt]
\texttt{master} & 24 & 13--11 & 81--77 & 21--3 & 93--45\\[1pt]
\texttt{diamond} & 24 & 9--15 & 74--79 & 24--0 & 96--36\\[1pt]
\texttt{medium-v2} & 24 & 17--7 & 88--63 & 24--0 & 96--27\\[1pt]
\texttt{plat} & 24 & 20--4 & 89--60 & 24--0 & 96--34\\[1pt]
\texttt{gold} & 24 & 21--3 & 93--49 & 24--0 & 96--18\\[1pt]
\texttt{silver} & 24 & 23--1 & 95--42 & 24--0 & 96--17\\[1pt]
\texttt{gm-v1} & 8 & 3--5 & 22--28 & 8--0 & 32--5\\[1pt]
\texttt{medium-v1} & 8 & 7--1 & 31--14 & 8--0 & 32--5\\*[1pt]
\midrule Total & 244 & 149--95 & 813--688 & 240--4 & 972--263\\\bottomrule
\end{longtable}}
\par All cells use only the expanded cohort. Stocks are taken--conceded, including self-destructs.
\par Single-character releases contribute 12 native mirrors each (six stages, two ports). The seven 12-character releases contribute 24 each, and the two four-character releases contribute eight each: $5\times12+7\times24+2\times8=244$ games per Faynt model.
\par
\subsection{Extended-Roster Matches}
\label{app:benchmark-extended-results}
Faynt controls a fighter outside the opponent's deployed roster; the opponent retains an assigned supported fighter. Each entry is from Faynt\textquotesingle s perspective. Stars identify cells containing separately rerun games described in Appendix~\ref{app:benchmark-audit}.
\par\vspace{5pt}
{\fontsize{8.5}{10}\selectfont\setlength{\tabcolsep}{3pt}
\addtocounter{table}{-1}
\renewcommand{\theHtable}{benchmark.expanded.tables.2}
\begin{longtable}{@{\extracolsep{\fill}}lrrrrr@{}}\toprule
 & & \multicolumn{2}{c}{75M RL} & \multicolumn{2}{c}{10M RL}\\
\cmidrule(lr){3-4}\cmidrule(l){5-6}
Opponent release & Games & W--L & Stocks & W--L & Stocks\\\midrule\endfirsthead
\toprule
 & & \multicolumn{2}{c}{75M RL} & \multicolumn{2}{c}{10M RL}\\
\cmidrule(lr){3-4}\cmidrule(l){5-6}
Opponent release & Games & W--L & Stocks & W--L & Stocks\\\midrule\endhead
\bottomrule\endfoot
\endlastfoot

\texttt{fox\_d21\_ditto\_v4} & 50 & 7--43 & 106--191 & 35--15 & 177--127\\[1pt]
\texttt{fox\_d24\_ditto\_v4} & 50 & 17--33 & 135--174 & 43--7$^{*}$ & 188--100\\[1pt]
\texttt{SFIL} & 50 & 15--35 & 146--174 & 41--9 & 188--98\\[1pt]
\texttt{fox\_d21\_ditto\_hax\_v3} & 50 & 30--20 & 173--148 & 46--4 & 193--72\\[1pt]
\texttt{falco\_d21\_ditto\_v4} & 50 & 4--46 & 97--193 & 35--15 & 175--128\\[1pt]
\texttt{gm} & 28 & 4--24 & 53--108 & 16--12 & 94--88\\[1pt]
\texttt{master} & 28 & 6--22 & 62--103 & 18--10 & 96--76\\[1pt]
\texttt{diamond} & 28 & 5--23 & 60--105 & 17--11 & 92--73\\[1pt]
\texttt{medium-v2} & 28 & 14--14 & 89--87 & 23--5 & 104--56\\[1pt]
\texttt{plat} & 28 & 12--16 & 89--93 & 26--2 & 110--58\\[1pt]
\texttt{gold} & 28 & 19--9 & 103--74 & 28--0 & 112--41\\[1pt]
\texttt{silver} & 28 & 25--3 & 106--62 & 26--2 & 107--40\\[1pt]
\texttt{gm-v1} & 44 & 10--34 & 107--164 & 31--13 & 154--114\\[1pt]
\texttt{medium-v1} & 44 & 30--14 & 153--118 & 42--2 & 173--76\\*[1pt]
\midrule Total & 534 & 198--336 & 1,479--1,794 & 427--107 & 1,963--1,147\\\bottomrule
\end{longtable}}
\par Each release contributes two games per outside-roster character. This gives $5\times50+7\times28+2\times44=534$ games per Faynt model in this condition.
\par
\subsection{Forced-Character Mirrors}
\label{app:benchmark-forced-results}
Both agents control the same fighter outside the opponent release's deployed roster. Each entry is from Faynt\textquotesingle s perspective. Stars identify cells containing separately rerun games described in Appendix~\ref{app:benchmark-audit}.
\par\vspace{5pt}
{\fontsize{8.5}{10}\selectfont\setlength{\tabcolsep}{3pt}
\addtocounter{table}{-1}
\renewcommand{\theHtable}{benchmark.expanded.tables.3}
\begin{longtable}{@{\extracolsep{\fill}}lrrrrr@{}}\toprule
 & & \multicolumn{2}{c}{75M RL} & \multicolumn{2}{c}{10M RL}\\
\cmidrule(lr){3-4}\cmidrule(l){5-6}
Opponent release & Games & W--L & Stocks & W--L & Stocks\\\midrule\endfirsthead
\toprule
 & & \multicolumn{2}{c}{75M RL} & \multicolumn{2}{c}{10M RL}\\
\cmidrule(lr){3-4}\cmidrule(l){5-6}
Opponent release & Games & W--L & Stocks & W--L & Stocks\\\midrule\endhead
\bottomrule\endfoot
\endlastfoot

\texttt{fox\_d21\_ditto\_v4} & 50 & 50--0 & 200--18 & 50--0$^{*}$ & 200--13\\[1pt]
\texttt{fox\_d24\_ditto\_v4} & 50 & 50--0$^{*}$ & 200--19 & 50--0$^{*}$ & 200--9\\[1pt]
\texttt{SFIL} & 50 & 50--0 & 200--20 & 50--0 & 200--17\\[1pt]
\texttt{fox\_d21\_ditto\_hax\_v3} & 50 & 50--0 & 200--11 & 50--0$^{*}$ & 200--7\\[1pt]
\texttt{falco\_d21\_ditto\_v4} & 50 & 50--0 & 200--16 & 50--0 & 200--16\\[1pt]
\texttt{gm} & 28 & 25--3 & 109--39 & 28--0 & 112--22\\[1pt]
\texttt{master} & 28 & 24--4 & 108--43 & 28--0 & 112--21\\[1pt]
\texttt{diamond} & 28 & 26--2 & 109--38 & 28--0 & 112--19\\[1pt]
\texttt{medium-v2} & 28 & 28--0$^{*}$ & 112--23 & 28--0 & 112--10\\[1pt]
\texttt{plat} & 28 & 28--0$^{*}$ & 112--30 & 28--0 & 112--14\\[1pt]
\texttt{gold} & 28 & 28--0$^{*}$ & 112--23 & 28--0 & 112--13\\[1pt]
\texttt{silver} & 28 & 28--0 & 112--16 & 28--0 & 112--13\\[1pt]
\texttt{gm-v1} & 44 & 24--20 & 145--139 & 42--2$^{*}$ & 174--74\\[1pt]
\texttt{medium-v1} & 44 & 42--2$^{*}$ & 174--84 & 43--1 & 175--46\\*[1pt]
\midrule Total & 534 & 503--31 & 2,093--519 & 531--3 & 2,133--294\\\bottomrule
\end{longtable}}
\par Each release contributes two games per outside-roster character. This gives $5\times50+7\times28+2\times44=534$ games per Faynt model in this condition.

\par
\subsection{Full-Roster Mirror Records}
\label{app:benchmark-full-roster}
Table~\ref{tab:benchmark-full-roster} combines supported and forced mirrors from the expanded suite. Every release is evaluated across all 26 characters. Specialists receive twelve native-character mirrors and two per other fighter, totaling 62 games. Multi-character releases receive two per fighter, totaling 52. These are scheduled game totals; the specialist native character receives greater weight. Extended-roster games are excluded.
\par\vspace{4pt}
\begin{table}[!htbp]
\caption{Full-roster mirror results after RL. Entries are Faynt wins--losses against each frozen release. Supported and forced components are listed separately in Appendices~\ref{app:benchmark-expanded-results} and~\ref{app:benchmark-forced-results}.}
\label{tab:benchmark-full-roster}
\centering\fontsize{9}{11}\selectfont\setlength{\tabcolsep}{12pt}
\begin{tabular}{@{}lrrr@{}}\toprule
Opponent release & Games & 75M RL & 10M RL\\\midrule
\texttt{fox\_d21\_ditto\_v4} & 62 & 52--10 & 62--0\\[2pt]
\texttt{fox\_d24\_ditto\_v4} & 62 & 53--9 & 62--0\\[2pt]
\texttt{SFIL} & 62 & 59--3 & 62--0\\[2pt]
\texttt{fox\_d21\_ditto\_hax\_v3} & 62 & 62--0 & 62--0\\[2pt]
\texttt{falco\_d21\_ditto\_v4} & 62 & 51--11 & 61--1\\[2pt]
\texttt{gm} & 52 & 34--18 & 52--0\\[2pt]
\texttt{master} & 52 & 37--15 & 49--3\\[2pt]
\texttt{diamond} & 52 & 35--17 & 52--0\\[2pt]
\texttt{medium-v2} & 52 & 45--7 & 52--0\\[2pt]
\texttt{plat} & 52 & 48--4 & 52--0\\[2pt]
\texttt{gold} & 52 & 49--3 & 52--0\\[2pt]
\texttt{silver} & 52 & 51--1 & 52--0\\[2pt]
\texttt{gm-v1} & 52 & 27--25 & 50--2\\[2pt]
\texttt{medium-v1} & 52 & 49--3 & 51--1\\[2pt]
\midrule Winning records & & 14/14 & 14/14\\\bottomrule
\end{tabular}
\end{table}

\subsection{Zero-Delay Specialist Evaluation}
\label{app:benchmark-zero-delay}
Table~\ref{tab:benchmark-zero-delay-complete} gives the native-character
mirror results against seven specialist checkpoints from
Phillip~\cite{firoiu2017}, the predecessor of Slippi-AI. Each opponent receives
16 games on Final Destination and a separate 18 games across six stages,
three per stage. The seven checkpoints cover six characters, including two
Fox specialists. The final endpoints are 10M RL step 632 and 75M RL step 980;
the earlier 75M checkpoint at step 222 is retained for the training comparison.

\begin{table}[!htbp]
\centering\fontsize{8.5}{10.5}\selectfont\setlength{\tabcolsep}{4pt}
\begin{tabular*}{\textwidth}{@{\extracolsep{\fill}}lrrrrrr@{}}
\toprule
& \multicolumn{2}{c}{10M RL 632} & \multicolumn{2}{c}{75M RL 222} & \multicolumn{2}{c}{75M RL 980}\\
\cmidrule(lr){2-3}\cmidrule(lr){4-5}\cmidrule(l){6-7}
Opponent & FD & Six stages & FD & Six stages & FD & Six stages\\
\midrule
\texttt{delay0/FoxFD} & 16--0 & 18--0 & 15--1 & 18--0 & 14--2 & 18--0\\
\texttt{delay0/FalcoFD} & 16--0 & 18--0 & 16--0 & 18--0 & 14--2 & 18--0\\
\texttt{FoxFD0} & 16--0 & 18--0 & 16--0 & 18--0 & 16--0 & 18--0\\
\texttt{MarthFD0} & 16--0 & 18--0 & 15--1 & 18--0 & 16--0 & 18--0\\
\texttt{PeachFD} & 16--0 & 18--0 & 16--0 & 18--0 & 16--0 & 18--0\\
\texttt{SheikFD} & 16--0 & 18--0 & 16--0 & 18--0 & 16--0 & 18--0\\
\texttt{FalconFalconBF} & 16--0 & 18--0 & 16--0 & 18--0 & 16--0 & 18--0\\
\midrule
Total & 112--0 & 126--0 & 110--2 & 126--0 & 108--4 & 126--0\\
Win rate & 100\% & 100\% & 98.2\% & 100\% & 96.4\% & 100\%\\
\bottomrule
\end{tabular*}
\caption{Results against zero-delay specialists. Entries are Faynt wins--losses.
FD denotes Final Destination. Each checkpoint plays 112 games in the FD
condition and 126 in the six-stage condition, using the same weights across
characters.}
\label{tab:benchmark-zero-delay-complete}
\end{table}
\par
All seven released Phillip configurations specify zero added action delay.
Their configured decision interval is two frames for \texttt{delay0/FoxFD}
and \texttt{FalconFalconBF}, four for \texttt{delay0/FalcoFD}, and three for
the remaining checkpoints. These are release settings; zero added delay
alone does not imply equal decision frequency across policies.
The source bundle includes the result records in
\nolinkurl{data/benchmark_zero_delay.json} and the release settings and
source URLs in \nolinkurl{data/benchmark_zero_delay_timing.json}.
The executed configurations and complete per-game evidence are retained
with the evaluation records.

\subsection{Privately Supplied Zero-Delay Slippi-AI Checkpoint}
\label{app:benchmark-private-slippi}
The Slippi-AI developers~\cite{slippiai} privately supplied a checkpoint with
zero added action delay and two conditioning settings, ``Master Player''
and ``Cody.'' We evaluate both settings using the same Final Destination
and six-stage protocols as above. Each Faynt checkpoint plays 16 games on
Final Destination and 18 across six stages per setting, for 68 games in
total. Both sides use zero added action delay. These games are reported
separately from the fourteen-release suite and the seven Phillip specialists.
\par
Table~\ref{tab:benchmark-private-slippi} gives results before and after RL.
The 10M RL checkpoint wins 61 of its 68 games without losing a stock.
The source table and transcribed results are included in
\nolinkurl{data/zero_delay_sources/image_0_delay_slippi.png} and
\nolinkurl{data/benchmark_private_slippi_zero_delay.json}.
The complete evaluation evidence includes the executed configurations
and per-game records for both conditioning settings.

\begin{table}[!htbp]
\centering\fontsize{8.5}{10.5}\selectfont\setlength{\tabcolsep}{4pt}
\begin{tabular*}{\textwidth}{@{\extracolsep{\fill}}lrrrrr@{}}
\toprule
& \multicolumn{2}{c}{Master Player} & \multicolumn{2}{c}{Cody} & \\
\cmidrule(lr){2-3}\cmidrule(lr){4-5}
Faynt checkpoint & FD & Six stages & FD & Six stages & Total\\
\midrule
10M supervised & 6--10 & 9--9 & 3--13 & 5--13 & 23--45\\
10M RL step 632 & 16--0 & 18--0 & 16--0 & 18--0 & 68--0\\
\addlinespace[3pt]
75M supervised & 9--7 & 8--10 & 5--11 & 3--15 & 25--43\\
75M RL step 980 & 16--0 & 17--1 & 13--3 & 12--6 & 58--10\\
\bottomrule
\end{tabular*}
\caption{Results against the privately supplied Slippi-AI checkpoint.
Entries are Faynt wins--losses. The two column groups use different
conditioning settings of the same opponent model. FD denotes Final
Destination; the six-stage condition contains three games per stage.
Supervised denotes the post-trained policy before RL.}
\label{tab:benchmark-private-slippi}
\end{table}

\FloatBarrier
\subsection{Released Match and Tournament Software}
\label{app:platform}
The \href{\platformurl}{\textcolor{resourcelink}{tournament repository}} provides local match runners and
a mirrored baseline scheduler. The \href{\benchmarkurl}{\textcolor{resourcelink}{benchmark repository}}
contains the original and expanded Faynt schedules, their local and cloud
execution tools, the reported result records, and the later Phillip and
zero-delay Slippi-AI evaluation runtime. The repositories document the commands,
source revisions, checkpoint formats, and external inputs for each path.
\par
\textbf{Policy interfaces.} The local \texttt{scripts/play} command selects
the existing Faynt, MIMIC, and Slippi-AI adapters. Faynt occupies port 1 and
can face MIMIC or Slippi-AI on port 2; a separate path supports CPU9.
Checkpoint paths, characters, stage, policy seed, and artifact label are
explicit inputs. The adapters retain each policy's feature construction,
controller decoder, state, and delay queue. They process consecutive game
frames, and the emulator advances after the required inference and paired
controller writes. Faynt's local runtime uses a continuous 256-frame
key--value cache, reset at the start of a game, FP32 arithmetic, and sampling
temperature 1. The 128-frame trajectory-context setting in the original actor
configuration is distinct from this cache's capacity.
\par
The local adapters validate supported checkpoint identities and formats. The
later evaluation runtime in \texttt{evaluation/} loads the native
\texttt{checkpoint.pt} files for all six Faynt releases and exposes
\texttt{melee\_rl.release\_cli} for inspecting and running the recorded
Phillip and Slippi-AI configurations. Integrating another policy requires a
source adapter that supplies its observations, controller commands, state
reset, and timing behavior, together with its source and checkpoint identity
checks. The command-line policy selectors enumerate the implemented adapters.
\par
\textbf{Scheduling.} The standard \texttt{scripts/tournament} scheduler
registers MIMIC's Fox checkpoint and Slippi-AI's \texttt{medium-v2} checkpoint.
For each requested stage and policy seed, an even number of Fox-mirror games
is divided equally between the two port assignments. Each mirrored pair
shares a derived policy-sampling seed; a separate order seed determines the
execution order. Policy seeds leave Dolphin's game randomness uncontrolled.
Every scheduled game runs in a fresh process. The Faynt benchmark queues
use the character allocations and acceptance rules of
Appendices~\ref{app:benchmark-original} and~\ref{app:benchmark-expanded}.
\par
\textbf{Scoring and resumption.} Tournament acceptance checks the declared
inputs, policy and controller traces, and an attributed scoring replay with
a conclusive result. A win earns one point and a draw earns one half;
technical failures invalidate the run. Atomic JSON reports retain the
schedule, attempts, source and asset hashes, per-port counts, and matched
port blocks. On resumption, the runner verifies the same configuration,
implementation, host, and input identities and revalidates accepted evidence
before continuing unstarted games. An interrupted or rejected game closes
the run as invalid. Its aggregates include descriptive Wilson intervals
over decisive games; paired inferential analysis remains a separate analysis
step.
\par
\textbf{External inputs.} Operators supply Melee NTSC 1.02, the declared
frame-synchronized Dolphin build, opponent source checkouts and checkpoints,
and the pinned Python environments. The local runtime also requires its
replay canary for cross-runtime input verification. Cloud runners require
an operator-prepared image and asset volumes. The release includes source
and dependency notices; the game image, emulator binaries, and third-party
checkpoint payloads are supplied separately.

\par
\section{Training-Stage Behavior Analysis}
\label{app:behavior}
This appendix gives the metric definitions, replay counts, and additional
results for Section~\ref{sec:behavior}. We compare damage received, stock
leads, and game outcomes after pretraining and supervised post-training.

\subsection{Replay Selection and Evaluation Conditions}
\label{app:behavior-comparability}
The two-stage analysis uses the same scheduled configurations at both
stages: 145 games per checkpoint for 75M and 139 for 10M. These fixed matched
cohorts retain the game selection used for the reported behavioral calculations.
The complete initial benchmark contains 152 verified outcomes per checkpoint.
All games come from the initial suite. Section~\ref{sec:benchmark}
describes the larger final evaluation.
\par
The selected replays pass the extraction checks for file size, SHA-256,
contiguous frames, initial and final stocks, and winner. There are no repeated
raw frames or further exclusions during extraction.
\par
Paired games retain the learned opponents' configurations, character
assignments, stage, ports, and policy evaluation seeds. The CPU's game-start
random seeds differ across training-stage evaluations. All results use the delay and decoder
settings specified for the benchmark.
\subsection{Measurement Definitions}
\label{app:behavior-metrics}
We extract Slippi player states~\cite{slippi-tools} from frame zero, after the
countdown, and count each playable simulation frame once. A game ends when
either player first reaches zero stocks. Durations use simulated game time.

\paragraph{Damage received per minute.}
For player $p$ in game $g$, let $q_{p,t}$ be the recorded damage percentage
and $s_{p,t}$ the remaining stocks. We sum positive changes in damage
percentage between consecutive frames on the same stock:
\[
 D_{p,g}=\sum_{t=1}^{T_g}
 \mathbf{1}\{s_{p,t}=s_{p,t-1}\}
 \max(0,q_{p,t}-q_{p,t-1}),
 \qquad
 R_{p,g}=\frac{60D_{p,g}}{\tau_g},
\]
where $T_g$ is the final frame index and $\tau_g$ is the game's duration in seconds. The rate $R_{p,g}$ has
units of damage percentage points per minute. We average these rates with
equal weight per game. The calculation includes damage from all sources;
it excludes transitions on which the stock count changes. Damage attribution
to individual attacks or players is unavailable in these aggregates.

\paragraph{Stock lead and first-stock outcomes.}
The stock lead is Faynt's remaining stocks minus its opponent's. If a game
ends before a fixed-time comparison, we carry its final stock counts forward
to that time. We also report the fraction of games won after taking or
conceding the first stock. Games with simultaneous first-stock losses are
listed separately and excluded from both denominators. The games in each
conditional group can change across checkpoints. These statistics record
the final outcome after the first stock loss; subsequent changes of lead
and adaptation within a game require separate analysis.

\paragraph{Low-percent stock losses.}
A low-percent stock loss occurs when Faynt's damage percentage immediately
before the loss is at most 50\%. We report the count and its share of all
Faynt stock losses. Such losses can result from edgeguards, trades,
self-destructs, and other events; the threshold alone leaves their causes
undetermined.

\paragraph{Controller activity and time in action states.}
A neutral controller has no physical buttons pressed, centered sticks, and
zero analog shoulder pressure, with numerical tolerance $10^{-6}$. A
controller change occurs when any of these fields changes between adjacent
frames. The denominator includes frames in which the character cannot act.
Time in damage-reaction states counts frames with a \texttt{DAMAGE}-prefixed
state in the pinned libmelee \texttt{Action} enumeration. Shield states use
the \texttt{SHIELD} prefix, including shield stun and shield break. Each
frequency is computed within a game, then averaged equally across games.
These frequencies depend on the opponent and the situations reached during
play, as well as the policy's actions. Table~\ref{tab:behavior-occupancy}
reports these quantities on the common two-stage cohorts.

\subsection{Pretrained and Supervised Results}
\label{app:behavior-trajectories}
Table~\ref{tab:behavior-trajectories} compares the same configurations at
both stages within each model size. Both models receive less damage per
minute after supervised post-training, while their opponents receive more. They also
hold larger stock leads after one minute. The pretrained 10M's games are shorter than its
supervised successor's, and it loses substantially more often.

\par\vspace{3pt}
\begin{table}[!htbp]
\caption{Two-stage results on the same configurations. Duration
is in seconds. Damage received (dmg/min) is in damage percentage points per
minute, averaged equally across games. Stock lead is evaluated at 60 seconds.}
\label{tab:behavior-trajectories}
\centering\fontsize{8.5}{10.5}\selectfont\setlength{\tabcolsep}{3pt}
\begin{tabular*}{\textwidth}{@{\extracolsep{\fill}}llrrrrrr@{}}\toprule
Model & Stage & Games & Wins & Duration & Faynt dmg/min & Opp.\ dmg/min & Lead\\\midrule
75M & Pretrained & 145 & 65 & 193.2 & 106.4 & 122.1 & $-0.055$\\
 & Supervised & 145 & 117 & 184.5 & 95.8 & 151.4 & $+0.531$\\\addlinespace[3pt]
10M & Pretrained & 139 & 46 & 177.1 & 118.2 & 110.2 & $-0.317$\\
 & Supervised & 139 & 95 & 186.5 & 98.7 & 144.6 & $+0.460$\\\bottomrule
\end{tabular*}
\end{table}
\par
Table~\ref{tab:behavior-first-stock} reports wins after taking or conceding
the first stock, with the number of games in each group. Both players lose
their first stock simultaneously in two games for the supervised 75M; Faynt
wins both. On the common two-stage cohort, 10M's mean time to the first stock
loss changes from 34.5 to 35.5 seconds.

\par\vspace{3pt}
\begin{table}[!htbp]
\caption{Wins after taking or conceding the first stock on the
common two-stage cohorts.
Each entry gives wins divided by eligible games, followed by the percentage.
Simultaneous first-stock losses are excluded.}
\label{tab:behavior-first-stock}
\centering\fontsize{9}{11}\selectfont\setlength{\tabcolsep}{4pt}
\begin{tabular*}{\textwidth}{@{\extracolsep{\fill}}llrr@{}}\toprule
Model & Stage & Win after taking first & Win after conceding first\\\midrule
75M & Pretrained & 42/65 (64.6\%) & 23/80 (28.8\%)\\
 & Supervised & 96/107 (89.7\%) & 19/36 (52.8\%)\\\addlinespace[3pt]
10M & Pretrained & 40/57 (70.2\%) & 6/82 (7.3\%)\\
 & Supervised & 76/93 (81.7\%) & 19/46 (41.3\%)\\\bottomrule
\end{tabular*}
\end{table}
\par\vspace{3pt}
\begin{table}[!htbp]
\caption{Faynt stock losses on the common cohorts. Low-percent
losses occur at 50\% damage or less, using the last recorded value before
the loss. The final column gives mean damage percentage before a stock loss.}
\label{tab:behavior-low-percent}
\centering\fontsize{9}{11}\selectfont\setlength{\tabcolsep}{4pt}
\begin{tabular*}{\textwidth}{@{\extracolsep{\fill}}llrrr@{}}\toprule
Model & Stage & Low-percent / all losses & Share & Mean loss percent\\\midrule
75M & Pretrained & 86/456 & 18.9\% & 99.6\\
 & Supervised & 51/325 & 15.7\% & 108.9\\\addlinespace[3pt]
10M & Pretrained & 98/469 & 20.9\% & 92.2\\
 & Supervised & 47/348 & 13.5\% & 106.4\\\bottomrule
\end{tabular*}
\end{table}
\par\vspace{3pt}
\begin{table}[!htbp]
\caption{Controller activity and time in action states on the
common cohorts. Values are percentages of frames, averaged equally across
games. Damage and shield columns use the state definitions above.}
\label{tab:behavior-occupancy}
\centering\fontsize{9}{11}\selectfont\setlength{\tabcolsep}{4pt}
\begin{tabular*}{\textwidth}{@{\extracolsep{\fill}}llrrrr@{}}\toprule
Model & Stage & Neutral & Changed & Damage & Shield\\\midrule
75M & Pretrained & 26.08 & 23.25 & 11.37 & 4.61\\
 & Supervised & 25.90 & 24.02 & 10.22 & 3.19\\\addlinespace[3pt]
10M & Pretrained & 22.83 & 23.12 & 12.39 & 4.00\\
 & Supervised & 25.67 & 23.73 & 10.46 & 3.15\\\bottomrule
\end{tabular*}
\end{table}

\subsection{Stock-Margin Accounting}
\label{app:behavior-stock-ledger}
For $N$ games, total opponent stocks taken $T$, and Faynt stocks lost $C$,
mean stock margin is $(T-C)/N$. For a change between two checkpoints,
\begin{equation}
 \Delta M=\frac{\Delta T-\Delta C}{N}.
 \label{eq:stock-decomposition}
\end{equation}
Table~\ref{tab:behavior-stocks} reports stock outcomes on the full 152-game initial
suite, including the historical pretrained games excluded from the replay
analysis. Stock losses include self-destructs.

\par\vspace{3pt}
\begin{table}[!htbp]
\caption{Stock outcomes on the full initial suite. Each row contains
152 games. Taken, conceded, and margin are per-game means; remaining is mean
Faynt stocks remaining conditional on winning.}
\label{tab:behavior-stocks}
\centering\fontsize{9}{11}\selectfont\setlength{\tabcolsep}{4pt}
\begin{tabular*}{\textwidth}{@{\extracolsep{\fill}}llrrrrr@{}}\toprule
Model & Stage & Wins & Taken & Conceded & Margin & Remaining\\\midrule
75M & Pretrained & 69 & 3.066 & 3.145 & $-0.079$ & 1.884\\
 & Supervised & 123 & 3.763 & 2.243 & $+1.520$ & 2.171\\\addlinespace[3pt]
10M & Pretrained & 49 & 2.559 & 3.401 & $-0.842$ & 1.857\\
 & Supervised & 106 & 3.546 & 2.480 & $+1.066$ & 2.179\\\bottomrule
\end{tabular*}
\end{table}
\par
After supervised training, 75M takes 106 more stocks and loses 137 fewer;
the corresponding changes for 10M are 150 and 140. Each opponent starts with four
stocks, which limits further increases in stocks taken as win rates rise.

\subsection{Scope of the Comparisons}
\label{app:behavior-checkpoints}
The replay analysis compares two checkpoints per model: pretrained and
supervised post-trained. Intermediate curriculum, WSD, and
distillation checkpoints have no corresponding replay comparisons. The
study also lacks compute-matched continuation controls and independent
training replications, so the contribution of each intervention cannot be
isolated. The aggregate statistics leave recovery success and the retention
of specific teacher behaviors unresolved.
\par
Weighted validation ranks the four pre-RL checkpoints in the same order as
win rate. Because it also selected the supervised checkpoints, this agreement
provides limited evidence of predictive value beyond these four policies.
The separate overall, Master-winner, and weighted losses are given in
Appendix~\ref{app:posttraining-checkpoints}.

\section{Inference Cost and Optimization}
\label{app:inference}
This appendix details the measurements summarized in Section~\ref{sec:benchmark-realtime}: the
setup (Appendix~\ref{app:inference-setup}), how the optimized implementation removes the reference
implementation's overhead (Appendix~\ref{app:inference-optimized}), the equivalence of their
outputs (Appendix~\ref{app:inference-parity}), and the CPU and GPU results
(Appendices~\ref{app:inference-cpu} and~\ref{app:inference-gpu}).

\subsection{Measurement Setup}
\label{app:inference-setup}
\textbf{Loop and timing.} For every frame, the loop records when the frame was due, when the loop
picked it up, when the model input was built, when the decision reached the host, and when the
controller command was issued. The \emph{processing time} runs from pickup to command, and its mean
over a run divided by the 16.67~ms frame is the RTF. \emph{Latency} runs from the due time to the
command, so it includes any wait for a loop that fell behind; a decision is \emph{late}, or misses the
frame, when its latency exceeds one frame. Frames are due every 1/60~s of wall-clock time whether or not the loop
is ready, as in live play. A loop that is slower than the game on average therefore \emph{falls
behind}: its frames queue, its latency grows with every frame, and we report how far behind the
game it is at the run's last frame. A loop that is fast enough on average can still be late on single
decisions, which the share of late decisions counts. Kernels are
loaded and the CUDA graph~\cite{cudagraphs} is captured before a run starts. Each Modal run then processes 300
unmeasured frames, which fill the 256-frame cache, and 1,200 measured frames, or 20~s of play; the
laptop runs of Appendix~\ref{app:inference-cpu} measure 1,800 frames, or 900 with trained weights.
\par
\textbf{Stages.} The input build converts a game state into the model's per-frame features and,
on a GPU, copies them to the device; synthetic frames are already in that format and need only the
copy, as one packed block per data type for the optimized implementation. The policy step runs the
frame encoder, one cached Transformer step over a 256-frame window ending at the current frame, and
the joint-controller head. Sampling draws the joint-controller category and then the main-stick
position conditioned on it (Equation~\ref{eq:controller-factorization}). The host decodes the
sample and issues it through libmelee's controller interface~\cite{libmelee} to a stub that
records the calls, so writing to an emulator's input pipe is not timed.
\par
\textbf{Frames and roles.} Synthetic runs cycle through 16 random valid game-state frames
(\emph{synthetic frames}), already in the model's input format. Replay runs read a recorded
two-player game (Fox versus Link on Pok\'emon Stadium) frame by frame from its \texttt{.slp}
file~\cite{slippi-tools} through libmelee's replay reader, which yields the same game-state objects as a connection to a
running game; no emulator ran and no game image was used. To \emph{play}, one forward pass computes
one row, and the policy's own samples become its previous controller input. To \emph{watch}, as a
coaching tool would, one forward pass computes one row per player, each given that player's
recorded inputs, and the loop reads back each player's per-head log-probability of its next input,
with the stick head conditioned on the sampled rather than the player's category. Batched runs
place several games in one forward pass, one row per game, on synthetic frames.
\par
\textbf{Hardware and software.} All runs except the laptop comparison in
Appendix~\ref{app:inference-cpu} use Modal~\cite{modal} cloud containers in a gVisor~\cite{gvisor} sandbox,
with PyTorch 2.11.0~\cite{paszke2019}, its CUDA 12.8 build wherever a GPU is attached, and FP32 arithmetic outside the settings
study. Each GPU type (NVIDIA T4, L4, or A10G) ran in two containers requesting 8 CPU cores and
16~GiB of memory, a first for the sync-free and optimized runs and a second for the reference runs
and the settings study; the output-parity test ran in a third T4 container. The second A10G
request was served by an NVIDIA A10, so that group's reference and settings runs were measured on
an A10. The CPU study requests 4 CPU cores and 8~GiB on three
of Modal's CPU-only hosts and on three T4 hosts used for their processors, with the GPU idle. The
sandbox hides the processor names, but the CPU flags identify the first class as AMD processors
with AVX2 and no AVX-512 (EPYC of the Zen~1--3 generations) and the second as Intel Xeon processors
with AVX-512 and VNNI but neither AVX-512 BF16 nor Ice Lake's VBMI (Cascade Lake). The hosts are shared with other tenants,
so we report the median of the three hosts; each GPU value comes from a single run. The CPU study
uses one PyTorch thread, and the GPU runs use four.
\par
\textbf{Weights.} Timing runs use random weights of each architecture, because the sequence of
operations does not depend on the weight values. On the laptop CPU of
Appendix~\ref{app:inference-cpu}, two trained RL checkpoints from the evaluated models' lineages,
the 10M at RL step 1,318 and the 75M at RL step 222, ran 6.2\% and 6.0\% faster than random weights
on the recorded game, in runs of 900 rather than 1,800 measured frames. This comparison found no
timing penalty from trained weights on that CPU.

\subsection{How the Optimized Implementation Works}
\label{app:inference-optimized}
\textbf{Why one game is bound by the host.} For one game, a policy step is a long chain of small
tensor operations, each launched from Python, with little arithmetic in any of them. On a GPU, the
step therefore takes about as long as the host needs to issue its operations, and every read of a
value back to the host adds a wait: the host stops until the device has finished all queued work,
and the device then idles until the host issues the next operation. Our measurements bear this out.
Settings that reduce the arithmetic barely change even the optimized step's processing time, or
make it longer, whereas removing the reads and then the individual launches makes the loop several
times faster (Appendix~\ref{app:inference-gpu}).
\par
\textbf{What the optimized implementation changes.} It reuses the policy's modules and parameters
and changes only how a single-frame step is orchestrated (Table~\ref{tab:inference-mechanisms}).
The main change concerns the key-value cache. Both implementations keep each layer's keys and
values in a ring of 256 frames and apply the rotary position phases~\cite{su2021} to the keys before storing
them, so attention does not depend on the order of the stored frames. The reference implementation
nevertheless reads each row's cache bookkeeping back to the host in every layer and copies the
window into chronological order before attending; the optimized implementation writes and attends
in place, in storage order, through a validity mask. With these reads and the input checks
removed, only the sampled action returns to the host when playing: this is the sync-free step. On
a GPU, a CUDA graph captured before the run then replays the frame encoder, the Transformer step,
and the joint-controller logits with one launch per frame. The two sampling draws and the stick
head, whose logits depend on the sampled category, run outside the graph and consume the
random-number generator exactly as the reference implementation does.
\begin{table}[!htbp]
  \centering
  \small
  \begin{tabular}{@{}p{0.19\textwidth}p{0.37\textwidth}p{0.37\textwidth}@{}}
    \toprule
    Per frame & Reference & Optimized \\
    \midrule
    Cache write & A loop over the batch in Python reads each row's validity flag and write
    position to the host, in every layer & One indexed write each for keys and values per
    layer; positions stay on the device \\
    \addlinespace[2pt]
    Cache read & Each layer's 256-frame window is copied into chronological order, reading each
    row's length and next write position to the host & Attention runs over the ring in storage
    order through a validity mask, which is exact because rotary phases are applied when keys are
    written \\
    \addlinespace[2pt]
    Grouped-query\newline attention~\cite{ainslie2023} & PyTorch's scaled dot-product attention on FP32 inputs expands
    keys and values to every query head & Batched matrix products per key-value head, without
    expansion \\
    \addlinespace[2pt]
    Checks & Eleven encoder range checks, three cache checks, a reset check, and the controller
    head's label checks, each a device-to-host read & None; inputs are valid by construction \\
    \addlinespace[2pt]
    Input transfer & One host-to-device copy per input tensor & One copy per data type into static
    buffers (GPU) \\
    \addlinespace[2pt]
    Dispatch & Every kernel launched from Python & One CUDA-graph replay, with sampling and the
    stick head outside the graph (GPU) \\
    \bottomrule
  \end{tabular}
  \vspace{4pt}
  \caption{Per-frame operations of the two implementations. The optimized implementation computes
  the same formulas with the same parameters.}
  \label{tab:inference-mechanisms}
\end{table}

\subsection{Output Parity}
\label{app:inference-parity}
To test equivalence, we ran both implementations with the evaluated RL checkpoints, the 10M at RL
step 632 and the 75M at RL step 980, on a Modal T4 container and on its host CPU, an Intel Xeon with
AVX-512 but no VNNI, a host of neither CPU-study class; we did not test parity on the EPYC hosts.
Each implementation processed the first 1,200 frames of the recorded game from both players'
perspectives, fed the same frames and the same recorded previous inputs, and we compared the logits
of both controller heads at every decision; the stick head's logits are conditioned on each
implementation's own sampled joint-controller category. We then let each implementation play from
the same random seed, feeding back its own samples, and recorded the first frame on which the chosen
actions differed. Table~\ref{tab:inference-parity} shows that the reference and sync-free outputs
differ only by floating-point rounding. The largest logit difference is $6.4\times10^{-4}$, the
per-head KL divergence between the action distributions never exceeds $1.2\times10^{-10}$~nats, and
neither the most likely actions nor the played actions differ in any of the 1,200 frames. The
residual comes from summation order, because the sync-free step evaluates the same formulas with
different kernels. The CUDA graph replays the sync-free step's own kernels, so its outputs match
that step bit for bit.
\begin{table}[!htbp]
  \centering
  \footnotesize
  \setlength{\tabcolsep}{4pt}
  \begin{tabular}{@{}lllrrrrl@{}}
    \toprule
    Model & Device & Pair & Max $|\Delta\text{logit}|$ & Mean KL & Max KL & Top-1 & Play \\
    \midrule
10M & CPU & Reference vs.\ sync-free & $6.1\times10^{-5}$ & $1.5\times10^{-13}$ & $2.6\times10^{-11}$ & 0 of 4,800 & none in 1,200 \\
10M & GPU & Reference vs.\ sync-free & $6.1\times10^{-5}$ & $1.5\times10^{-13}$ & $1.6\times10^{-11}$ & 0 of 4,800 & none in 1,200 \\
10M & GPU & Sync-free vs.\ optimized & 0 & 0 & 0 & 0 of 4,800 & none in 1,200 \\
    \addlinespace[2pt]
75M & CPU & Reference vs.\ sync-free & $6.4\times10^{-4}$ & $1.5\times10^{-13}$ & $1.2\times10^{-10}$ & 0 of 4,800 & none in 1,200 \\
75M & GPU & Reference vs.\ sync-free & $4.0\times10^{-4}$ & $2.0\times10^{-13}$ & $1.2\times10^{-10}$ & 0 of 4,800 & none in 1,200 \\
75M & GPU & Sync-free vs.\ optimized & 0 & 0 & 0 & 0 of 4,800 & none in 1,200 \\
    \bottomrule
  \end{tabular}
  \vspace{4pt}
  \caption{Output parity between implementations on the first 1,200 frames of the recorded game,
  from both players' perspectives, with the evaluated checkpoints (10M RL step 632, 75M RL step 980)
  on a Modal T4 container (GPU rows) and its host CPU (CPU rows). The first three columns of numbers
  compare the logits of both controller heads; KL divergences between action distributions are per
  head and in nats. Top-1 counts head outputs whose most likely category differs (1,200 frames
  $\times$ 2 players $\times$ 2 heads). Play reports the first frame on which actions sampled from the
  same seed differ. Optimized: the sync-free step replayed from a CUDA graph.}
  \label{tab:inference-parity}
\end{table}

\subsection{CPU Results}
\label{app:inference-cpu}
\begin{figure}[!htbp]
  \centering
  \includegraphics[width=\textwidth]{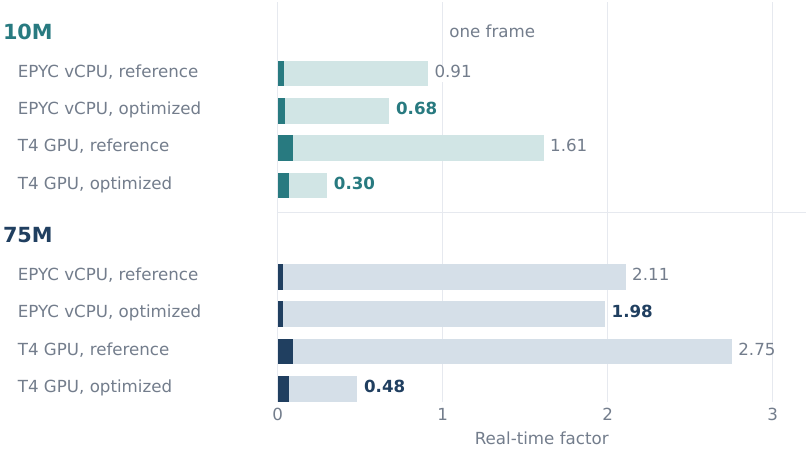}\par
  \caption{\textbf{Where the time goes.} Real-time factor of the decision loop, paced at 60~Hz with
  Faynt playing one game, split into building the model input (dark segment) and the policy step
  with sampling, decoding, and the controller write (light segment), teal for the 10M and navy for
  the 75M. EPYC rows are the median host's own run on the recorded game; T4 rows use synthetic
  frames, and their input build includes the copy to the GPU. The dashed line marks one
  frame.}\label{fig:inference}
\end{figure}
The cloud CPU study uses one PyTorch thread. When playing, the
optimized implementation saves 6--26\% of the reference's processing time. The policy step with
sampling and decoding accounts for 93--99\% of the processing time when playing
(Figure~\ref{fig:inference} shows the split on the EPYC): building the input takes 0.5--0.7~ms and
issuing the command under 0.1~ms. The shared hosts differ. The optimized 10M's 99th-percentile
latency on the EPYC is 14.0--18.6~ms across the three hosts and exceeds the frame on one of them,
where 6.0\% of its decisions are late, against 0 and 1.0\% on the other two. No other run of the cloud CPU study keeps pace. The 10M's reference on the EPYC and its optimized implementation on the Xeon average just
under a frame, but 47\% and 98\% of their decisions are late. The other runs fall behind: by
their last frame, the Xeon's reference 10M is 2--4~s behind the game and every 75M run 18--52~s, across
hosts. Following both players keeps pace on no CPU host: on the Xeon it
falls behind, and on the EPYC the optimized implementation is late on 2--96\% of decisions,
depending on the host.
\par
For comparison, we ran the loop on a laptop CPU not shared with other cloud tenants (Intel Core
Ultra 9 285H, in a Linux container under WSL2, outside Modal) with four PyTorch threads. On the
recorded game, over 1,800 measured frames, it runs the optimized 10M in 9.1~ms per frame
(99th percentile 12.3~ms) and the optimized 75M in 20.6~ms, which falls behind; the reference takes 10.7
and 23.3~ms. On this CPU the 10M therefore keeps pace with either implementation, while the 75M
falls behind on every CPU we tested.

\subsection{GPU Results}
\label{app:inference-gpu}
Figure~\ref{fig:realtime} separates the two optimizations on the T4 and L4 GPUs.
The reference and sync-free steps run eagerly, with every kernel launched from Python. On the T4
and L4, the sync-free step saves 22--35\% of the reference's processing time, but neither eager
step keeps pace on any GPU; the 10M's sync-free step on the A10G averages under a frame, but 13\%
of its decisions are late. The other eager runs fall behind: by their last frame, they are
1.3--36~s behind the game. Replaying the sync-free step's kernels from a CUDA graph then makes the
loop 3.5--7.1 times faster, so on the T4 and L4 the optimized implementation is 5.4--9.2 times
faster than the reference. With the graph, building the input becomes a noticeable share of the
processing time: on the T4 it takes 1.1~ms of the 10M's 5.0~ms, against 1.6~ms of the reference's
26.9~ms (Figure~\ref{fig:inference}). Pacing itself costs time. Fed at 60~Hz rather than back to
back, the T4's optimized step takes about 2~ms longer, probably because the GPU lowers its clock
between frames; we did not record clock speeds.
\par
Table~\ref{tab:inference-replay} repeats the optimized measurements on the recorded game.
Converting libmelee's game state into the model's features takes 0.3--0.5~ms per game and frame,
so the loop costs up to 0.6~ms more than on synthetic frames, and following both players costs
0.6--1.5~ms more than playing.
\begin{table}[!htbp]
  \centering
  \small
  \setlength{\tabcolsep}{4.5pt}
  \begin{tabular}{@{}lrrrrrrrr@{}}
    \toprule
    & \multicolumn{4}{c}{10M} & \multicolumn{4}{c}{75M} \\
    \cmidrule(lr){2-5}\cmidrule(l){6-9}
    & \multicolumn{2}{c}{Play} & \multicolumn{2}{c}{Watch} & \multicolumn{2}{c}{Play} & \multicolumn{2}{c}{Watch} \\
    \cmidrule(lr){2-3}\cmidrule(lr){4-5}\cmidrule(lr){6-7}\cmidrule(l){8-9}
    GPU & Time & p99 & Time & p99 & Time & p99 & Time & p99 \\
    \midrule
NVIDIA T4 & 5.2 & 6.8 & 6.8 & 8.9 & 8.7 & 12.1 & 9.5 & 11.9 \\
NVIDIA L4 & 3.2 & 4.7 & 4.0 & 5.5 & 5.4 & 6.4 & 6.2 & 7.5 \\
NVIDIA A10G & 3.1 & 4.6 & 3.7 & 5.1 & 4.6 & 5.9 & 5.6 & 6.9 \\
    \bottomrule
  \end{tabular}
  \vspace{4pt}
  \caption{The whole loop of the optimized implementation on the recorded game, paced at 60~Hz:
  mean processing time per frame (Time) and 99th-percentile latency (p99), in ms. Play and watch are as defined in Appendix~\ref{app:inference-setup}. Decisions later than one frame are 0.6\% for the T4 75M and 0.3\%
  for the A10G 10M when playing, and none in any other cell.}
  \label{tab:inference-replay}
\end{table}
\par
\begin{figure}[!htbp]
  \centering
  \includegraphics[width=\textwidth]{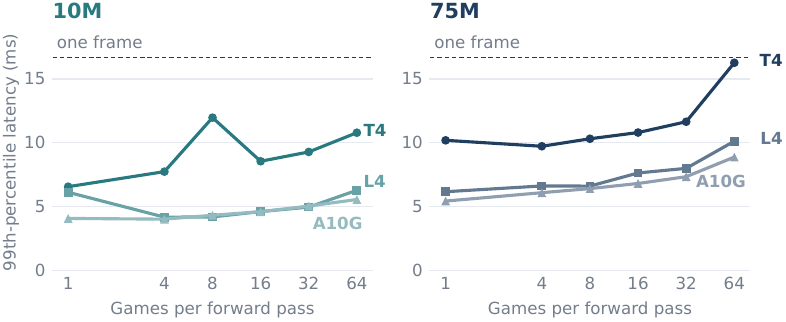}\par
  \caption{\textbf{Serving several games from one forward pass.} 99th-percentile latency of the
  optimized implementation, one row per game, paced at 60~Hz on synthetic frames, on an NVIDIA T4,
  L4, and A10G GPU; teal is the 10M and navy the 75M. The dashed line marks one frame. Each point is
  a single run, and decisions later than one frame occur only for the T4 10M at 8 games (0.8\%) and
  the T4 75M at 64 games (0.4\%).}\label{fig:inference-games}
\end{figure}

Figure~\ref{fig:inference-games} shows one forward pass serving several games, one row per game. At
64 games per forward, the 99th-percentile latency stays within 10.1~ms on the L4 and A10G for both
models; on the T4 it is 10.8~ms for the 10M and 16.3~ms, at the edge of the frame, for the 75M. In a
tight loop, with each frame started as soon as the previous one ends, one GPU completes
11,047--15,541 game-frames per second for the 10M and 4,626--8,564 for the 75M at 64 games per
forward: a throughput ceiling of 184--259 and 77--143 games at 60~Hz with the GPU fully busy, of
which we measured up to 64 within the frame budget. These runs use synthetic frames, and their
processing time includes packing the frames and issuing the commands. A server would also have to
convert every game's state, which at 64 games needs about one to two CPU cores.
\par
\textbf{Settings with little effect.} At one game per forward, even the optimized step is bound by
per-operation overhead rather than by arithmetic. Shortening the context from 256 to 64 or 128
frames or enabling TF32 matrix products changes its processing time by at most 0.3~ms, and BF16 or
FP16 autocast at context 256 makes it 0.3--1.6~ms slower, because the casts cost more than the smaller matrix
products save. Half precision would also change the outputs, so we did not adopt it.

\par
\end{document}